\documentclass[acmtog]{acmart}

\usepackage{booktabs}
\usepackage{subfigure}
\usepackage{colortbl}
\usepackage{soul}
\usepackage[normalem]{ulem}

\makeatletter
\renewcommand{\@seccntformat}[1]{%
	\ifcsname format@#1\endcsname
	\csname format@#1\endcsname
	\else
	\csname the#1\endcsname\quad 
	\fi
}
\g@addto@macro\appendix{%
	\def\format@section{Appendix \thesection: }%
}
\makeatother

\setcitestyle{square}
\definecolor{rc}{rgb}{0.93,0.93,1}

\newcommand{\figref}[1]{Figure~\ref{fig:#1}}
\newcommand{\tabref}[1]{Table~\ref{tab:#1}}
\newcommand{\eqnref}[1]{Equation~\ref{eqn:#1}}
\newcommand{\secref}[1]{Section~\ref{sec:#1}}
\newcommand{\Real}{\ensuremath{I\!R}}
\newcommand{\vect}[1]{\boldsymbol{#1}}
\newcommand{\loss}{\mathcal{L}}
\newcommand{\etal}{\emph{et al.\ }}

\newcommand{\eg}{\emph{e.g.\ }}
\newcommand{\hdrp}{H}
\newcommand{\rhdrp}{\hat{\hdrp}}
\newcommand{\yp}{\hat{y}}

\newcommand{\ldrp}{D}
\newcommand{\illp}{I}
\newcommand{\reflp}{R}
\newcommand{\hdr}{\vect{\hdrp}}
\newcommand{\rhdr}{\vect{\rhdrp}}
\newcommand{\y}{\vect{\yp}}

\newcommand{\ldr}{\vect{\ldrp}}
\newcommand{\ill}{\vect{\illp}}
\newcommand{\refl}{\vect{\reflp}}
\newcommand{\cc}{f}
\newcommand{\msk}{\alpha}
\newcommand{\feat}{\vect{h}}
\newcommand\belowfigspace{-2pt}
\newcommand\customsection[1]{\subsection{#1}}

\setcopyright{acmcopyright}
\acmJournal{TOG}
\acmYear{2017}
\acmVolume{36}
\acmNumber{6}
\acmArticle{178}
\acmMonth{11}
\acmDOI{10.1145/3130800.3130816}

\begin{document}
\title{HDR image reconstruction from a single exposure using deep CNNs}

\author{Gabriel Eilertsen}
\affiliation{%
  \institution{Link\"oping University, Sweden}
}
\email{gabriel.eilertsen@liu.se}

\author{Joel Kronander}
\affiliation{%
	\institution{Link\"oping University, Sweden}
}

\author{Gyorgy Denes}
\affiliation{%
	\institution{University of Cambridge, UK}
}

\author{Rafa{\l} K. Mantiuk}
\affiliation{%
	\institution{University of Cambridge, UK}
}

\author{Jonas Unger}
\affiliation{%
	\institution{Link\"oping University, Sweden}
}

\renewcommand{\shortauthors}{G. Eilertsen, J. Kronander, G. Denes, R. K. Mantiuk and J. Unger}

\begin{abstract}
Camera sensors can only capture a limited range of luminance simultaneously, and in order to create high dynamic range (HDR) images a set of different exposures are typically combined. In this paper we address the problem of predicting information that have been lost in saturated image areas, in order to enable HDR reconstruction from a single exposure. We show that this problem is well-suited for deep learning algorithms, and propose  a deep convolutional neural network (CNN) that is specifically designed taking into account the challenges in predicting HDR values. To train the CNN we gather a large dataset of HDR images, which we augment by simulating sensor saturation for a range of cameras. To further boost robustness, we pre-train the CNN on a simulated HDR dataset created from a subset of the MIT Places database.
We demonstrate that our approach can reconstruct high-resolution visually convincing HDR results in a wide range of situations, and that it generalizes well to reconstruction of images captured with arbitrary and low-end cameras that use unknown camera response functions and post-processing. 
Furthermore, we compare to existing methods for HDR expansion, and show high quality results also for image based lighting.
Finally, we evaluate the results in a subjective experiment performed on an HDR display. This shows that the reconstructed HDR images are visually convincing, with large improvements as compared to existing methods.
\end{abstract}

\ccsdesc[300]{Computing methodologies~Image processing}
\ccsdesc[300]{Computing methodologies~Neural networks}

\keywords{HDR reconstruction, inverse tone-mapping, deep learning, convolutional network}

\begin{teaserfigure}
	\centering
    \includegraphics[width=\textwidth]{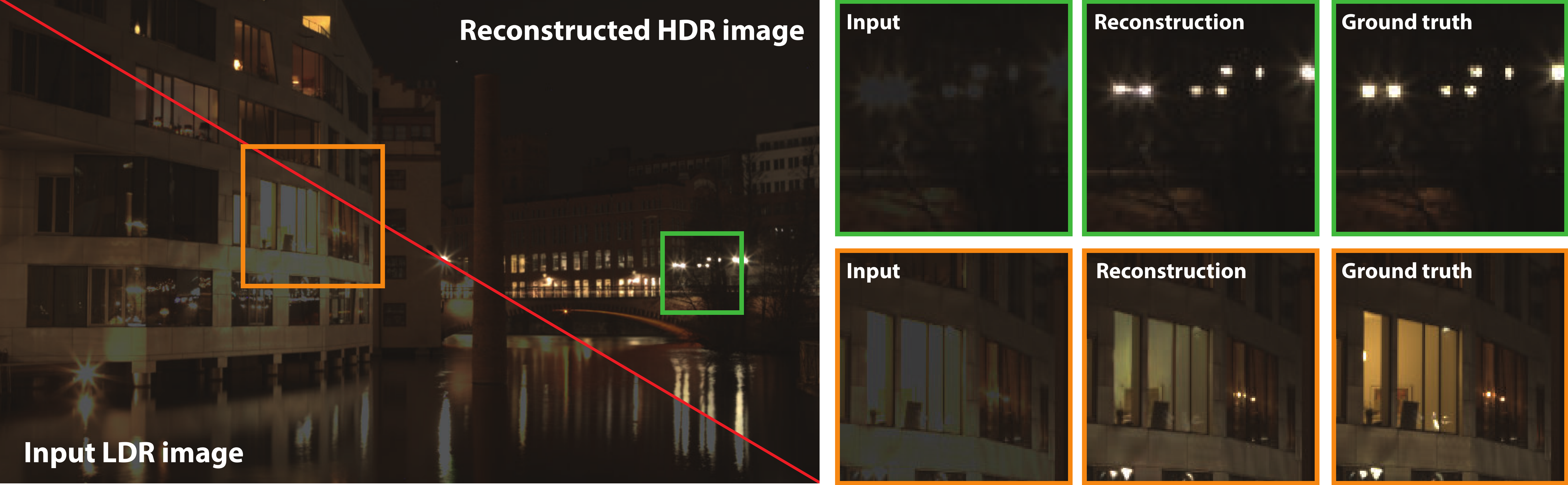}
    \caption{The exposure of the input LDR image in the bottom left has been reduced by $3$ stops, revealing loss of information in saturated image regions. Using the proposed CNN trained on HDR image data, we can reconstruct the highlight information realistically (top right). The insets show that the high luminance of the street lights can be recovered (top row), as well as colors and details of larger saturated areas (bottom row). The exposures of the insets have been reduced by $5$ and $4$ stops in the top and bottom rows, respectively, in order to facilitate comparisons. All images have been gamma corrected for display.}
    \label{fig:teaser}
\end{teaserfigure}

\maketitle


\section{Introduction}

High dynamic range (HDR) images can significantly improve the viewing experience -- viewed on an HDR capable display or by means of tone-mapping. With the graphics community as an early adopter, HDR images are now routinely used in many applications including photo realistic image synthesis and a range of post-processing operations; for an overview see ~\cite{Reinhard2010,Banterle2011,Dufaux2016}. The ongoing rapid development of HDR technologies and cameras has now made it possible to collect the data required to explore recent advances in deep learning for HDR imaging problems.

In this paper, we propose a novel method for reconstructing HDR images from low dynamic range (LDR) input images, by estimating missing information in bright image parts, such as highlights, lost due to saturation of the camera sensor. We base our approach on a fully convolutional neural network (CNN) design in the form of a \textit{hybrid dynamic range autoencoder}. Similarly to deep autoencoder architectures~\cite{hinton2006reducing,Vincent2008}, the LDR input image is transformed by an encoder network to produce a compact feature representation of the spatial context of the image. The encoded image is then fed to an HDR decoder network, operating in the log domain, to reconstruct an HDR image. Furthermore, the network is equipped with skip-connections that transfer data between the LDR encoder and HDR decoder domains in order to make optimal use of high resolution image details in the reconstruction. For training, we first gather data from a large set of existing HDR image sources in order to create a training dataset. For each HDR image we then simulate a set of corresponding LDR exposures using a virtual camera model. The network weights are optimized over the dataset by minimizing a custom HDR loss function. 
As the amount of available HDR content is still limited we utilize transfer-learning, where the weights are pre-trained on a large set of simulated HDR images, created from a subset of the MIT Places database \cite{Zhou2014}.

Expansion of LDR images for HDR applications is commonly referred to as inverse tone-mapping (iTM). Most existing inverse tone-mapping operators (iTMOs) are not very successful in reconstruction of saturated pixels. This has been shown in a number of studies \cite{Akyuz2007,Masia2009}, in which na\"ive methods or non-processed images were more preferred than the results of those operators. The existing operators focus on boosting the dynamic range to look plausible on an HDR display, or to produce rough estimates needed for image based lighting (IBL). The proposed method demonstrates a step improvement in the quality of reconstruction, in which the structures and shapes in the saturated regions are recovered. It offers a range of new applications, such as exposure correction, tone-mapping, or glare simulation. 

The main contributions of the paper can be summarized as:

\begin{enumerate}
	\item A deep learning system that can reconstruct a high quality HDR image from an arbitrary single exposed LDR image, provided that saturated areas are reasonably small.
	\item A \textit{hybrid dynamic range autoencoder} that is tailored to operate on LDR input data and output HDR images. It utilizes HDR specific transfer-learning, skip-connections, color space and loss function.
	\item The quality of the HDR reconstructions is confirmed in a subjective evaluation on an HDR display, where predicted images are compared to HDR and LDR images as well as a representative iTMO using a random selection of test images in order to avoid bias in image selection.
	\item The HDR reconstruction CNN together with trained parameters are made available online, enabling prediction from any LDR images: {\it\url{https://github.com/gabrieleilertsen/hdrcnn}}.
\end{enumerate}


\section{Related work}

\customsection{HDR reconstruction}
In order to capture the entire range of luminance in a scene it is necessary to use some form of exposure multiplexing. While static scenes commonly are captured using multiplexing exposures in the time domain \cite{Mann1994,Debevec1997,Unger2007}, dynamic scenes can be challenging as robust exposure alignment is needed. This can be solved by techniques such as multi-sensor imaging \cite{Tocci2011,Kronander2014} or by varying the per-pixel exposure \cite{Nayar2000} or gain \cite{hajisharif2015adaptive}. Furthermore, saturated regions can be encoded in glare patterns \cite{Rouf2011} or with convolutional sparse coding \cite{Serrano2016}. However, all these approaches introduce other limitations such as bulky and custom built systems, calibration problems, or decreased image resolution. 
Here, we instead tackle the problem by reconstructing visually convincing HDR images from single images that have been captured using standard cameras without any assumptions on the imaging system or camera calibration.

\customsection{Inverse tone-mapping} 
Inverse tone-mapping is a general term used to describe methods that utilize LDR images for HDR image applications \cite{Banterle2006}. The intent of different iTMOs may vary. If it is to display standard images on HDR capable devices, maximizing the subjective quality, there is some evidence that global pixel transformations may be preferred \cite{Masia2009}.  Given widely different input materials, such methods are less likely to introduce artifacts compared to more advanced strategies. The transformation could be a linear scaling \cite{Akyuz2007} or some non-linear function \cite{Masia2009,Masia2017}. These methods modify all pixels without reconstructing any of the lost information.

A second category of iTMOs attempt to reconstruct saturated regions to mimic a true HDR image. These are expected to generate results that look more like a reference HDR, which was also indicated by the pair-wise comparison experiment on an HDR display performed by Banterle \etal \citeyear{Banterle2009}. 
Meylan \etal \citeyear{Meylan2006} used a linear transformation, but applied different scalings in highlight regions. Banterle \etal \citeyear{Banterle2006} first linearized the input image, followed by boosting highlights using an expand map derived from the median cut algorithm. The method was extended for video processing, and with additional features such as automatic calibration and cross-bilateral filtering of the expand map \cite{Banterle2008}. Rempel \etal \citeyear{Rempel2007} also utilized an expand map, but computed this from Gaussian filtering in order to achieve real-time performance. Wang \etal \citeyear{Wang2007} applied inpainting techniques on the reflectance component of highlights. The method is limited to textured highlights, and requires some manual interaction. Another semi-manual method was proposed by Didyk \etal \citeyear{Didyk2008}, separating the image into diffuse, reflections and light sources. The reflections and light sources were enhanced, while the diffuse component was left unmodified. More recent methods includes the iTMO by Kovaleski and Oliviera \citeyear{Kovaleski2014}, that focus on achieving good results over a wide range of exposures, making use of a cross-bilateral expand map \cite{Kovaleski2009}.

For an in-depth overview of inverse tone-mapping we refer to the survey by Banterle \etal \citeyear{Banterle2009b}. Compared to the existing iTMOs, our approach achieves significantly better results by learning from exploring a wide range of different HDR scenes.
Furthermore, the reconstruction is completely automatic with no user parameters and runs within a second on modern hardware.

\customsection{Bit-depth extension}
A standard 8-bit LDR image is affected not only by clipping but also by quantization. If the contrast or exposure is significantly increased, quantization can be revealed as banding artifacts. Existing methods for decontouring, or bit-depth extension, include dithering methods that use noise in order to hide the banding artifacts \cite{Daly2003}. Decontouring can also be performed using low-pass filtering followed by quantization, in order to detect false contours \cite{Daly2004}. There are also a number of edge-preserving filters used for the same purpose. In this work we do not focus on decontouring, which is mostly a problem in under-exposed images. Also, since we treat the problem of predicting saturated image regions, the bit depth will be increased with the reconstructed information.

\customsection{Convolutional neural networks} 
CNNs have recently been applied to a large range of computer vision tasks, significantly improving on the performance of classical supervised tasks such as image classification~\cite{Simonyan2014}, object detection~\cite{ren2015faster} and semantic segmentation~\cite{Long2015}, among others. Recently CNNs have also shown great promise for image reconstruction problems related to the challenges faced in inverse tone-mapping, such as compression artifact reduction \cite{Svoboda2016}, super-resolution~\cite{Ledig2016}, and colorization~\cite{Iizuka2016}.
Recent work on inpainting~\cite{Pathak2016,yang2016high} have also utilized variants of Generative Adversarial Networks (GANs)~\cite{Goodfellow2014} to produce visually convincing results. However, as these methods are based on adversarial training, results tend to be unpredictable and can vary widely from one training iteration to the next. To stabilize training, several tricks are used in practice, including restricting the output intensity, which is problematic for HDR generation. Furthermore, these methods are limited to a single image resolution, with results only shown so far for very low resolutions.

Recently, deep learning has also been successfully applied for improving classical HDR video reconstruction from multiple exposures captured over time \cite{Kalantari2017}. 
In terms of reconstructing HDR from one single exposed LDR image, the recent work by Zhang and Lalonde \citeyear{Zhang2017} is most similar to ours. They use an autoencoder~\cite{hinton2006reducing} in order to reconstruct HDR panoramas from single exposed LDR counterparts. 
However, the objective of this work is specifically to recoverer high intensities near the sun in order to use the prediction for IBL. Also, the method is only trained using rendered panoramas of outdoor environments where the sun is assumed to be in the same azimuthal position in all images.
Given these restrictions, and that predictions are limited to $128 \times 64$ pixels, the results are only applicable for IBL of outdoor scenes. Compared to this work, we propose a solution to a very general problem specification without any such assumptions, and where any types of saturated regions are considered. We also introduce several key modifications to the standard autoencoder design~\cite{hinton2006reducing}, and show that this significantly improves the performance.

Finally, it should be mentioned that the concurrent work by Endo \etal \citeyear{Endo2017} also treats inverse tone-mapping using deep learning algorithms, by using a different pipeline design. Given a single exposure input image, the method uses autoencoders in order to predict a set of LDR images with both shorter and longer exposures. These are subsequently combined using standard methods, in order to reconstruct the final HDR image.

\section{HDR reconstruction model}\label{sec:method}

\subsection{Problem formulation and constraints}\label{sec:problem}
\begin{figure}[t]
	\newcommand\ww{0.13}
	\centering
	\subfigure[$\cc^{-1}(\ldr)$]{\includegraphics[width=\ww\textwidth]{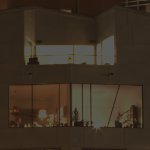}}
	\subfigure[$\exp(\y)$]{\includegraphics[width=\ww\textwidth]{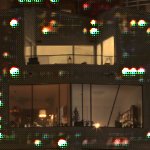}}
	\subfigure[$\vect{\msk}$]{\includegraphics[width=\ww\textwidth]{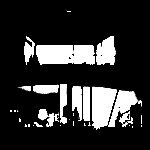}}\\
	\vspace{-5pt}
	\includegraphics[width=0.4\textwidth]{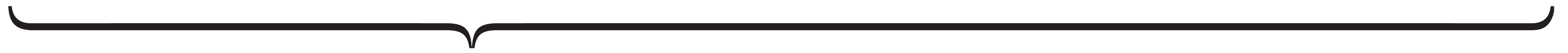}\\
	\vspace{-2pt}
	\subfigure[$\rhdr$]{\includegraphics[width=\ww\textwidth]{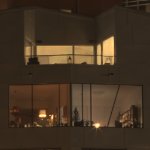}}
	\hspace{10pt}
	\subfigure[$\hdr$]{\includegraphics[width=\ww\textwidth]{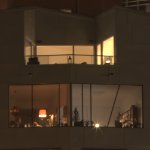}}
	\vspace{-10pt}
	\caption{\label{fig:msk} Zoom-in of an example of the components of the blending operation in \eqnref{masking}, compared to the ground truth HDR image. (a) is the input image, (b) is prediction, (c) is the blending mask, (d) is the blending of (a-b) using (c), and (e) is ground truth. Gamma correction has been applied to the images, for display purpose.}
	\vspace{\belowfigspace}
\end{figure}

Our objective is to predict values of saturated pixels given an LDR image produced by any type of camera. In order to produce the final HDR image, the predicted pixels are combined with the linearized input image. The final HDR reconstructed pixel $\rhdrp_{i,c}$ with spatial index $i$ and color channel $c$ is computed using a pixel-wise blending with the blend value $\msk_i$,
\begin{equation}
\rhdrp_{i,c} = (1-\msk_i) \cc^{-1}(\ldrp_{i,c}) + \msk_i \exp(\yp_{i,c}),
\label{eqn:masking}
\end{equation}
where $\ldrp_{i,c}$ is the input LDR image pixel and $\yp_{i,c}$ is the CNN output (in the log domain). The inverse camera curve $\cc^{-1}$ is used to transform the input to the linear domain. The blending is a linear ramp starting from pixel values at a threshold $\tau$, and ending at the maximum pixel value, 
\begin{equation}
\msk_i = \frac{\max(0, \max_c(\ldrp_{i,c})-\tau)}{1-\tau}.
\label{eqn:mask}
\end{equation}
 In all examples we use $\tau = 0.95$, where the input is defined to be in the range $[0,1]$. The linear blending prevents banding artifacts between predicted highlights and their surroundings, as compared to a binary mask. It is also used to define the loss function in the training, as described in \secref{loss}. For an illustration of the components of the blending, see \figref{msk}. Due to the blending predictions are focused on reconstructing around the saturated areas, and artifacts may appear in other image regions (\figref{msk}(b)).

\begin{figure*}
	\centering
	\includegraphics[width=0.9\linewidth]{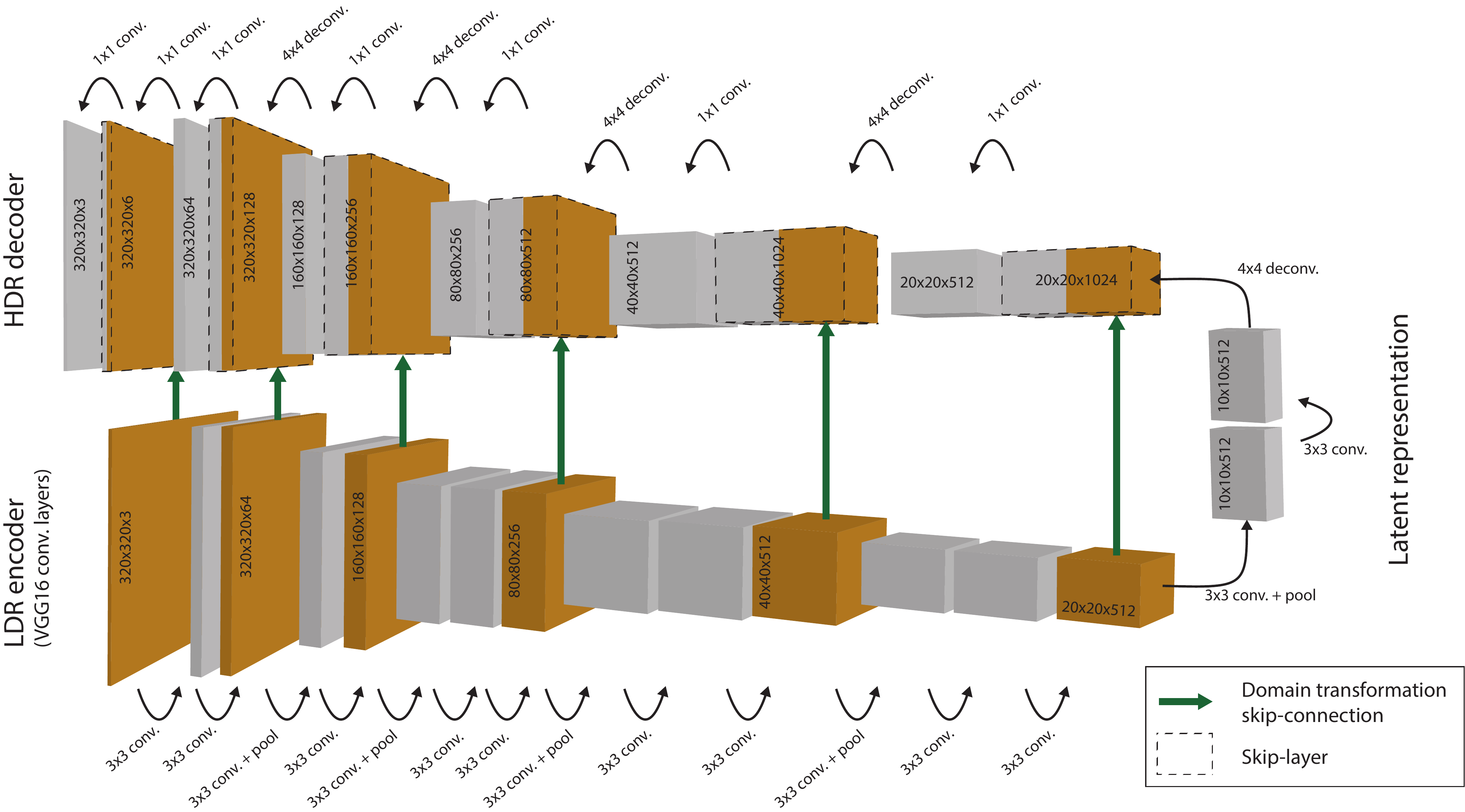}
	\vspace{-1mm}
	\caption{\label{fig:network} Fully convolutional deep hybrid dynamic range autoencoder network, used for HDR reconstruction. The encoder converts an LDR input to a latent feature representation, and the decoder reconstructs this into an HDR image in the log domain. The skip-connections include a domain transformation from LDR display values to logarithmic HDR, and the fusion of the skip-layers is initialized to perform an addition. The network is pre-trained on a subset of the Places database, and deconvolutions are initialized to perform bilinear upsampling. While the specified spatial resolutions are given for a $320\times320$ pixels input image, which is used in the training, the network is not restricted to a fixed image size.}
	\vspace{\belowfigspace}
\end{figure*}

The blending means that the input image is kept unmodified in the non-saturated regions, and linearization has to be made from either knowledge of the specific camera used or by assuming a certain camera curve $\cc$. We do not attempt to perform linearization or color correction with the CNN. Furthermore, information lost due to quantization is not recovered. We consider these problems separate for the following reasons:
\begin{enumerate}
	\item {\bf Linearization:} The most general approach would be to linearize either within the network or by learning the weights of a parametric camera curve. We experimented with both these approaches, but found them to be too problematic given any input image. Many images contain too little information in order to evaluate an accurate camera curve, resulting in high variance in the estimation. On average a carefully chosen assumed transformation performs better. 
	\item {\bf Color correction:} The same reasoning applies to color correction. Also, this would require all training data to be properly color graded, which is not the case. This means that given a certain white balancing transformation of the input, the saturated regions are predicted within this transformed color space.
	\item {\bf Quantization recovery:} Information lost due to quantization can potentially be reconstructed from a CNN. However, this problem is more closely related to super-resolution and compression artifact reduction, for which deep learning techniques have been successfully applied \cite{Dong2015,Ledig2016,Svoboda2016}. Furthermore, a number of filtering techniques can reduce banding artifacts due to quantization \cite{Daly2004,Bhagavathy2007}.
\end{enumerate}

\noindent Although we only consider the problem of reconstructing saturated image regions, we argue that this is the far most important part when transforming LDR images to HDR, and that it can be used to cover a wide range of situations. Typical camera sensors can capture between 8 and 12 stops of dynamic range, which is often sufficient to register all textured areas. However, many scenes contain a small number of pixels that are very bright and thus saturated. These can be reconstructed with the proposed method, instead of capturing multiple exposures or using dedicated HDR cameras. Our method is not intended to recover the lower end of the dynamic range, which is below the noise floor of a sensor. Instead, the problem of under-exposed areas is best addressed by increasing exposure time or gain (ISO). This will result in more saturated pixels, which then can be recovered using our approach.

\subsection{Hybrid dynamic range autoencoder}\label{sec:network}
Autoencoder architectures transform the input to a low-dimensional latent representation, and a decoder is trained to reconstruct the full-dimensional data \cite{hinton2006reducing}. A denoising autoencoder is trained with a corrupted input, with the objective of reconstructing the original uncorrupted data \cite{Vincent2008}. This is achieved by mapping to a higher level representation that is invariant to the specific corruption. We use the same concept for reconstruction of HDR images. In this case the corruption is clipped highlights, and the encoder maps the LDR to a representation that can be used by the decoder for HDR reconstruction. This means that the encoder and decoder work in different domains of pixel values, and we design them to optimally account for this. Since our objective is to reconstruct larger images than is practical to use in training, the latent representation is not a fully connected layer, but a low-resolution multi-channel image. Such a fully convolutional network (FCN) enables predictions at any resolution that is a multiple of the autoencoder downscaling factor.

The complete autoencoder design is depicted in \figref{network}. Convolutional layers followed by max-pooling encodes the input LDR in a $\frac{W}{32} \times \frac{H}{32} \times 512$ latent image representation, where $W$ and $H$ are the image width and height, respectively. The encoder layers correspond to the well-known VGG16 network \cite{Simonyan2014}, but without the fully connected layers.

While the encoder operates directly on the LDR input image, the decoder is responsible for producing HDR data.
For this reason the decoder operates in the log domain. This is accomplished using a loss function that compares the output of the network to the log of the ground truth HDR image, as explained in \secref{loss}. For the image upsampling, we use deconvolutional layers with a spatial resolution of $4\times4$ initialized to perform  bilinear upsampling \cite{Long2015}. While nearest neighbor up-sampling followed by convolution has been shown to alleviate artifacts that are common in decoder deconvolutions \cite{Odena2016}, we have not experienced such problems, and instead use the more general deconvolutional layers.
All layers of the network use ReLU activation functions, and after each layer of the decoder a batch normalization layer~\cite{ioffe2015batch} is used.

\subsection{Domain transformation and skip-connections}
The encoding of the input image means that much of the high resolution information in earlier layers of the encoder are lost.
The information could potentially be used by the decoder to aid reconstruction of high frequency details in saturated regions.
Thus, we introduce skip-connections that transfer data between both high and low level features in the encoder and decoder. 

Skip-connections have been shown to be useful for constructing deeper network architectures which improve performance in a variety of tasks~\cite{He2016}. For autoencoders, where layers have different spatial resolution, a separate residual stream can be maintained in full resolution, with connections to each layer within the autoencoder \cite{Pohlen2016}. Alternatively, skip-connections between layers of equal resolution in encoder and decoder have also been shown to boost performance in a variety of imaging tasks using autoencoders \cite{Ronneberger2015,Zhang2017}.

\begin{figure}[t]
	\vspace{5pt}
	\newcommand\ww{0.116}
	\centering
	\includegraphics[width=\ww\textwidth]{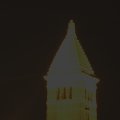}
	\includegraphics[width=\ww\textwidth]{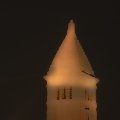}
	\includegraphics[width=\ww\textwidth]{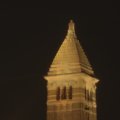}
	\includegraphics[width=\ww\textwidth]{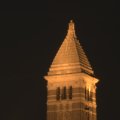}\\
	\vspace{-2pt}
	\subfigure[Input]{\includegraphics[width=\ww\textwidth]{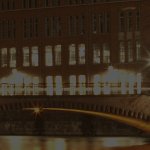}}
	\subfigure[Without skip]{\includegraphics[width=\ww\textwidth]{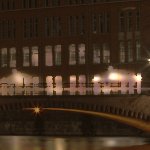}}
	\subfigure[With skip]{\includegraphics[width=\ww\textwidth]{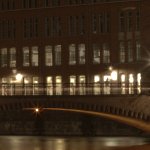}}
	\subfigure[Ground truth]{\includegraphics[width=\ww\textwidth]{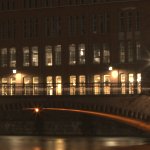}}
	\vspace{-10pt}
	\caption{\label{fig:skip} Zoom-ins of reconstruction without (b) and with (c) the domain transformation skip-connections. The plain autoencoder architecture can reconstruct high luminance, but without skip-connections the detail information around saturated regions cannot be fully exploited.}
	\vspace{\belowfigspace}
\end{figure}

Our autoencoder uses skip-connections to transfer each level of the encoder to the corresponding level on the decoder side. Since the encoder and decoder process different types of data (see \secref{network}), the connections include a domain transformation described by an inverse camera curve and a log transformation, mapping LDR display values to a logarithmic HDR representation. Since the camera curve is unknown, we have to assume its shape. Although a sigmoid function fits well with camera curves in general \cite{Grossberg2003}, its inverse is not continuous over $\Real^+$. The linearization of the skip-connections is therefore done using a gamma function $\cc^{-1}(x) = x^\gamma$, where $\gamma = 2$.

A skip-connected layer is typically added to the output of the layer at the connection point. However, to allow for additional freedom, we concatenate the two layers along the feature dimension. That is, given two $W\times H\times K$ dimensional layers, the concatenated layer is $W\times H\times 2K$. The decoder then makes a linear combination of these, that reduces the number of features back to $K$. This is equivalent to using a convolutional layer with a filter size of $1 \times 1$, where the number of input and output channels are $2K$ and $K$, respectively, as depicted in \figref{network}. More specifically, the complete LDR to HDR skip connection is defined as
\begin{equation}
\tilde{\feat}_i^D = \sigma \left(\vect{W} 
\begin{bmatrix}
\feat_i^D \\  \log \left( \cc^{-1}\left( \feat_i^E \right) + \epsilon \right)
\end{bmatrix}
+ \vect{b} \right).
\end{equation}
The vectors $\feat_i^E$ and $\feat_i^D$ denote the slices across all the feature channels $k \in \{1,...,K\}$ of the encoder and decoder layer tensors $\vect{y}^E, \vect{y}^D \in \Real^{W\times H\times K}$, for one specific pixel $i$. 
Furthermore, $\tilde{\feat}_i^D$ is the decoder feature vector with information fused from the skip-connected vector $\feat_i^E$.
$\vect{b}$ is the bias of the feature fusion, and $\sigma$ is the activation function, in our case the rectified linear unit (ReLU). A small constant $\epsilon$ is used in the domain transformation in order to avoid zero values in the log transform. Given $K$ features, $\feat^E$ and $\feat^D$ are $1\times K$ vectors, and $\vect{W}$ is a $2K \times K$ weight matrix, which maps the $2K$ concatenated features to $K$ dimensions. This is initialized to perform an addition of encoder and decoder features, setting the weights as
\begin{equation}
\vect{W}_0 =  
\begin{bmatrix}
1 & 0 & \dots & 0 & 1 & 0 & \dots & 0 \\ 
0 & 1 &  \dots & 0 & 0 & 1 &  \dots & 0 \\ 
\vdots & \vdots & \ddots & \vdots & \vdots & \vdots & \ddots & \vdots \\
0 & 0 &  \dots & 1 & 0 & 0 &  \dots & 1
\end{bmatrix}, \;\;\;\; \vect{b}_0 = \vect{0}.
\label{eqn:skip_weights}
\end{equation}
During training, these weights can be optimized to improve the performance of the skip-connection. 
Since the linear combination of features is performed in the log domain, it corresponds to multiplications of linear HDR data. This is an important characteristic of the domain transformation skip-connections as compared to existing skip architectures.

An example of the impact of the described skip-connection architecture is shown in \figref{skip}. The autoencoder design is able to reconstruct HDR information from an encoded LDR image. However, all information needed by the decoder has to travel trough the intermediate encoded representation. Adding the skip-connections enables a more optimal use of existing details.

\subsection{HDR loss function}\label{sec:loss}
A cost function formulated directly on linear HDR values will be heavily influenced by high luminance values, leading to underestimation of important differences in the lower range of luminaces. The few existing deep learning systems that predict HDR have treated this problem by defining the objective function in terms of tone-mapped luminance \cite{Zhang2017,Kalantari2017}. In our system the HDR decoder is instead designed to operate in the log domain. Thus, the loss is formulated directly on logarithmic HDR values, given the predicted log HDR image $\y$ and the linear ground truth $\hdr$,
\begin{equation}
\loss(\y,\hdr) = \frac{1}{3N}\sum_{i,c}\left| \msk_i \left(\yp_{i,c} - \log \left(\hdrp_{i,c} + \epsilon\right) \right) \right|^2,
\label{eqn:mse_loss}
\end{equation}
where $N$ is the number of pixels.
Since $\hdrp_{i,c} \in \Real^+$, the small constant $\epsilon$ removes the singularity at zero pixel values. The cost formulation is perceptually motivated by the the close to logarithmic response of the human visual system (HVS) in large areas of the luminance range, according to the Weber-Fechner law \cite{Fechner1965}. The law implies a logarithmic relationship between physical luminance and the perceived brightness. Thus, a loss formulated in the log domain makes perceived errors spread approximately uniformly across the luminance range.

As described in \secref{problem}, we use only the information from the predicted HDR image $\y$ around saturated areas. This is also reflected by the loss function in \eqnref{mse_loss} where the blend map $\vect{\msk}$ from \eqnref{mask} is used to spatially weight the error.

\begin{figure}[t]
	\vspace{5pt}
	\newcommand\ww{0.116}
	\centering
	\includegraphics[width=\ww\textwidth]{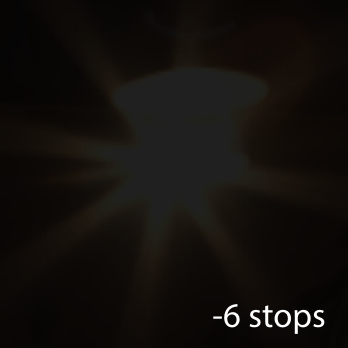}
	\includegraphics[width=\ww\textwidth]{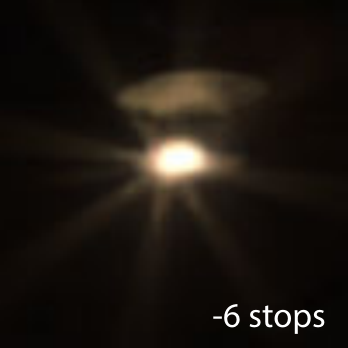}
	\includegraphics[width=\ww\textwidth]{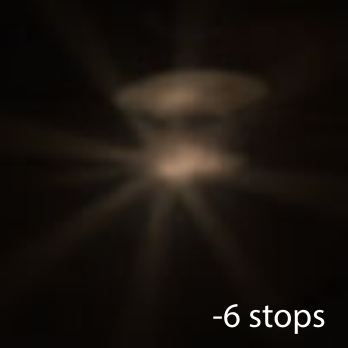}
	\includegraphics[width=\ww\textwidth]{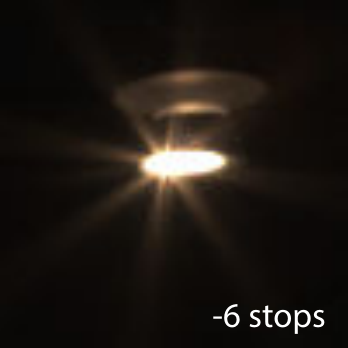}\\
	\vspace{-2pt}
	\subfigure[Input]{\includegraphics[width=\ww\textwidth]{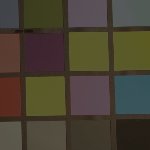}}
	\subfigure[$\lambda=0.9$]{\includegraphics[width=\ww\textwidth]{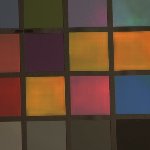}}
	\subfigure[$\lambda=0.05$]{\includegraphics[width=\ww\textwidth]{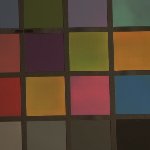}}
	\subfigure[Ground truth]{\includegraphics[width=\ww\textwidth]{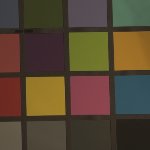}}
	\vspace{-10pt}
	\caption{\label{fig:lambda} Zoom-ins of reconstructions with different relative weight of illuminance and reflectance, $\lambda$ in \eqnref{ir_loss}. A higher weight of illuminance will in general better predict high intensity regions (b), while a higher weight of reflectance is better at deducing local colors and details (c).}
	\vspace{\belowfigspace}
\end{figure}

Treating the illuminance and reflectance components separately makes sense from a perceptual standpoint, as the visual system may indirectly perform such separation when inferring reflectance or discounting illumination \cite{Gilchrist1984}. 
We therefore also propose another, more flexible loss function that treats illuminance and reflectance separately.
The illumination component $\ill$ describes the global variations, and is responsible for the high dynamic range. The reflectance $\refl$ stores information about details and colors. This is of lower dynamic range and modulates the illuminance to create the final HDR image, $\hdrp_{i,c} = \illp_i \reflp_{i,c}$. We approximate the log illuminance by means of a Gaussian low-pass filter $G_\sigma$ on the log luminance $\vect{L}^{\y}$,
\begin{equation}
\begin{split}
\log\left( \illp_i^{\y} \right) &= \left( G_\sigma \ast \vect{L}^{\y} \right)_i, \\
\log\left( \reflp^{\y}_{i,c} \right) &= \yp_{i,c} - \log\left( \illp_i^{\y} \right).
\end{split}
\end{equation}
Since the estimation is performed in the log domain, the log reflectance is the difference between $\y$ and log illuminance.
$\vect{L}^{\y}$ is a linear combination of the color channels, $L^{\y}_i = \log(\sum_c w_c \exp(\yp_{i,c}))$, where $w = \{0.213, 0.715, 0.072\}$. The standard deviation of the Gaussian filter is set to $\sigma = 2$. The resulting loss function using $\ill$ and $\refl$ is defined as
\begin{equation}
\begin{split}
\loss_{IR}(\y,\hdr) = \frac{\lambda}{N} & \sum_i \left| \msk_i \left(\log \left(\illp_i^{\y}\right) - \log \left(\illp_i^{\vect{y}}\right) \right) \right|^2 \\
+ \frac{1-\lambda}{3N} &\sum_{i,c} \left| \msk_i \left(\log \left(\reflp_{i,c}^{\y}\right) - \log \left(\reflp_{i,c}^{\vect{y}}\right) \right) \right|^2,
\end{split}
\label{eqn:ir_loss}
\end{equation}
where $\vect{y} = \log (\hdr + \epsilon)$ to simplify notation. The user-specified parameter $\lambda$ can be tuned for assigning different importance to the illuminance and reflectance components. If not stated otherwise, we use the illuminance + reflectance (I/R) loss with $\lambda = 0.5$ for all results in this paper. This puts more importance on the illuminance since the error in this component generally is larger. 
\figref{lambda} shows examples of predictions where the optimization has been performed with different values of $\lambda$. With more relative weight on illuminance, high intensity areas are in general better predicted, which \eg could benefit IBL applications. If the reflectance is given more importance, local colors and details can be recovered with higher robustness, for better quality \eg in post-processing applications.

\begin{figure}[t]
	\newcommand\ww{0.116}
	\centering
	\subfigure[Input]{\includegraphics[width=\ww\textwidth]{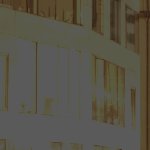}}
	\subfigure[Direct loss (eq. \ref{eqn:mse_loss})]{\includegraphics[width=\ww\textwidth]{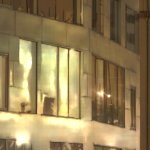}}
	\subfigure[I/R loss (eq. \ref{eqn:ir_loss})]{\includegraphics[width=\ww\textwidth]{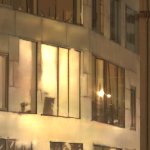}}
	\subfigure[Ground truth]{\includegraphics[width=\ww\textwidth]{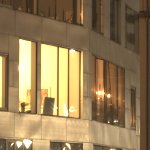}}
	\vspace{-10pt}
	\caption{\label{fig:loss} Zoom-in of a reconstruction with different loss functions. The input (a) is exposure corrected and clipped to have a large amount of information lost. The direct pixel loss (b) is more prone to generating artifacts as compared to the illuminance + reflectance loss (c).}
	\vspace{\belowfigspace}
\end{figure}

The visual improvements from using the I/R loss compared to the direct loss in \eqnref{mse_loss} are subtle. However, in general it tends to produce less artifacts in large saturated areas, as exemplified in \figref{loss}. One possible explanation is that the Gaussian low-pass filter in the loss function could have a regularizing effect, since it makes the loss in a pixel influenced by its neighborhood. This observation is further supported by the comparison in \tabref{error}, where the I/R loss lowers the final error in \eqnref{mse_loss} by more than 5\%, demonstrating better generalization to the test data.


\section{HDR image dataset}\label{sec:hdr_db}
\begin{figure}
	\vspace{-2pt}
	\centering
	\includegraphics[width=0.95\linewidth]{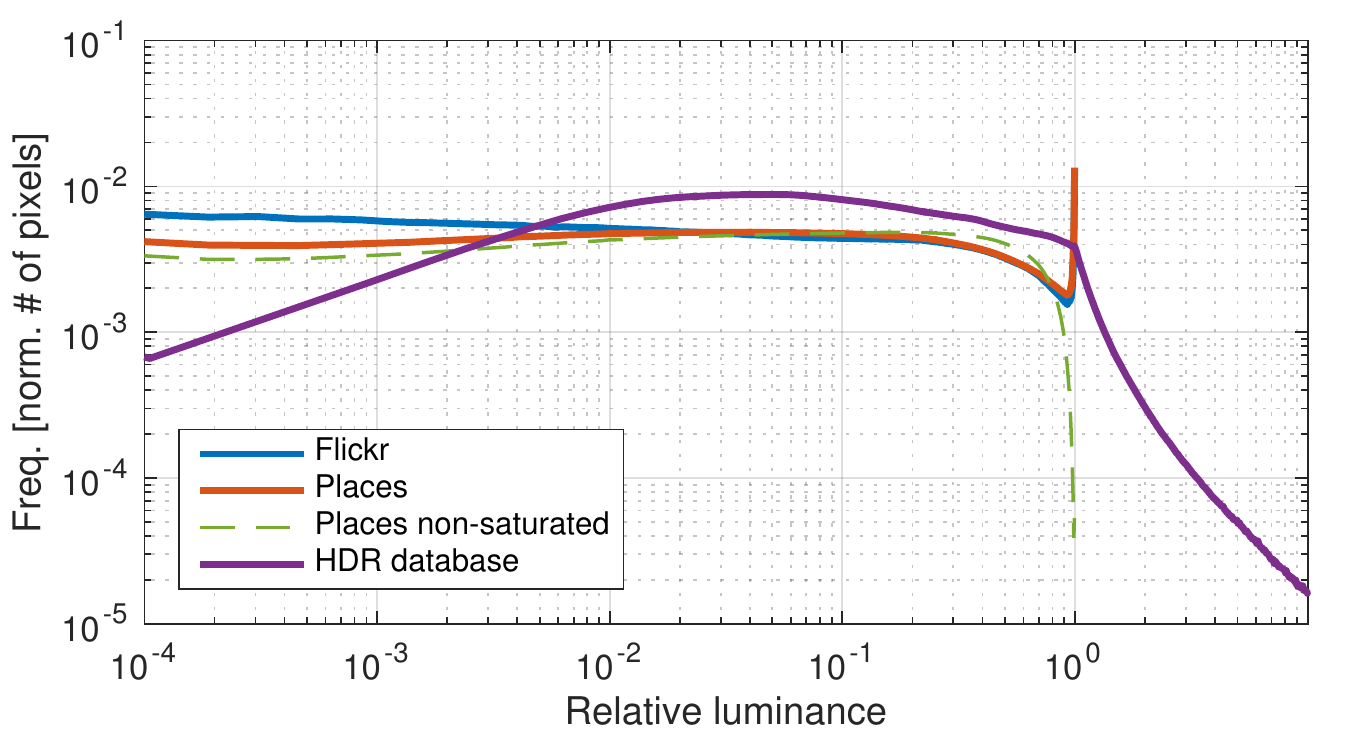}
	\vspace{-10pt}
	\caption{\label{fig:hist} Histograms over two LDR datasets and the pre-processed HDR dataset. The creation of the HDR data is described in \secref{hdr_db}. For the LDR data, the probabilities show a large increase close to 1, indicating saturated information. The HDR dataset contains such information, represented by the tail of decreasing frequency.} 
	\vspace{\belowfigspace}
\end{figure}

A key challenge for a learning-based HDR reconstruction is to obtain a sufficiently large set of well structured training data.  However, an increasing amount of HDR content has become available, in particular through recent HDR video datasets \cite{Froehlich2014,Kronander2014,Azimi2014,Boitard2014}. We were able to gather a total of 1121 HDR images and 67 HDR video sequences. The sources of the data are specified in the supplementary document. 4 video sequences and 95 images are separated from these to create the test set used throughout this paper, and the rest are used for training. Since consecutive frames from a video sequence are expected to be very similar, we use every 10th frame. Together with the static images, this results in a total of $\sim\!\!3700$ HDR images. Using a virtual camera, a carefully designed set of data augmentation operations are then applied in order to improve robustness of the predictions.

Considering each HDR image a real-world scene, we set up a virtual camera that captures a number of random regions of the scene using a randomly selected camera calibration. This provides us with an augmented set of LDR and corresponding HDR images that are used as input and ground truth for training, respectively. The regions are selected as image crops with random size and position, followed by random flipping and resampling to $320 \times 320$ pixels. The camera calibration incorporates parameters for exposure, camera curve, white balance and noise level. These are randomly selected, with the camera curve defined as a parametric function fitted to the database of camera curves collected by Grossberg and Nayar \citeyear{Grossberg2003}. For details on the augmentation, we refer to Appendix \ref{app:augmentation}.

In total we capture  $\sim\!\!125$K training samples from the HDR dataset using the virtual camera. This augmentation is responsible for creating a final trained model that generalizes well to a wide range of images captured with different cameras.

\customsection{Image statistics} It is important to note that the dynamic range statistics of LDR and HDR images differ considerably. \figref{hist} shows averaged histograms over two typical LDR datasets, as well as our HDR dataset of $125$K images. The LDR data are composed of around $2.5$M and $200$K images for Places \cite{Zhou2014} and Flickr, respectively. Inspecting the LDR histograms, they show a relatively uniform distribution of pixel values, except for distinct peaks near the maximum value representing information lost due to saturation. In the HDR histogram on the other hand, pixels are not saturated, and are instead represented by an exponentially decaying long tail. Although there are not many pixels with extreme intensities, these are very important to properly learn a plausible HDR image.

\begin{figure}[t]
	\newcommand\ww{0.116}
	\centering
	\subfigure[Input]{\includegraphics[width=\ww\textwidth]{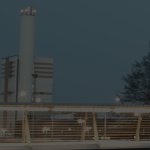}}
	\subfigure[No pre-training]{\includegraphics[width=\ww\textwidth]{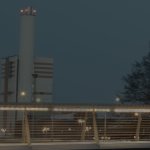}}
	\subfigure[Pre-trained]{\includegraphics[width=\ww\textwidth]{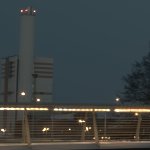}}
	\subfigure[Ground truth]{\includegraphics[width=\ww\textwidth]{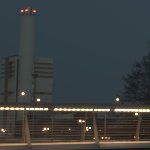}}
	\vspace{-10pt}
	\caption{\label{fig:init} A $150\times 150$ pixels zoom-in of a reconstruction. Using pre-training the CNN is in general more consistent and can reconstruct smaller highlights better.}
	\vspace{\belowfigspace}
\end{figure}

\section{Training}\label{sec:training}
To initialize the weights in the network we use different strategies for different parts of the network.
As we use the convolutional layers from the well-known VGG16 network, we can use pre-trained weights available for large scale image classification on the Places database \cite{Zhou2014} to initialize the encoder. The decoder deconvolutions are initiated for bilinear upsampling, and the fusions of skip-connection layers are initiated to perform addition of features (\eqnref{skip_weights}). For the convolutions within the latent image representation (right-most of \figref{network}) and the final feature reduction (top-left in \figref{network}) we use Xavier initializaion~\cite{Glorot2010}.

Minimization is performed with the ADAM optimizer \cite{Kingma2014}, with a learning rate of $5 \times 10^{-5}$, on the loss function in \eqnref{ir_loss}. In total $800$K steps of back-propagation are performed, with a mini-batch size of 8, taking approximately 6 days on an Nvidia Titan X GPU.

\begin{figure*}
	\vspace{5pt}
	\newcommand\ww{0.241}
	\newcommand\hpad{\hspace{3pt}}
	\newcommand\vpad{\vspace{3pt}}
	\centering
	\rotatebox[origin=l]{90}{\footnotesize \hspace{20pt}(a) Input}\hspace{2pt}\includegraphics[width=\ww\textwidth]{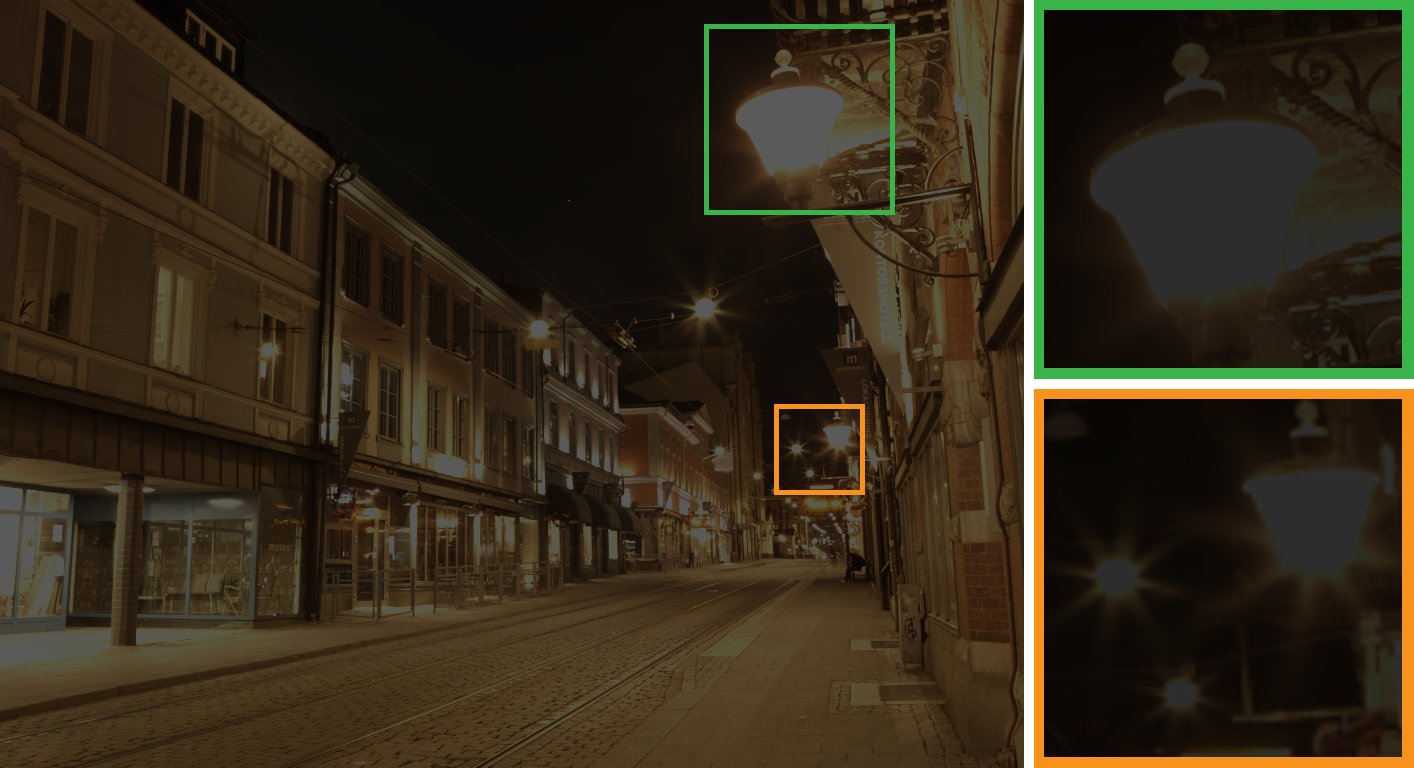}\hpad
	\includegraphics[width=\ww\textwidth]{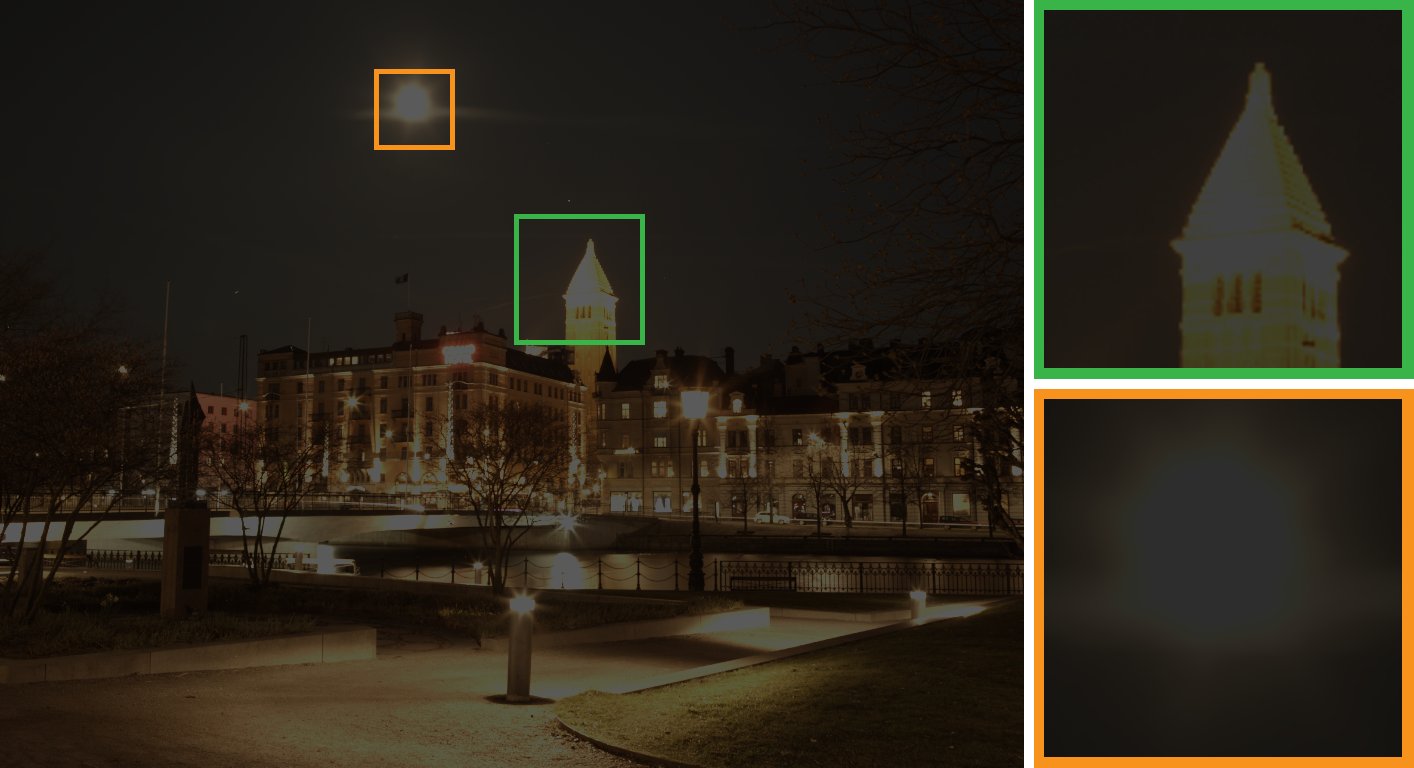}\hpad
	\includegraphics[width=\ww\textwidth]{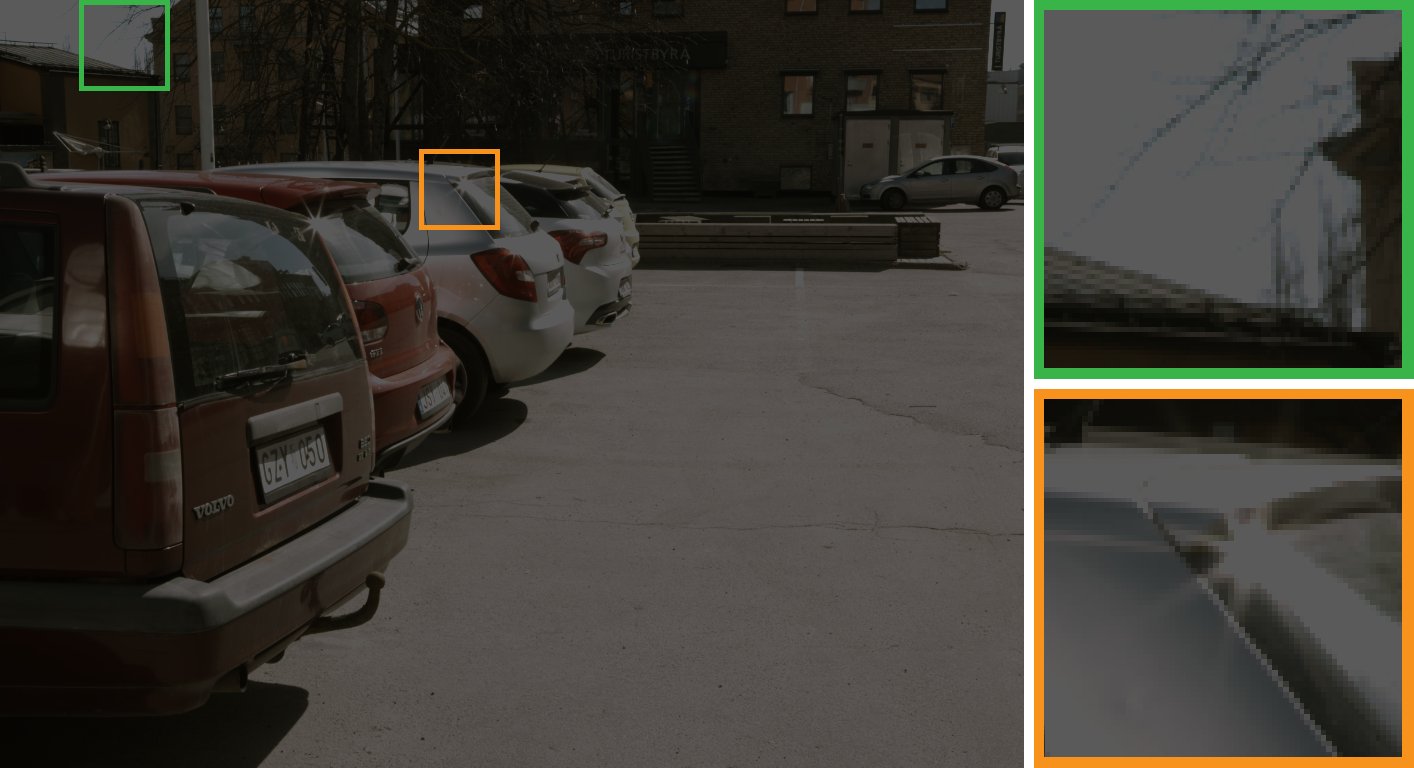}\hpad
	\includegraphics[width=\ww\textwidth]{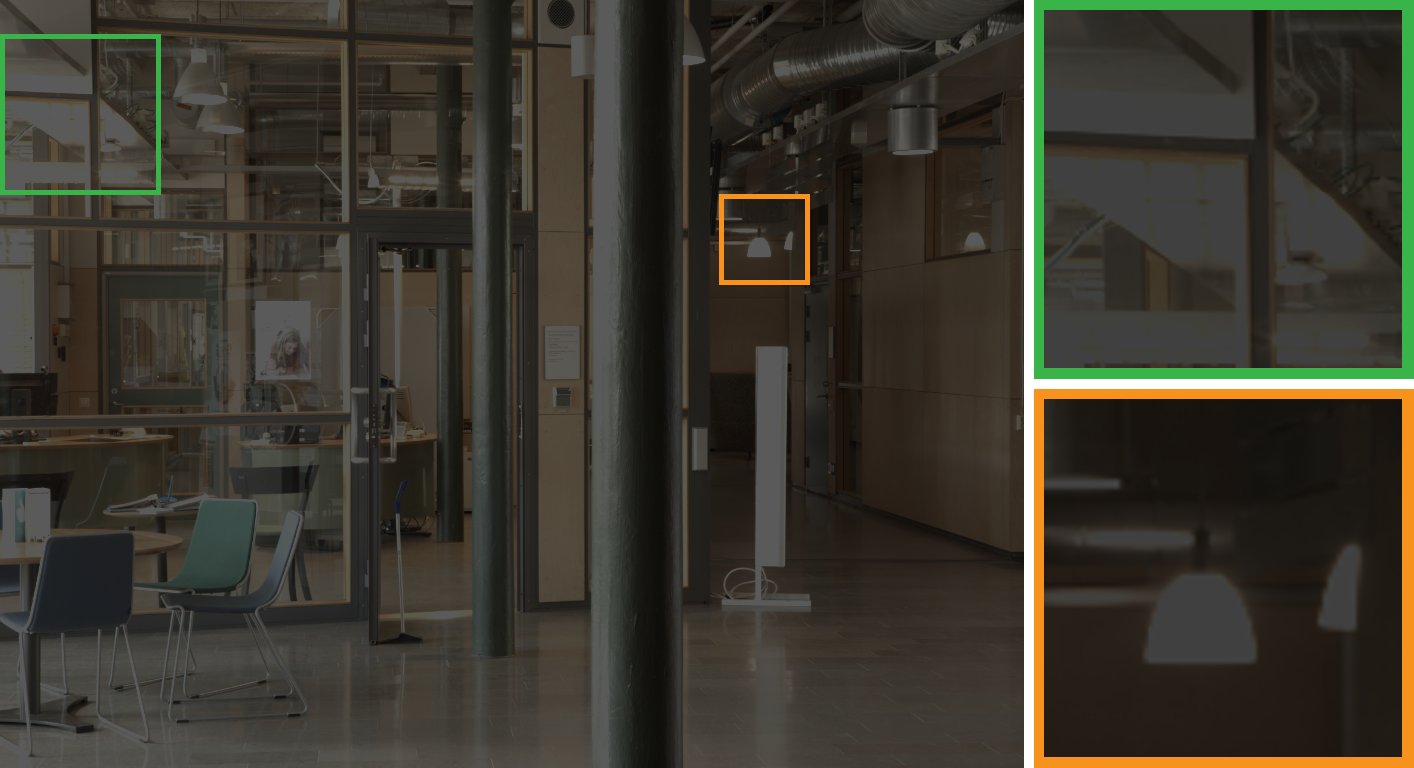}\vpad\\
	\rotatebox[origin=l]{90}{\footnotesize \hspace{4pt}(b) Reconstruction}\hspace{2pt}\includegraphics[width=\ww\textwidth]{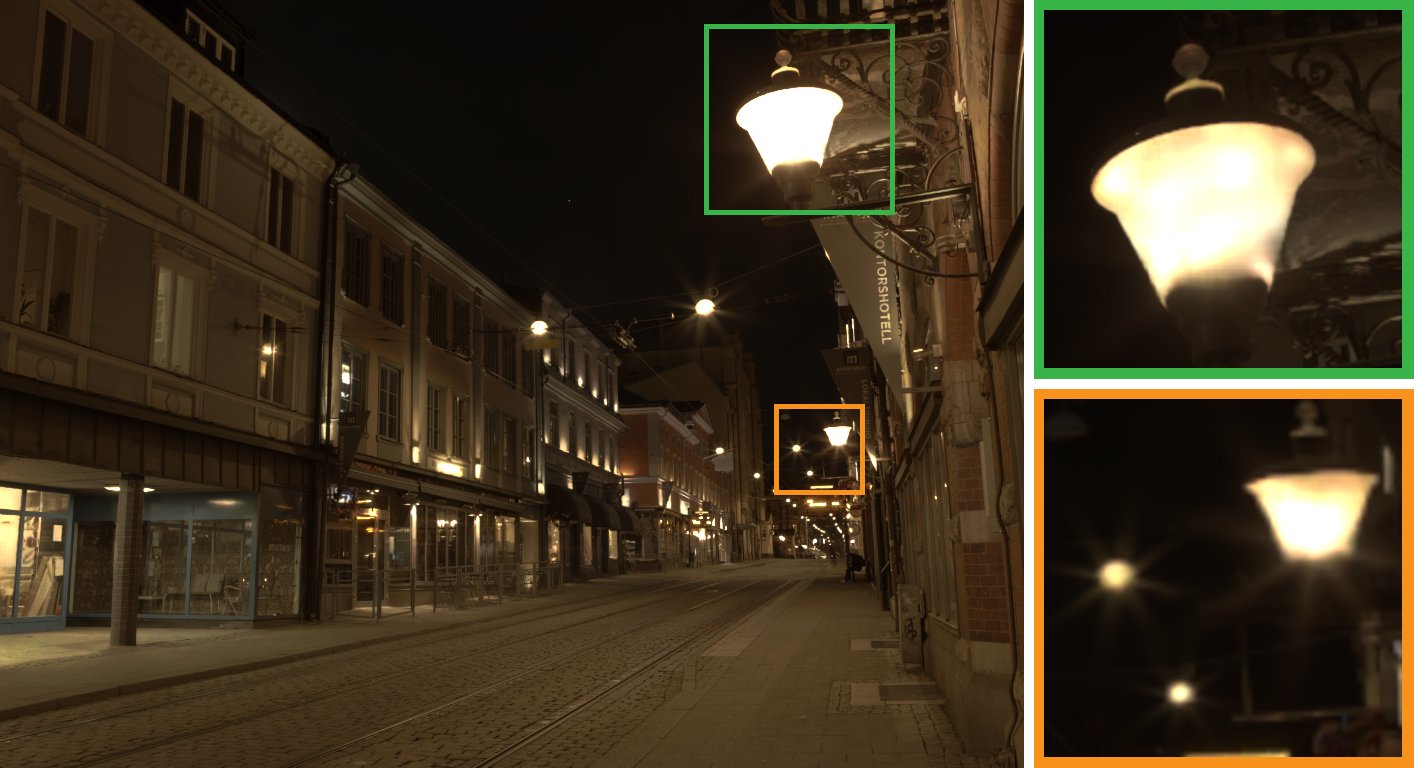}\hpad
	\includegraphics[width=\ww\textwidth]{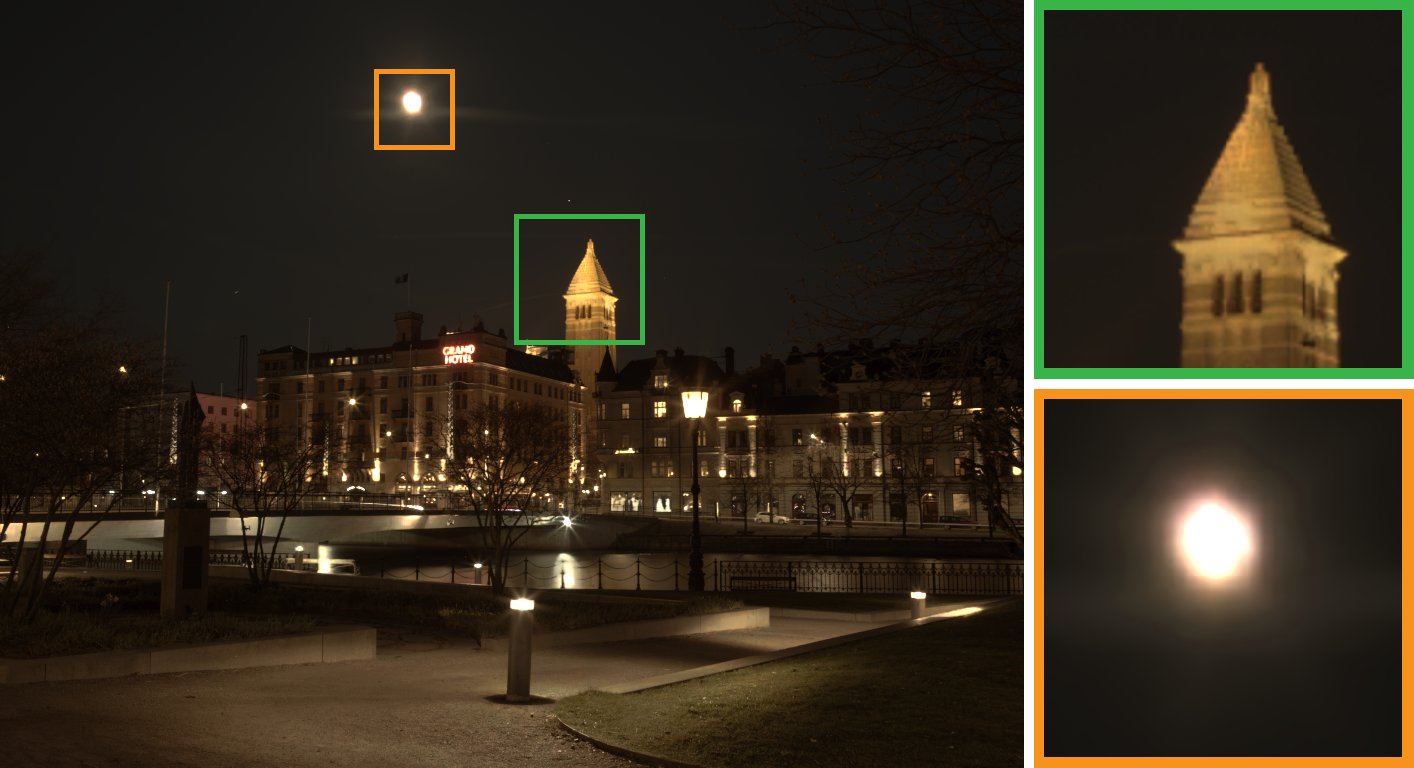}\hpad
	\includegraphics[width=\ww\textwidth]{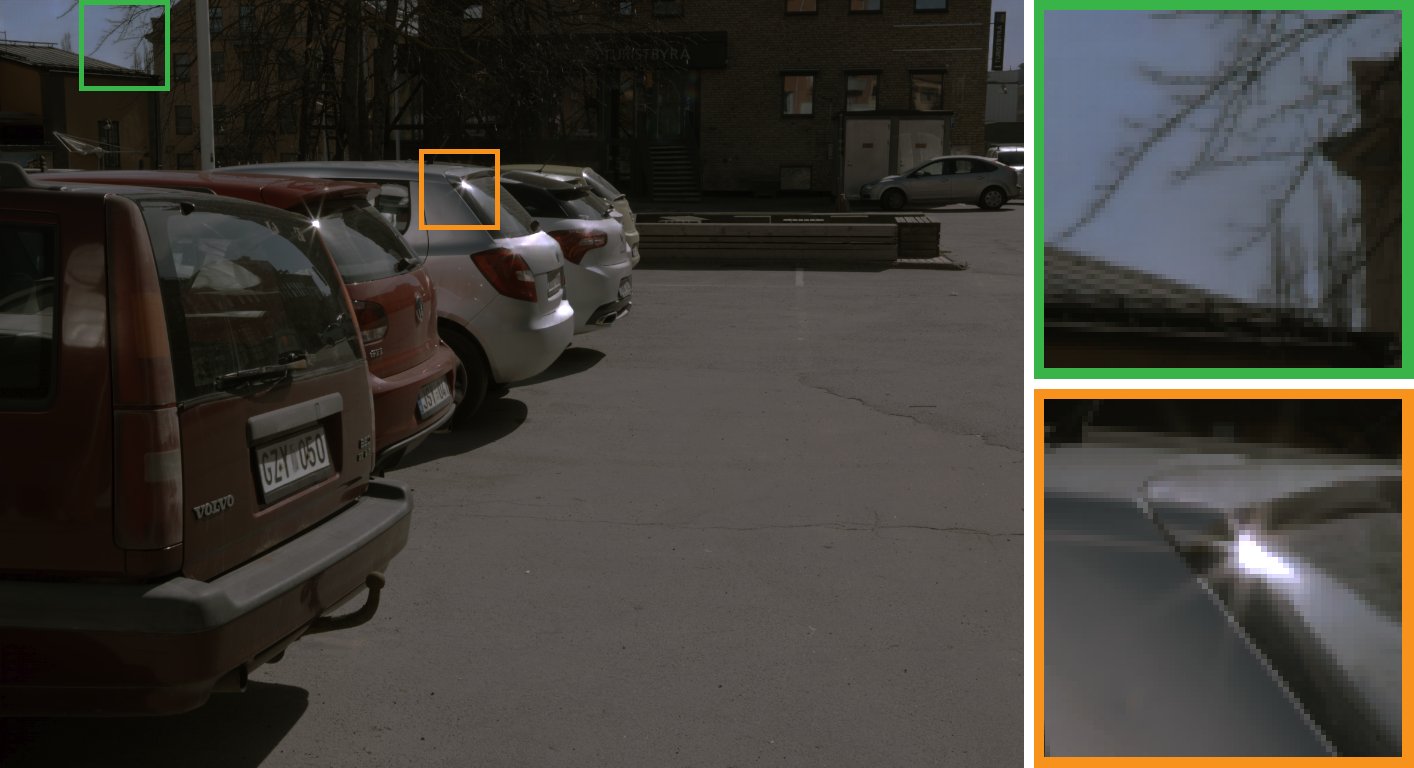}\hpad
	\includegraphics[width=\ww\textwidth]{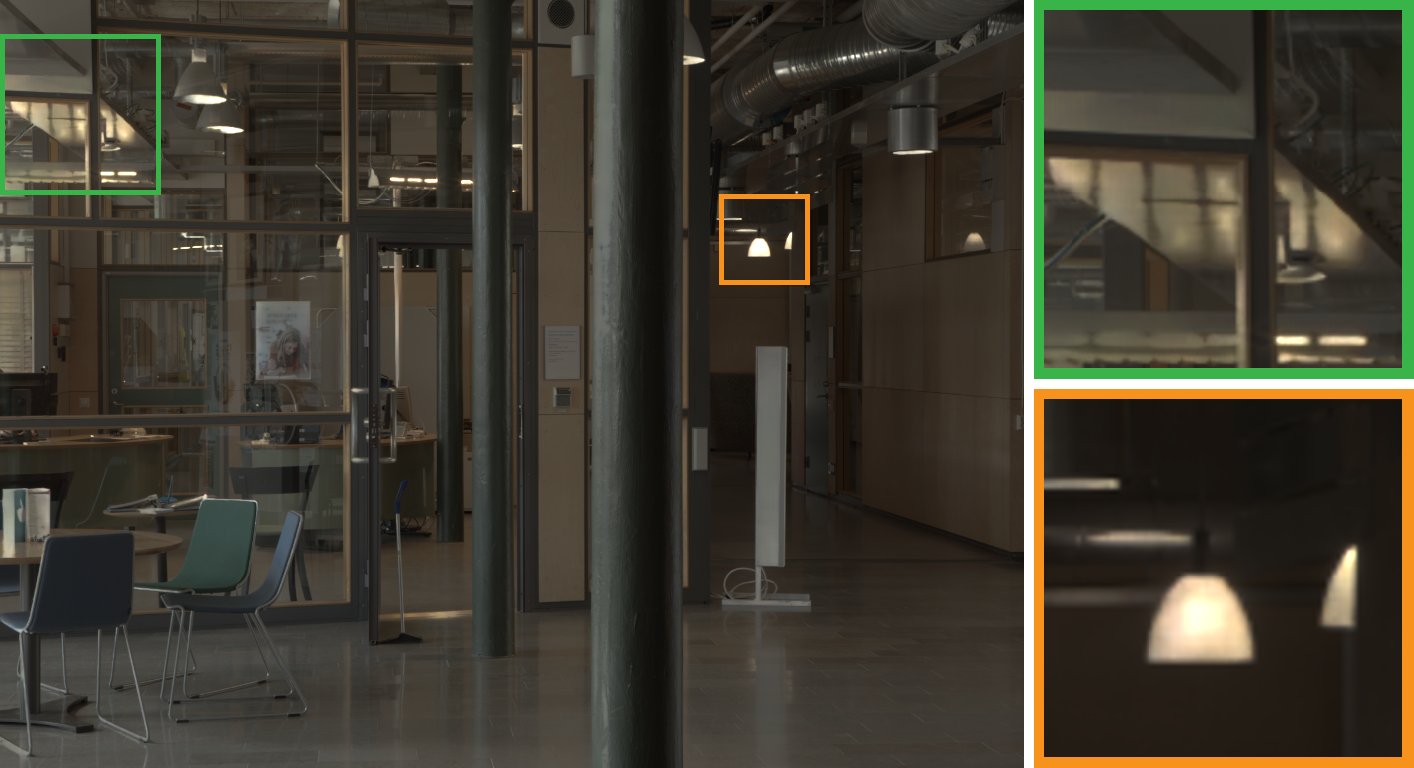}\vpad\\
	\rotatebox[origin=l]{90}{\footnotesize \hspace{8pt}(c) Ground truth}\hspace{2pt}\includegraphics[width=\ww\textwidth]{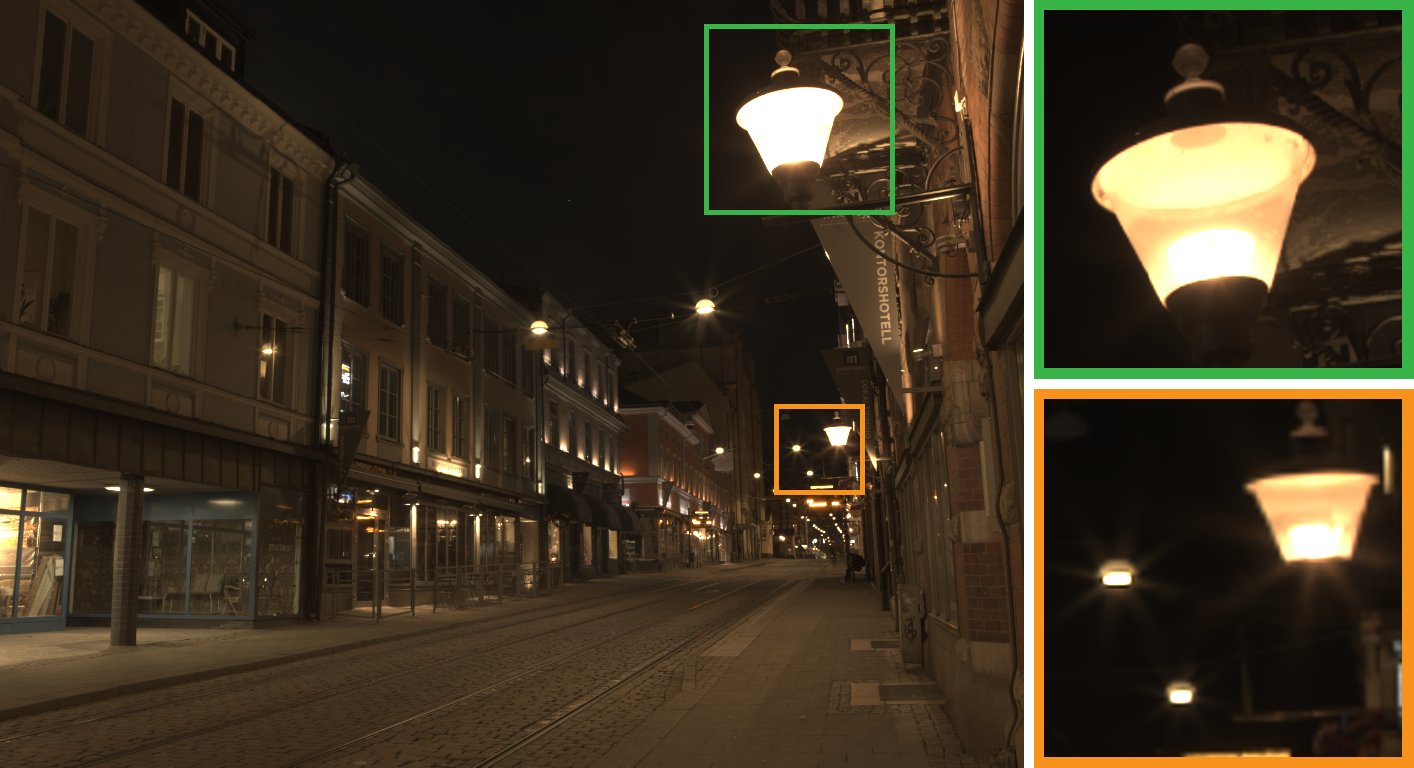}\hpad
	\includegraphics[width=\ww\textwidth]{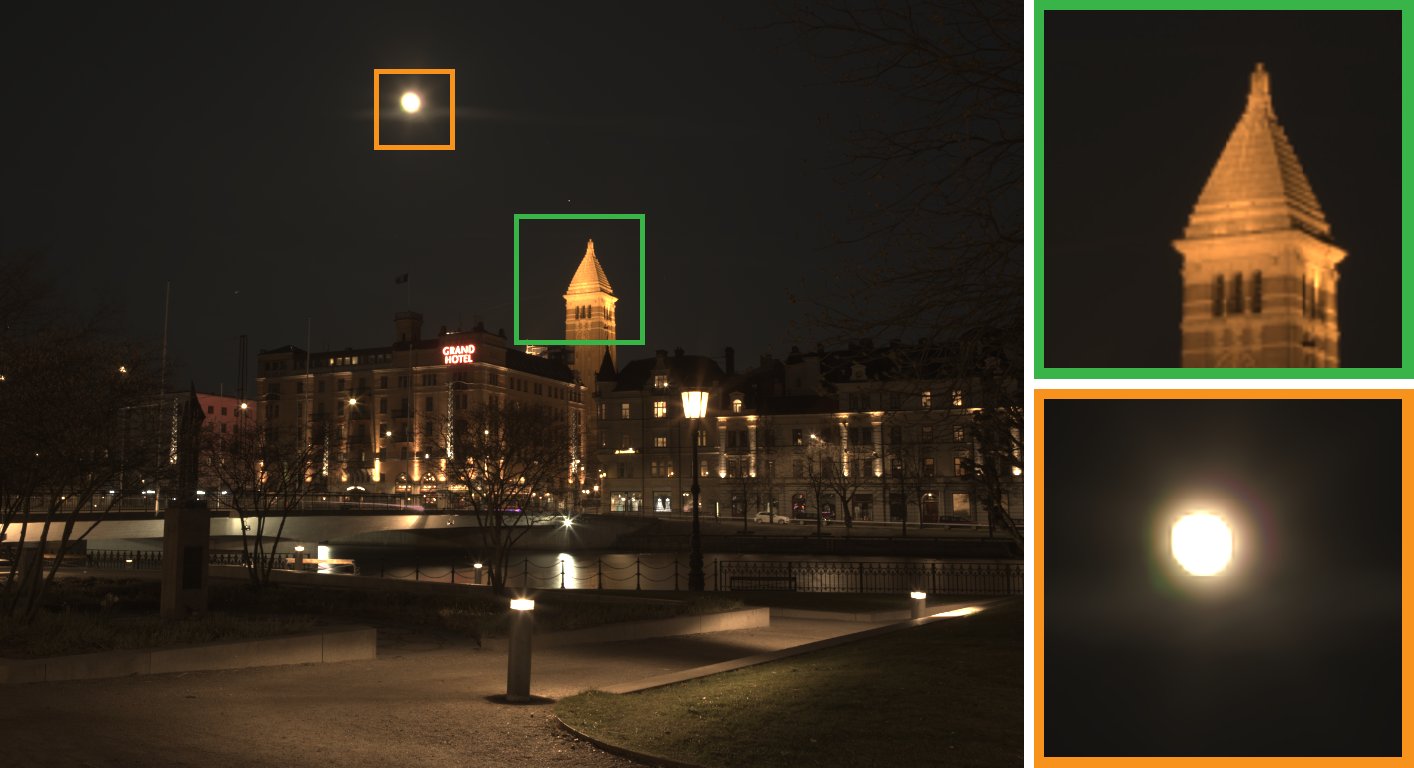}\hpad
	\includegraphics[width=\ww\textwidth]{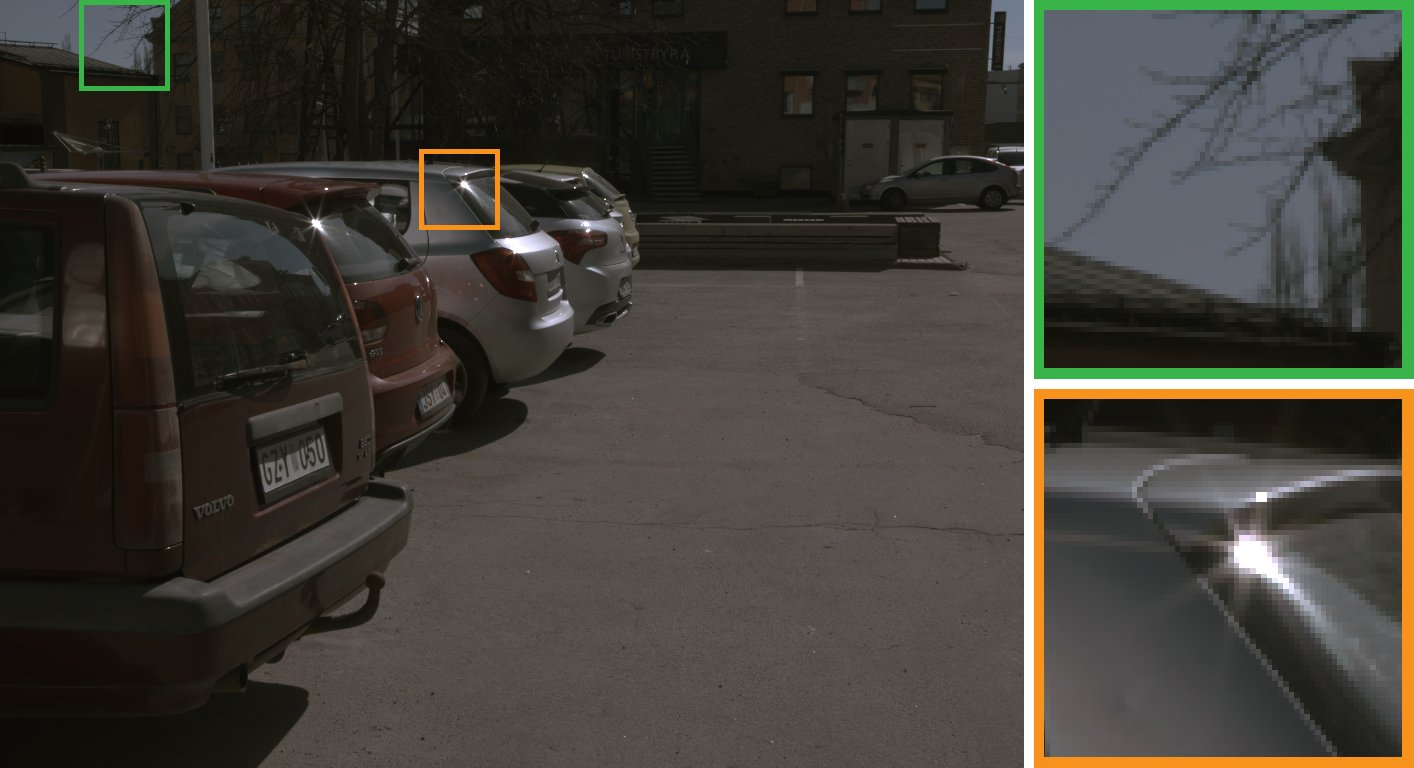}\hpad
	\includegraphics[width=\ww\textwidth]{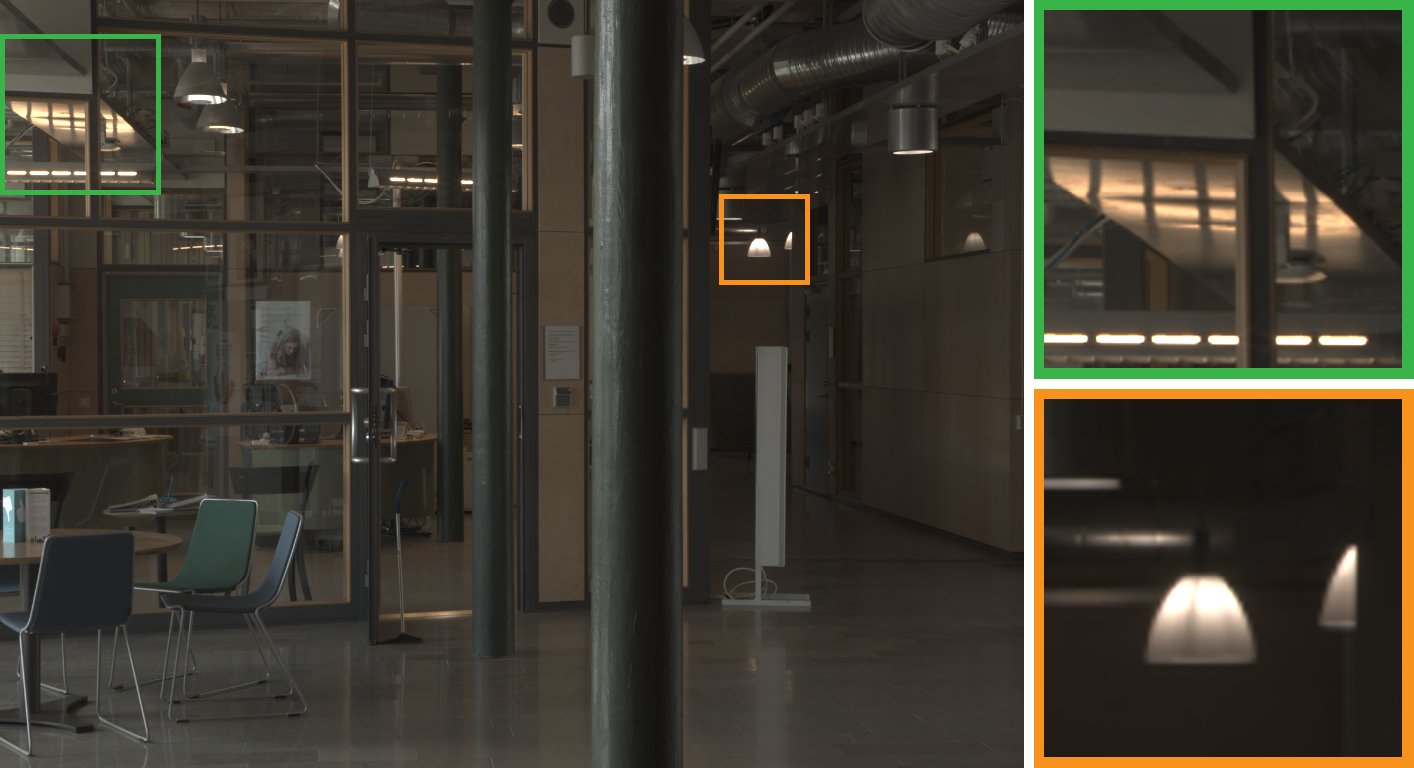}\vpad\\
	\vspace{-2pt}
	\caption{\label{fig:itm_hdr} The input images (a) have been exposure corrected, followed by camera transformation, quantization and clipping. $5\%$ of the pixels are saturated and contain no information. Visually convincing reconstructions (b) can be made in a wide range of situations. The reconstructions correspond well to the ground truth HDR images (c). The exposures of the images have been reduced to show the differences, and all images have been gamma corrected for display.}
\end{figure*}

\subsection{Pre-training on simulated HDR data}\label{sec:pretrain}
As we have a limited amount of HDR data at hand, we use transfer learning by pre-training the entire network on a large simulated HDR dataset. To this end, we select a subset of the images in the Places database~\cite{Zhou2014}, requiring that the images should not contain saturated image regions. Given the set of all Places images $\mathbb{P}$, this subset $\mathbb{S} \subset \mathbb{P} $ is defined as
\begin{equation}
\mathbb{S} = \left\lbrace \ldr \; | \; \ldr \in \mathbb{P}, \; p_{\ldr}(255) < \xi \right\rbrace,
\end{equation}
where $p_{\ldr}$ is the image histogram. For the threshold we use $\xi = 50/256^2$. Thus, if less than $50$ pixels ($0.076\%$ of the $256^2$ pixels of an image) have maximum value, we use this in the training set. For the Places database this gives $\sim\!\!600$K images of the total set size of $\sim\!\!2.5M$. The averaged histogram over the subset $\mathbb{S}$, plotted in \figref{hist}, does not show the peak of saturated pixels as the original set $\mathbb{P}$. By linearizing the images $\ldr \in \mathbb{S}$ using the inverse of the camera curve $\cc$ in \eqnref{cc} and increasing the exposure, $\hdr = s \cc^{-1}(\ldr)$, we create a simulated HDR training dataset.

The simulated HDR dataset is prepared in the same manner as in \secref{hdr_db}, but at $224 \times 224$ pixels resolution and without resampling.
The CNN is trained using the ADAM optimizer with learning rate $2\times 10^{-5}$ for $3.2$M steps with a batch size of 4.

The result of this pre-training on synthetic data leads to a significant improvement in performance. Small highlights, which sometimes are underestimated, are better recovered, as illustrated in \figref{init}, and less artifacts are introduced in larger saturated regions. \tabref{error} shows that the error is reduced by more than $10\%$.


\begin{table}[b]
	\newcommand\crot{0}
	\def\arraystretch{1.4}
	\setlength\tabcolsep{0.3cm}
	\centering
	\caption{Different MSEs evaluated over the test set. Rows show different training strategies, while columns evaluate with different errors. The direct MSE is from \eqnref{mse_loss}, while the I/R, I and R MSEs use \eqnref{ir_loss} with $\lambda = 0.5$, $1$ and $0$, respectively. The reference is the input image without reconstruction. Adding skip-connections improves the result to a large extent. The illuminance + reflectance loss lowers both direct MSE and in terms of illuminance and reflectance, as compared to optimizing for only the direct loss. Pre-training has a significant impact on the result.}
	\begin{tabular}{l|llll}
		&\hspace{0cm}\rotatebox[origin=l]{\crot}{Direct} &
		\hspace{0cm}\rotatebox[origin=l]{\crot}{I/R} &
		\hspace{0cm}\rotatebox[origin=l]{\crot}{I} &
		\hspace{0cm}\rotatebox[origin=l]{\crot}{R}\\
		\hline
		\rowcolor{rc}
		Reference   & 0.999  &  0.890  &  0.712  &  0.178 \\
		Without skip-conn.   & 0.249  &  0.204  &  0.102 &  0.102  \\
		\rowcolor{rc}
		Direct loss (eq. \ref{eqn:mse_loss})    & 0.189  &  0.159  &  0.090  &  0.069  \\
		I/R loss (eq. \ref{eqn:ir_loss}) & 0.178  &  0.150  &  0.081  &  0.068 \\
		\rowcolor{rc}
		Pre-train. + I/R loss & \bf{0.159}  &  \bf{0.134}  &  \bf{0.069} &   \bf{0.066} \\
		
	\end{tabular}
	\label{tab:error}
\end{table}

\section{Results}
In this section we present a number of examples, verifying the quality of the HDR reconstructions. Additional visual examples can be found in the supplementary material and video. Furthermore, for prediction using any LDR image the CNN together with trained parameters can be downloaded from: {\it\url{https://github.com/gabrieleilertsen/hdrcnn}}.

\customsection{Test errors}
To justify the different model and training strategies explained in \secref{method} and \ref{sec:training}, we evaluate the success of the optimization in \tabref{error}. The errors of the different configurations have been averaged over the test set of 95 HDR images, reconstructed at $1024 \times 768$ pixels resolution. The LDR images used for reconstruction use virtual exposures and clipping such that $5\%$ of the pixels in each image are saturated. 

\tabref{error} shows how different errors are affected by the different training strategies. The CNN without skip-connections can drastically reduce the MSE of the input. However, adding the skip-connections reduces error by an additional $24\%$, and creates images with substantial improvements in details (\figref{skip}). Comparing the two different loss functions in \eqnref{mse_loss} and \ref{eqn:ir_loss}, the latter I/R loss shows a lower error both in terms of I/R and direct MSE, with a reduction of $5.8\%$. Finally, with pre-training and the I/R loss the best training performance is accomplished, lowering the error by $10.7\%$ as compared to no pre-training. All the trainings/pre-trainings that do no use our pre-trained parameters have been initialized with VGG16 encoder weights trained for classification of the Places dataset \cite{Zhou2014}.

\begin{figure}
	\vspace{5pt}
	\newcommand\ww{0.235}
	\centering
	\includegraphics[width=\ww\textwidth]{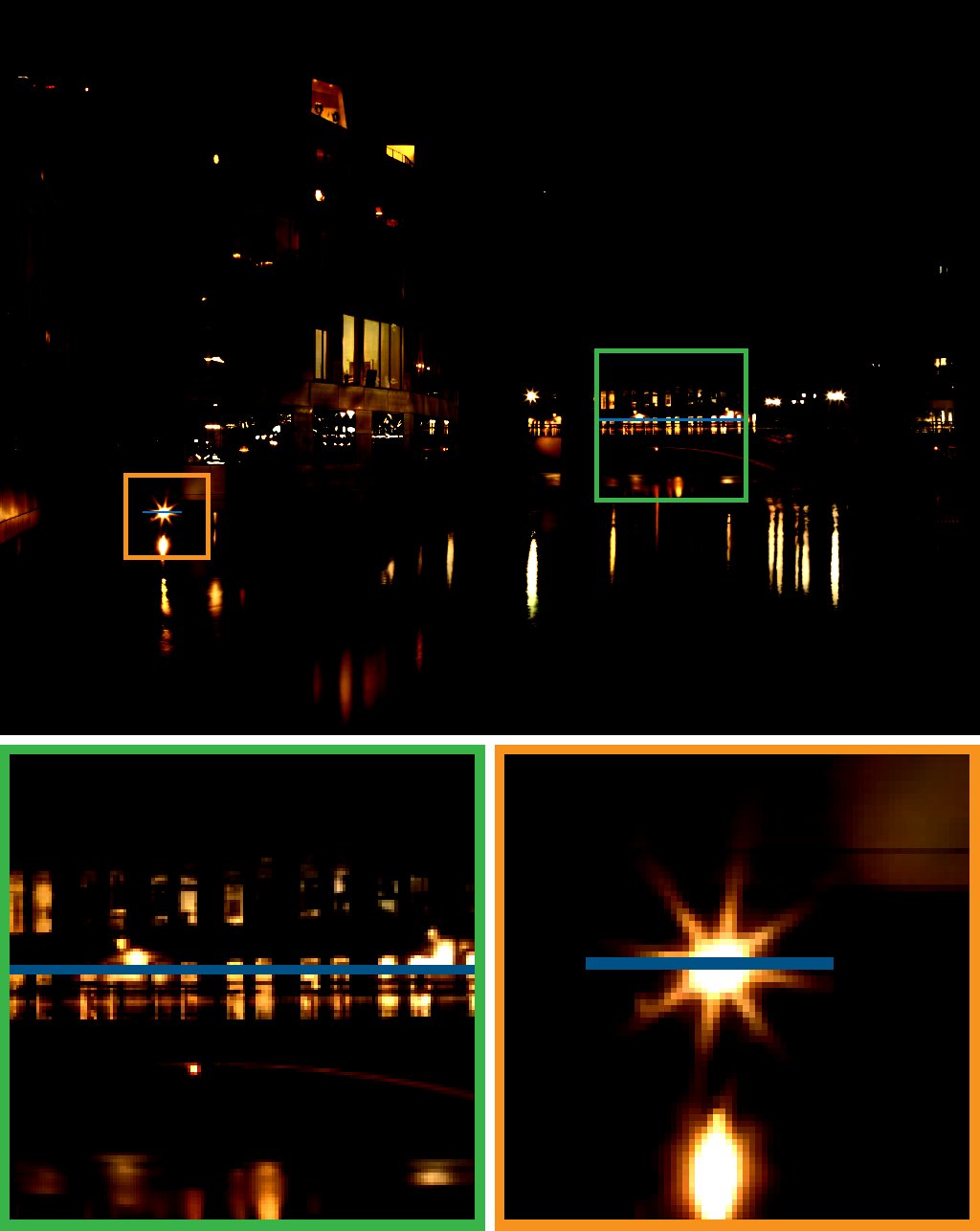}
	\includegraphics[width=\ww\textwidth]{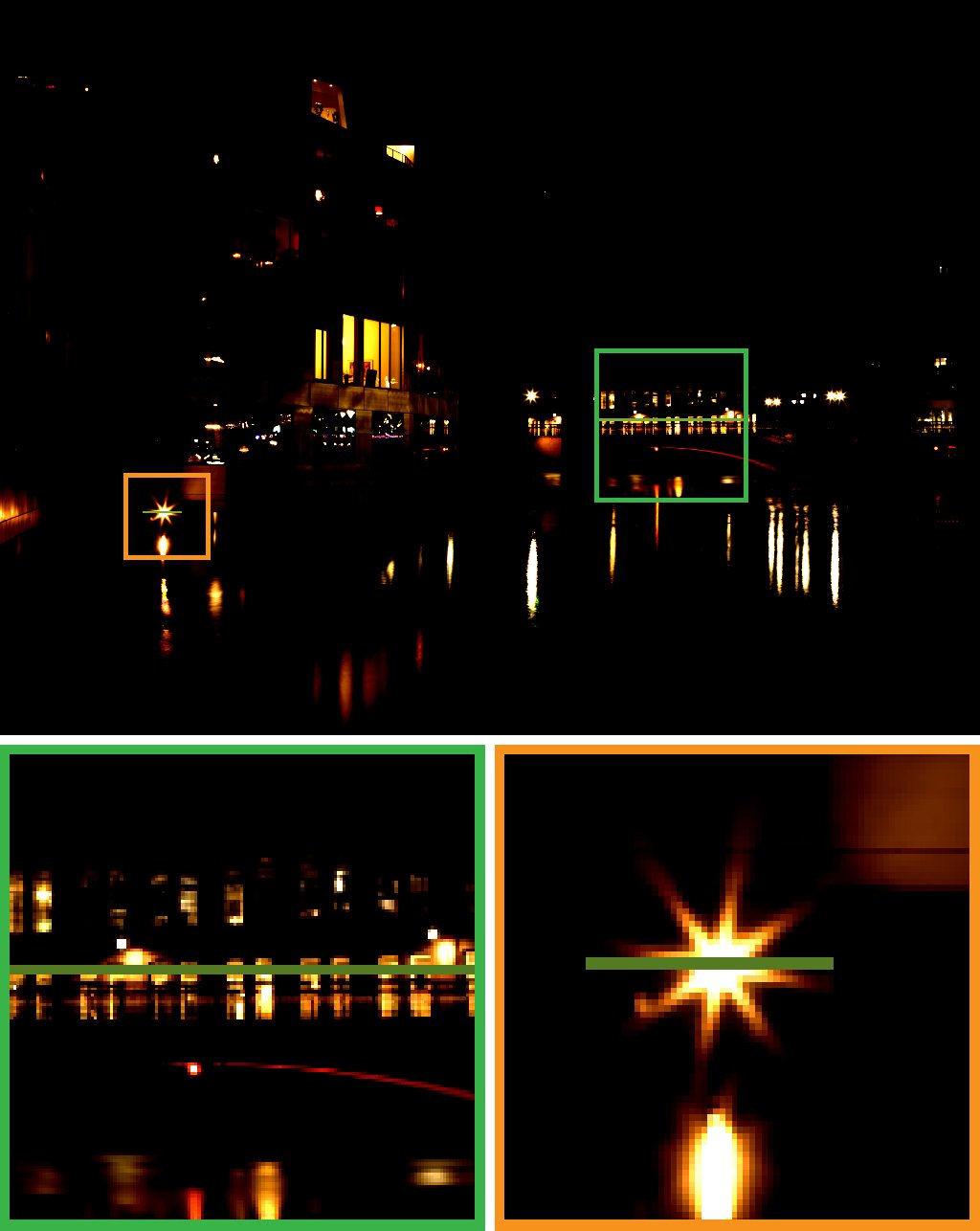}
	\vspace{-2pt}\\
	\includegraphics[width=0.49\linewidth]{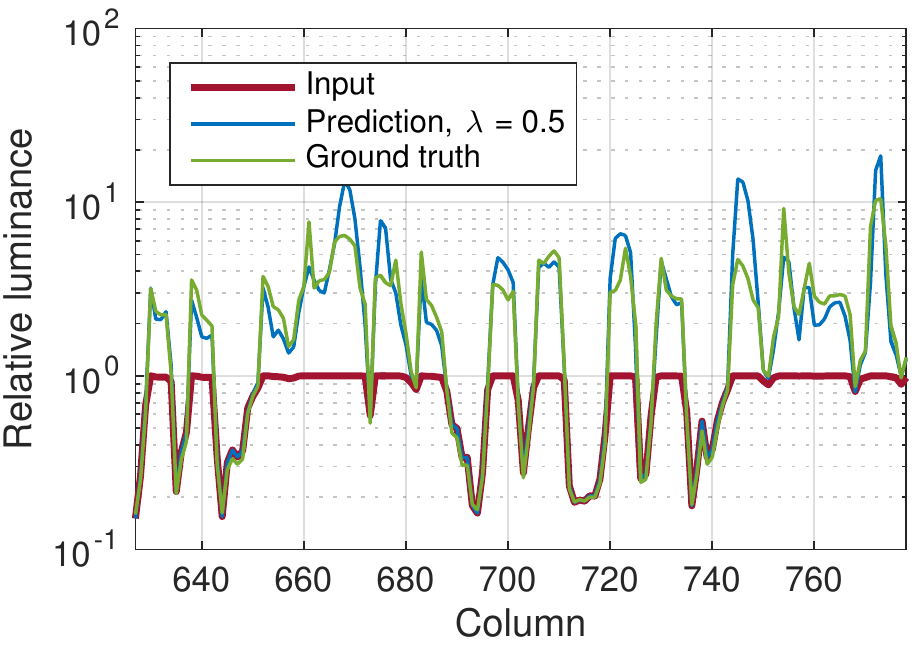}
	\includegraphics[width=0.49\linewidth]{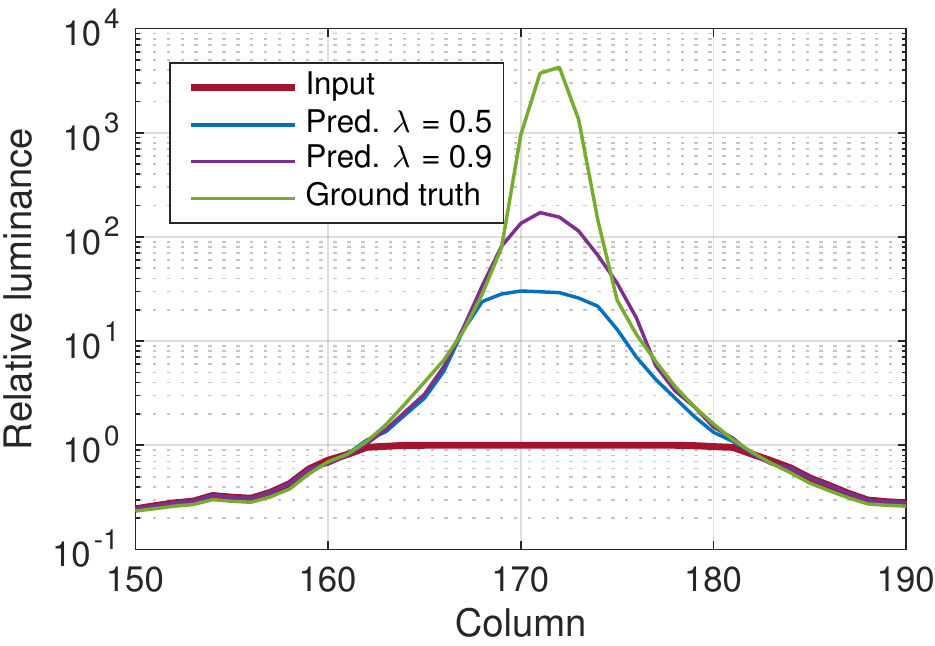}
	\vspace{-7pt}\\
	\caption{\label{fig:residual} The learned residual of the image in \figref{teaser} (top left), together with the ground truth residual (top right). The plots show relative luminance values across the two scanlines marked in the images. Complex regions are predicted convincingly (bottom left), but for the very intense spotlight the signal is underestimated (bottom right). The underestimation is less pronounced when training with higher illuminance weight ($\lambda$ in \eqnref{ir_loss}). }
	\vspace{\belowfigspace}
\end{figure}

\customsection{Comparisons to ground truth}
\figref{itm_hdr} demonstrates a set of predictions on HDR images from the test set that have been transformed to LDR by the virtual camera described in \secref{hdr_db}. The examples demonstrate successful HDR reconstruction in a variety of situations.
In night scenes, colors and intensities of street lights and illuminated facades can be restored with very convincing quality. The same goes for specular reflections and other smaller highlights in day scenes. Furthermore, in situations where there is some small amount of information left in any of the color channels, details and colors of larger areas can be reconstructed to be very close to the ground truth HDR image. For example, in the right-most column the large area light source does not contain any visual details when inspecting the input image. However, some small amount of invisible information enables recovery of details. Also, while all channels are saturated in the sky in the third column, the spatial context can be utilized to infer a blue color.

In order to visualize the information that is reconstructed by the CNN, \figref{residual} shows the residual, $\vect{\hat{r}} = \max \left(0, \rhdr-1 \right)$. That is, only the information in highlights is shown, which is not present in the input image $\ldr \in [0, 1]$. The information corresponds well with the ground truth residual, $\vect{r} = \max \left(0, \hdr-1 \right)$. The complete input and output signals are also plotted for two different scanlines across the images. Complex lighting of street lights and windows can be recovered convincingly (bottom left). However, in some situations of very intense light sources, the luminance is underestimated (bottom right). We elaborate on such limitations in \secref{conclusion}.

\begin{figure}
	\vspace{5pt}
	\newcommand\ww{0.236}
	\centering
	\includegraphics[width=\ww\textwidth]{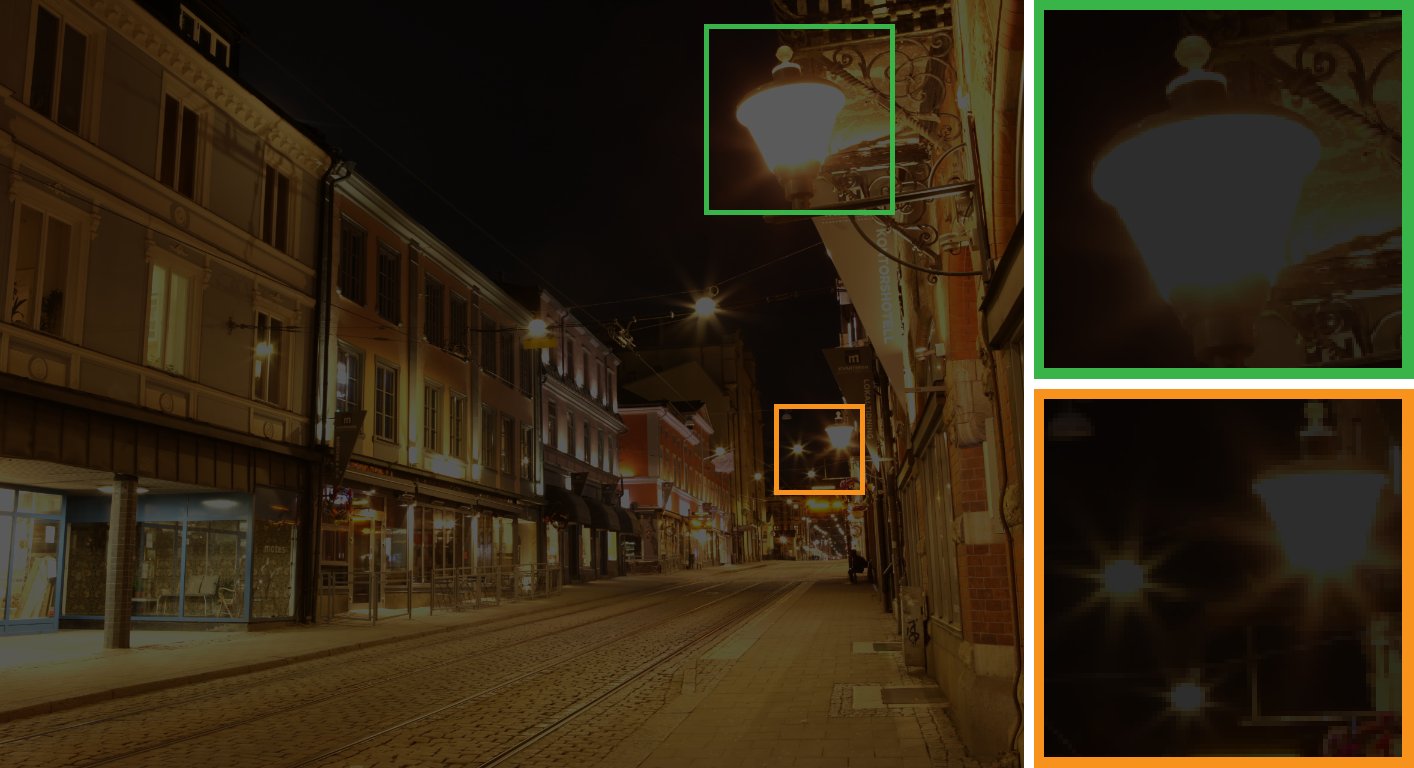}
	\includegraphics[width=\ww\textwidth]{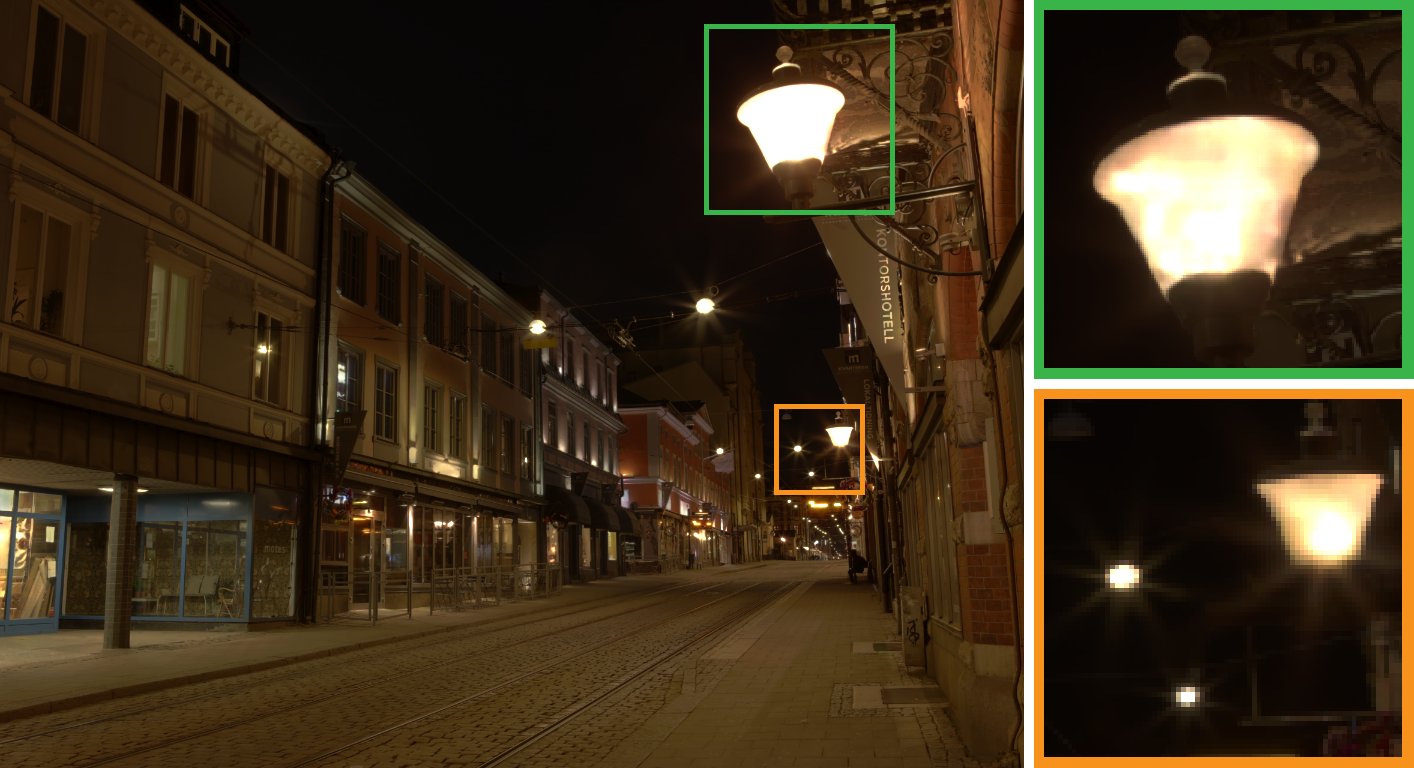}\vspace{3pt}\\
	\vspace{-5pt}
	\subfigure[Camera JPEG input]{\includegraphics[width=\ww\textwidth]{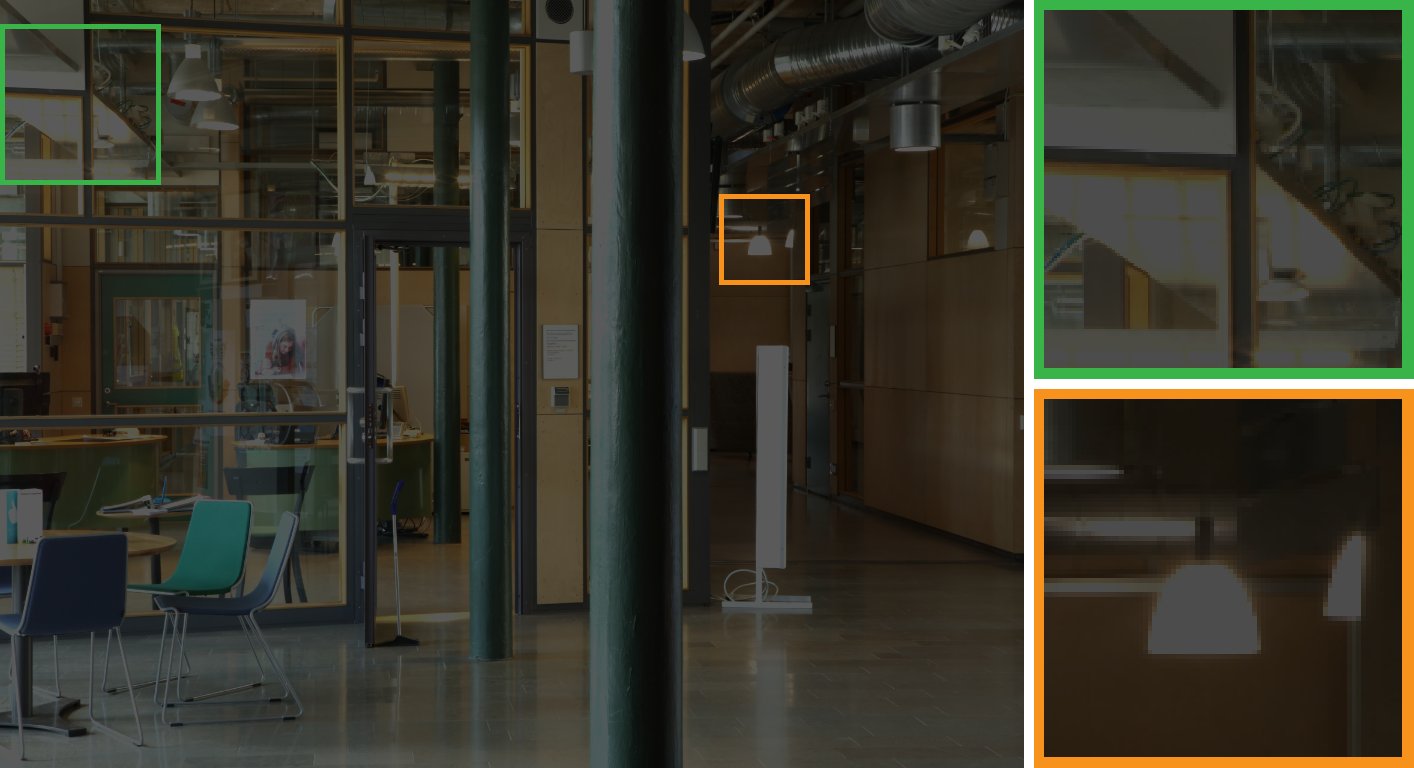}}
	\subfigure[Reconstruction]{\includegraphics[width=\ww\textwidth]{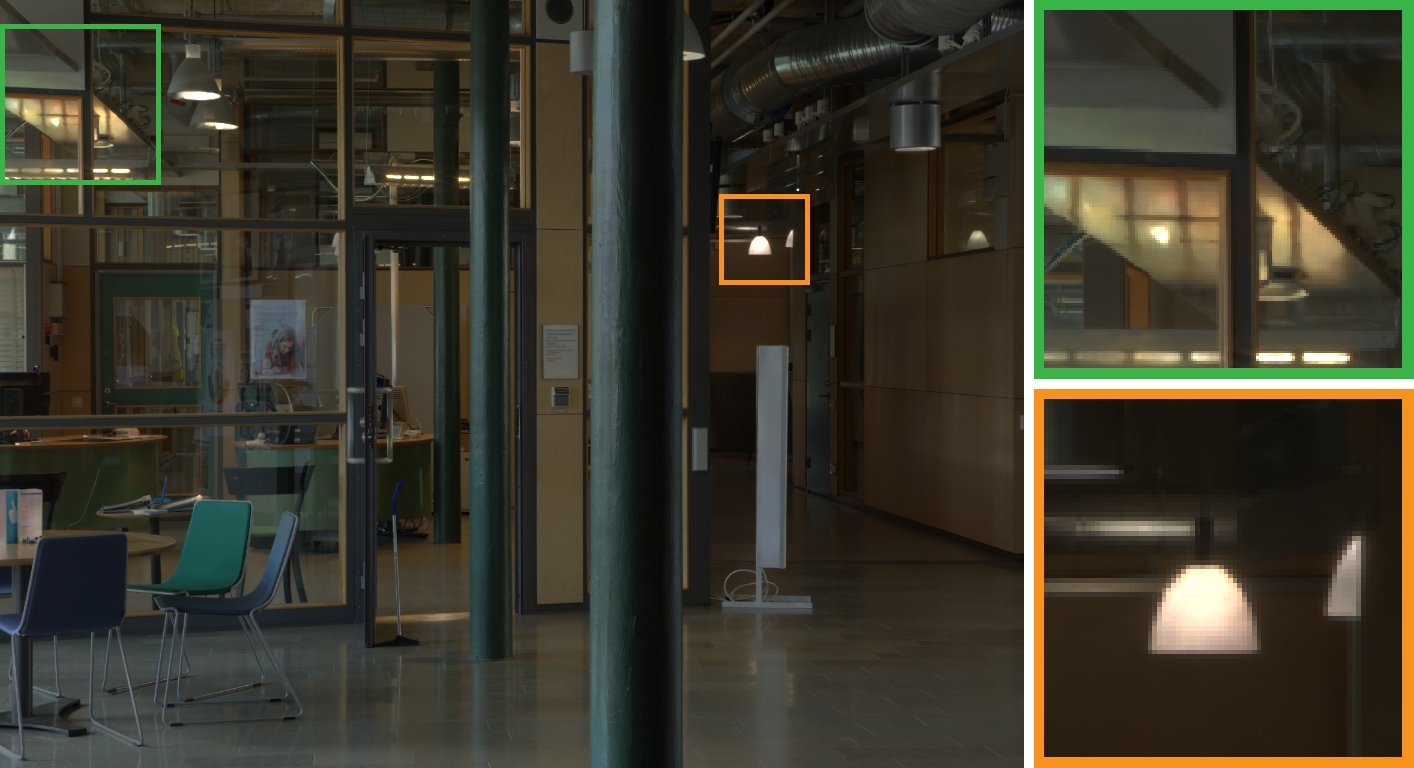}}\vspace{3pt}\\
	\includegraphics[width=\ww\textwidth]{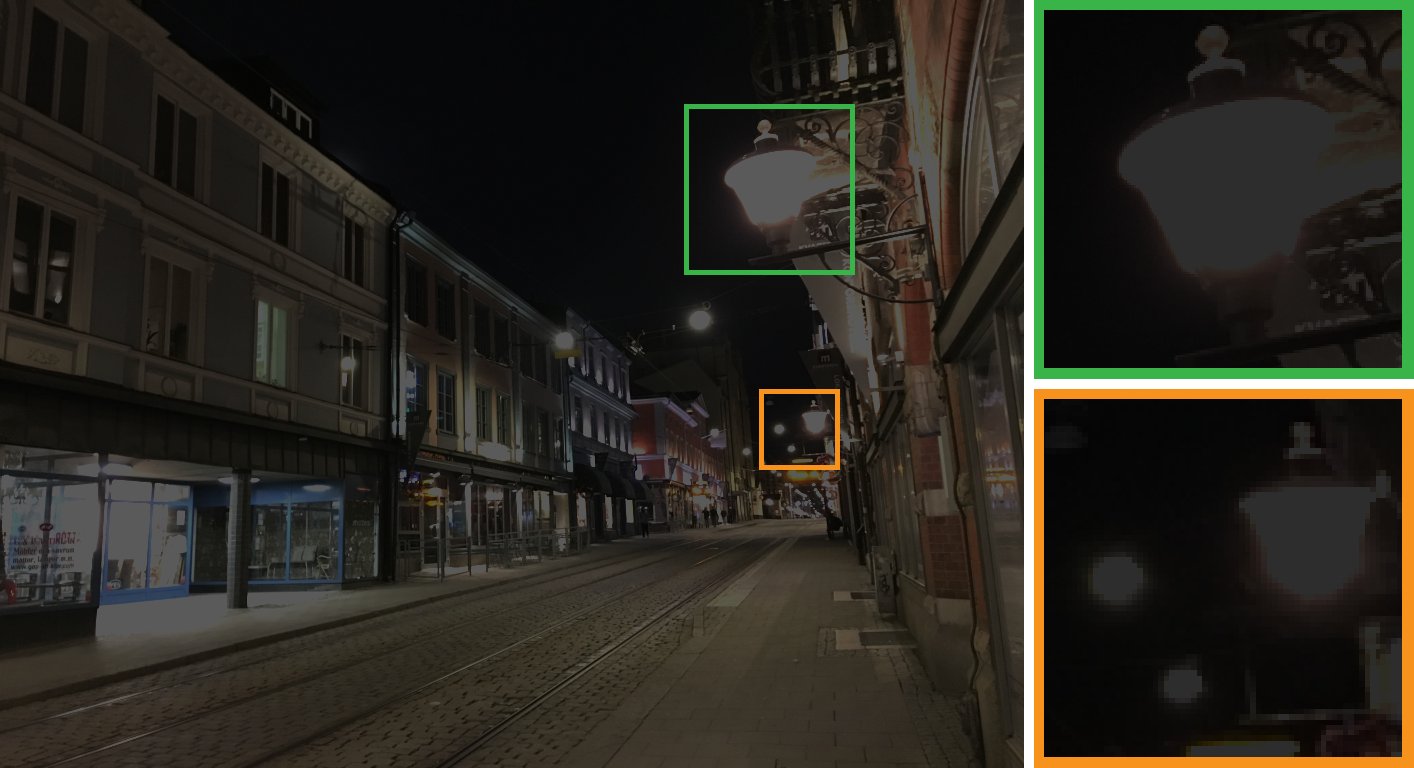}
	\includegraphics[width=\ww\textwidth]{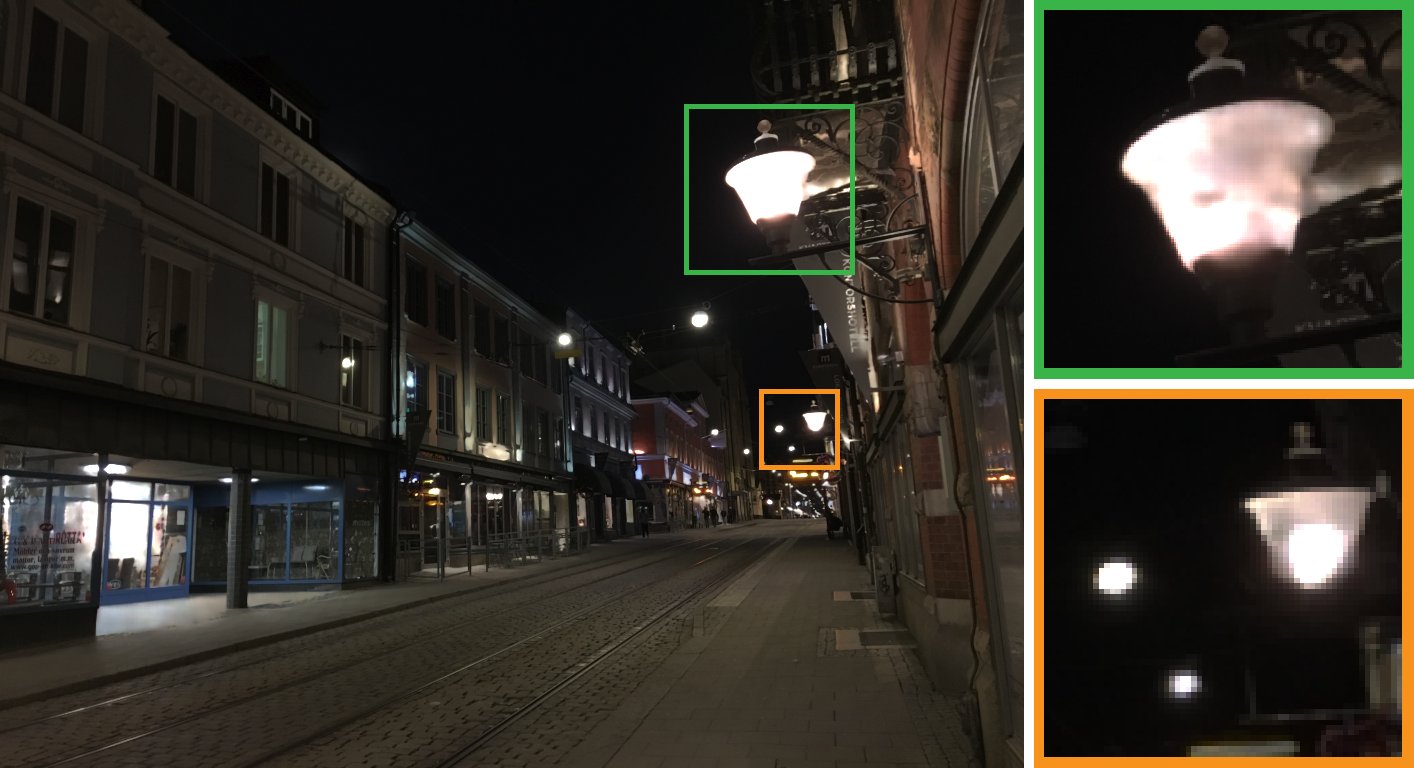}\vspace{3pt}\\
	\vspace{-5pt}
	\subfigure[iPhone image input]{\includegraphics[width=\ww\textwidth]{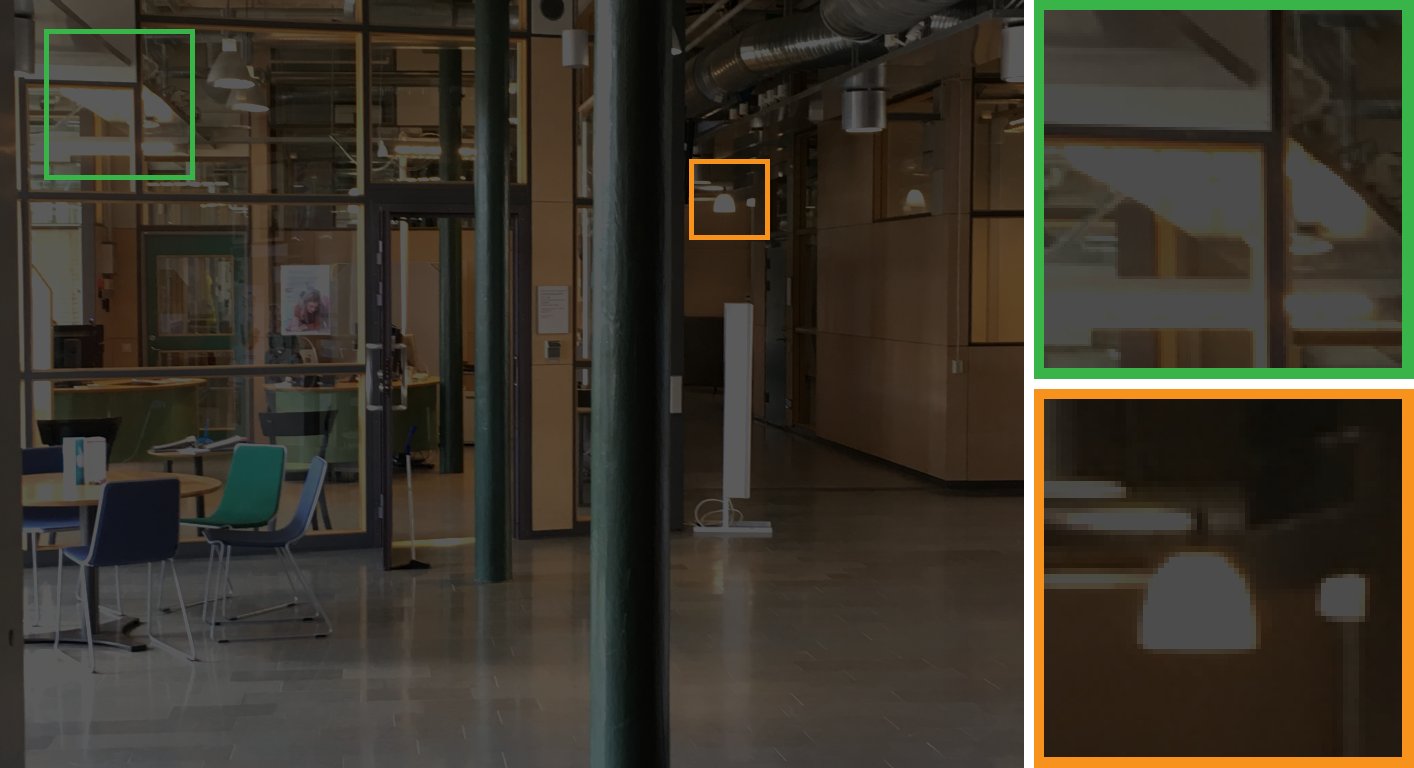}}
	\subfigure[Reconstruction]{\includegraphics[width=\ww\textwidth]{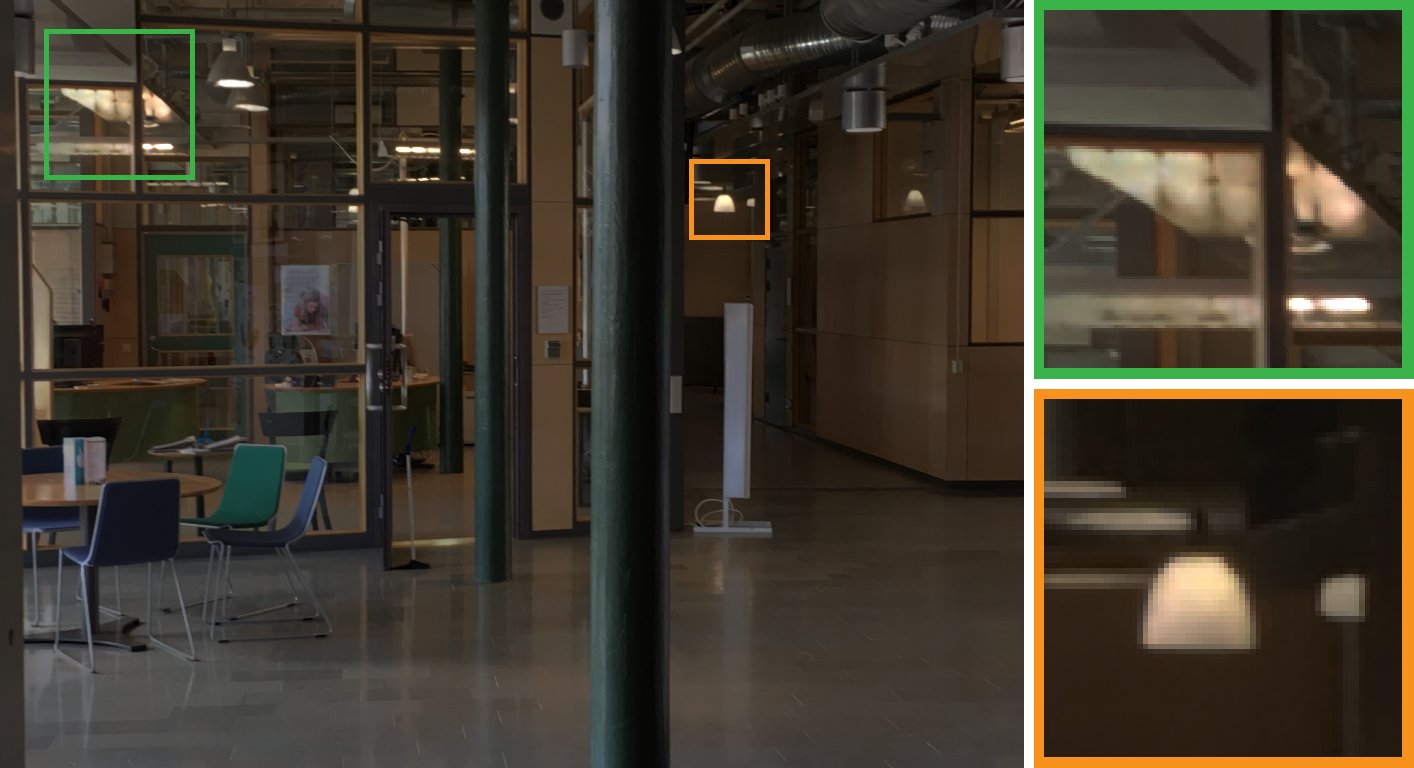}}
	\vspace{-10pt}
	\caption{\label{fig:itm_ldr} Reconstruction from Canon 5DS R camera JPEG (top). The examples are the same as in \figref{itm_hdr}, but with the camera set to store JPEG, applying unknown image transformations and compression. Reconstruction can also be performed with hand-held low-end cameras, such as the iPhone 6S (bottom), where the input image is more degraded.}
\end{figure}

\begin{figure}
	\vspace{5pt}
	\newcommand\ww{0.236}
	\centering
	\includegraphics[width=\ww\textwidth]{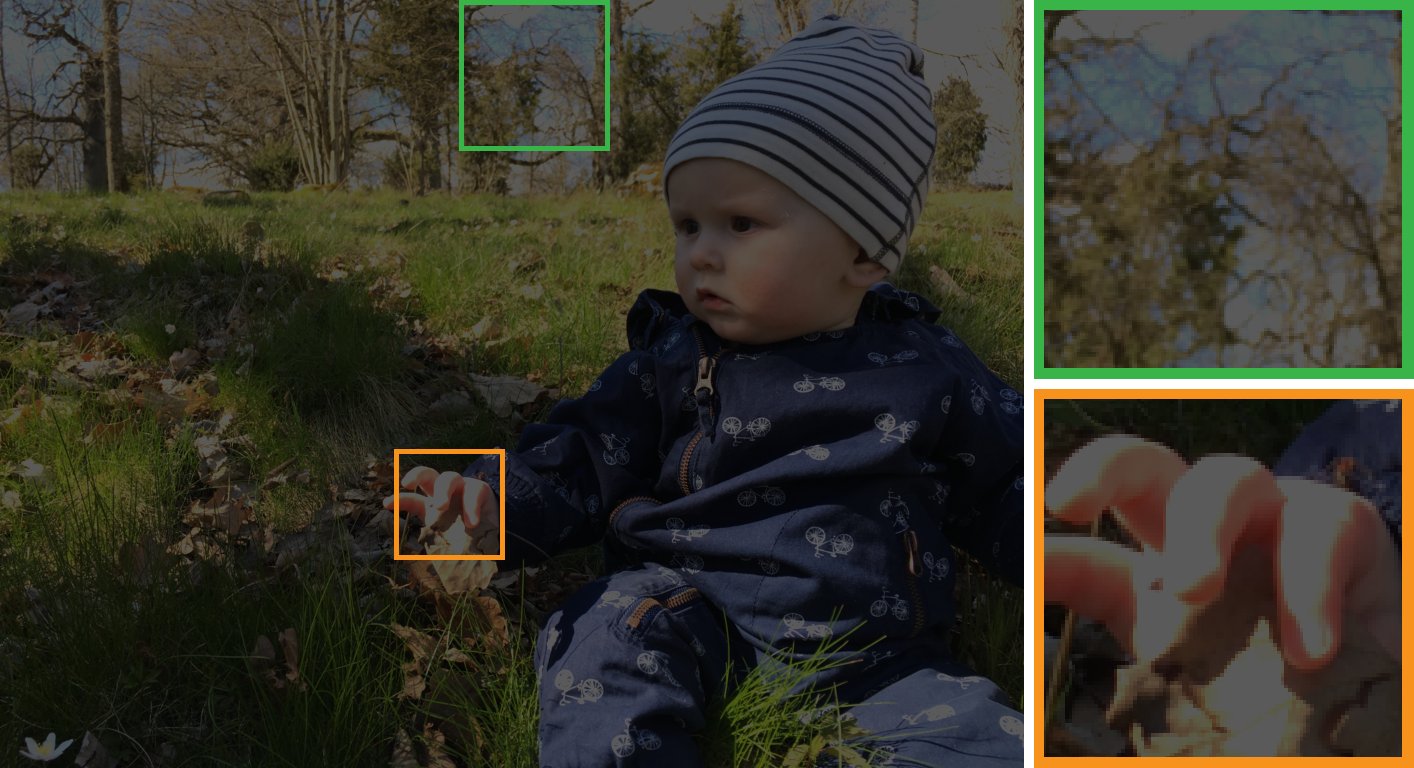}
	\includegraphics[width=\ww\textwidth]{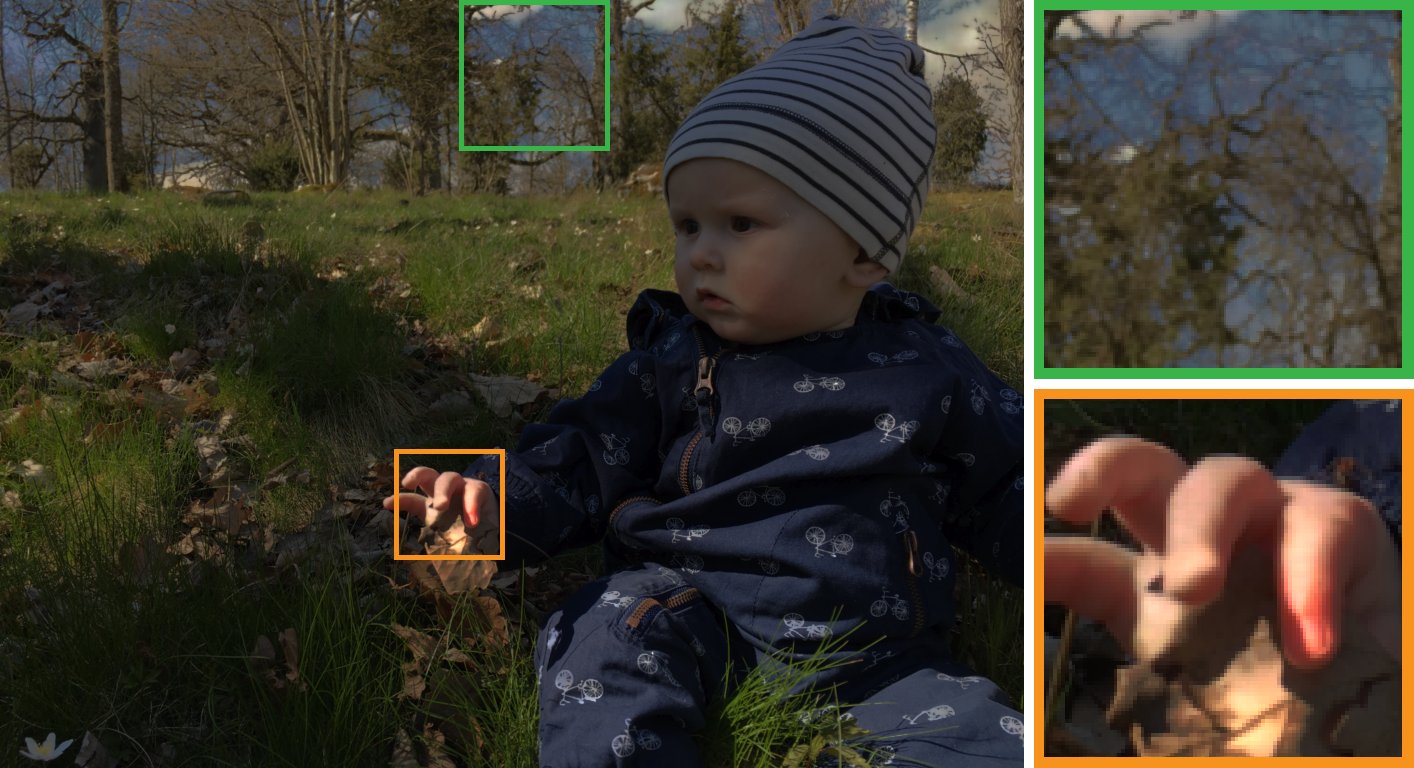}\vspace{3pt}\\
	\includegraphics[width=\ww\textwidth]{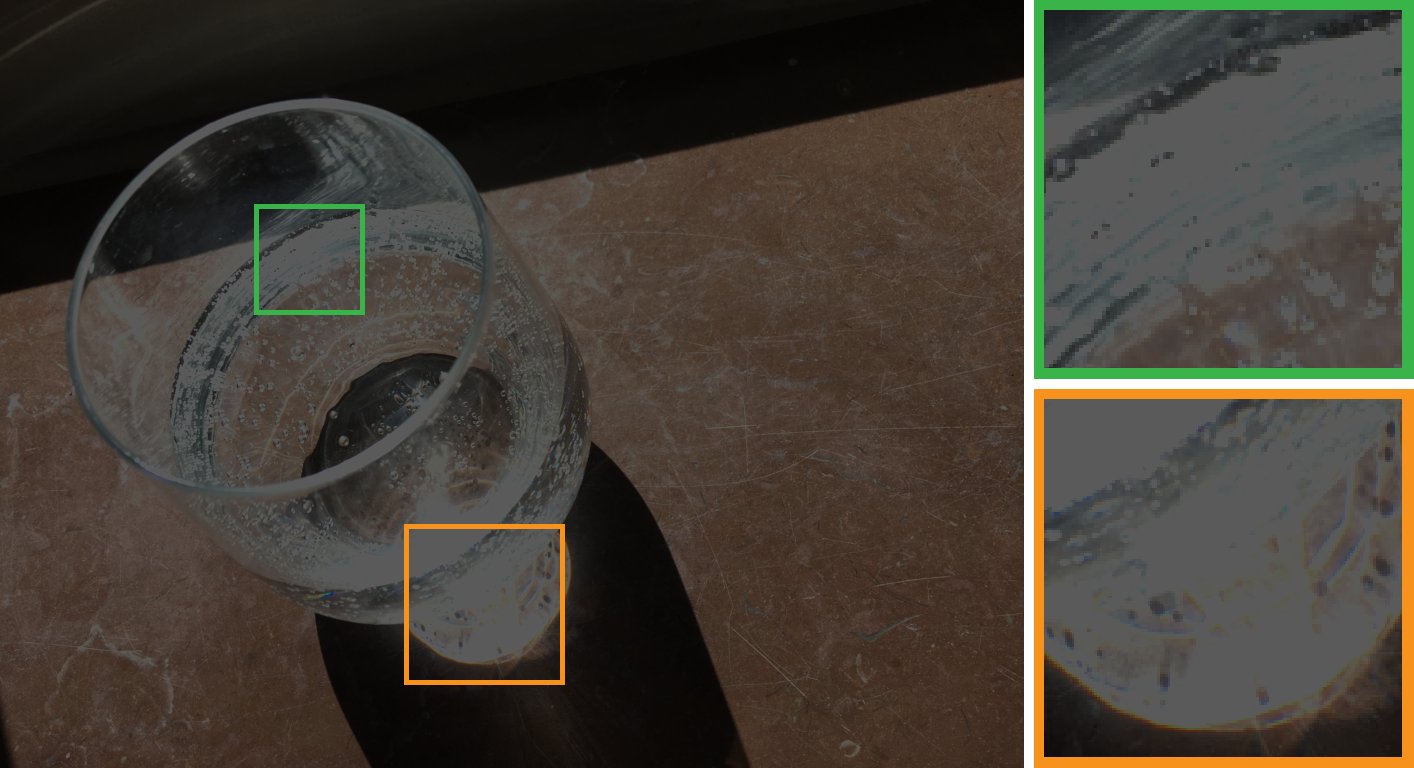}
	\includegraphics[width=\ww\textwidth]{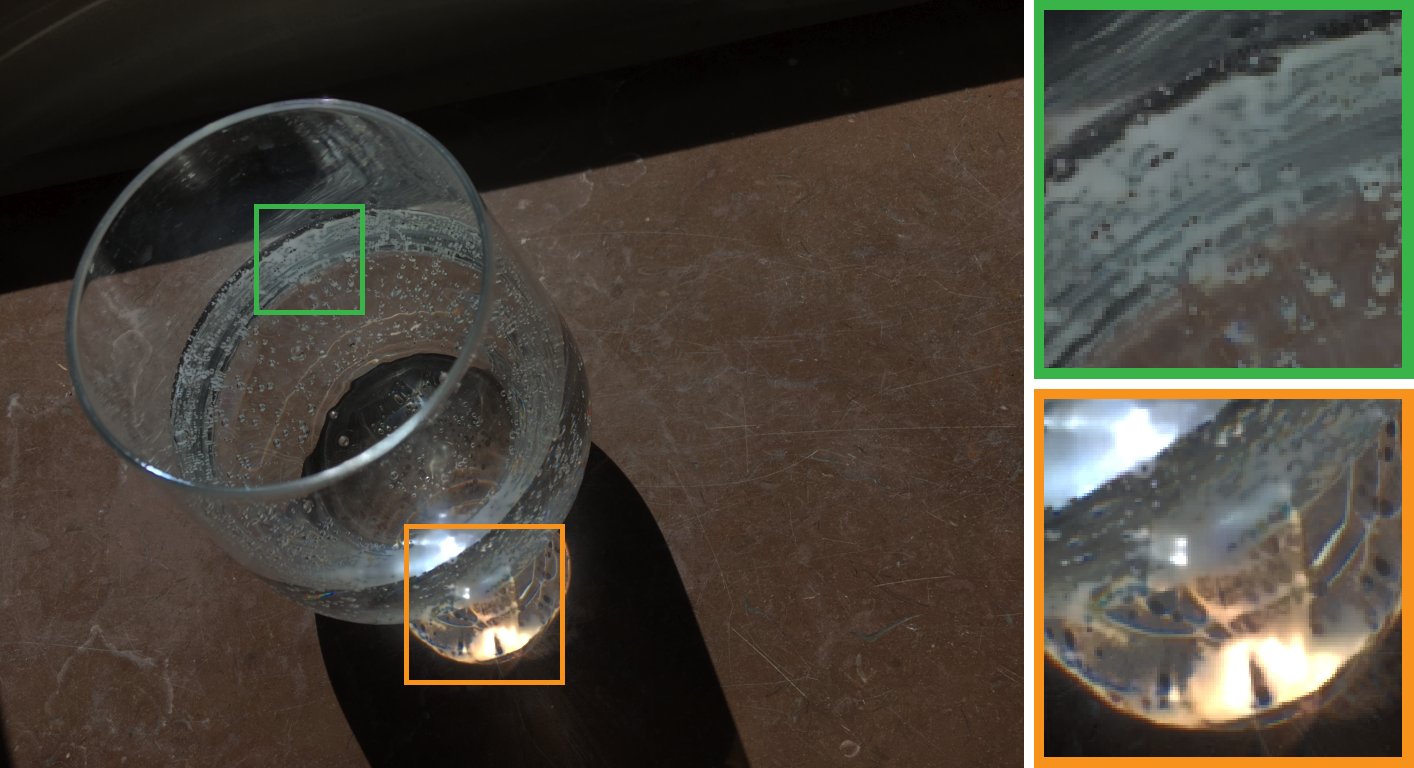}\vspace{3pt}\\
	\includegraphics[width=\ww\textwidth]{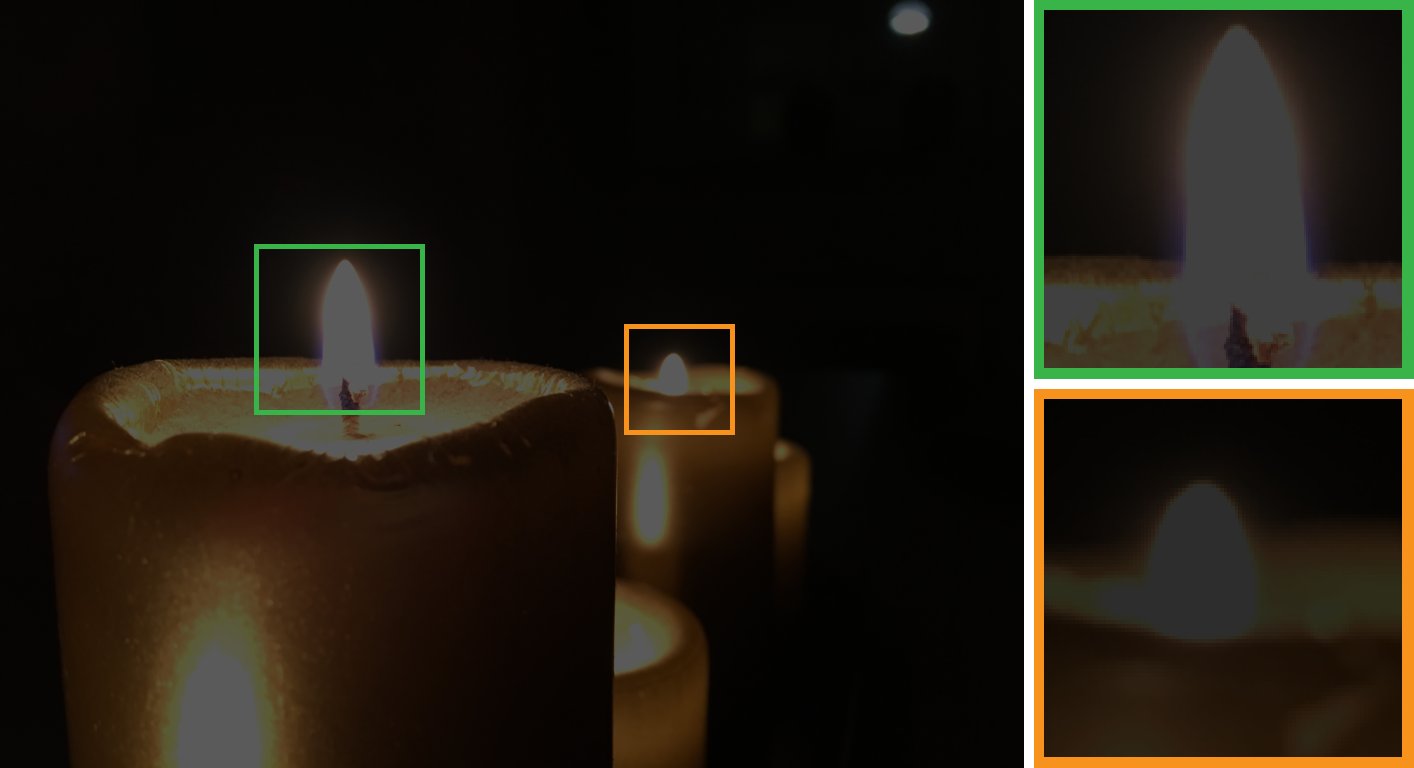}
	\includegraphics[width=\ww\textwidth]{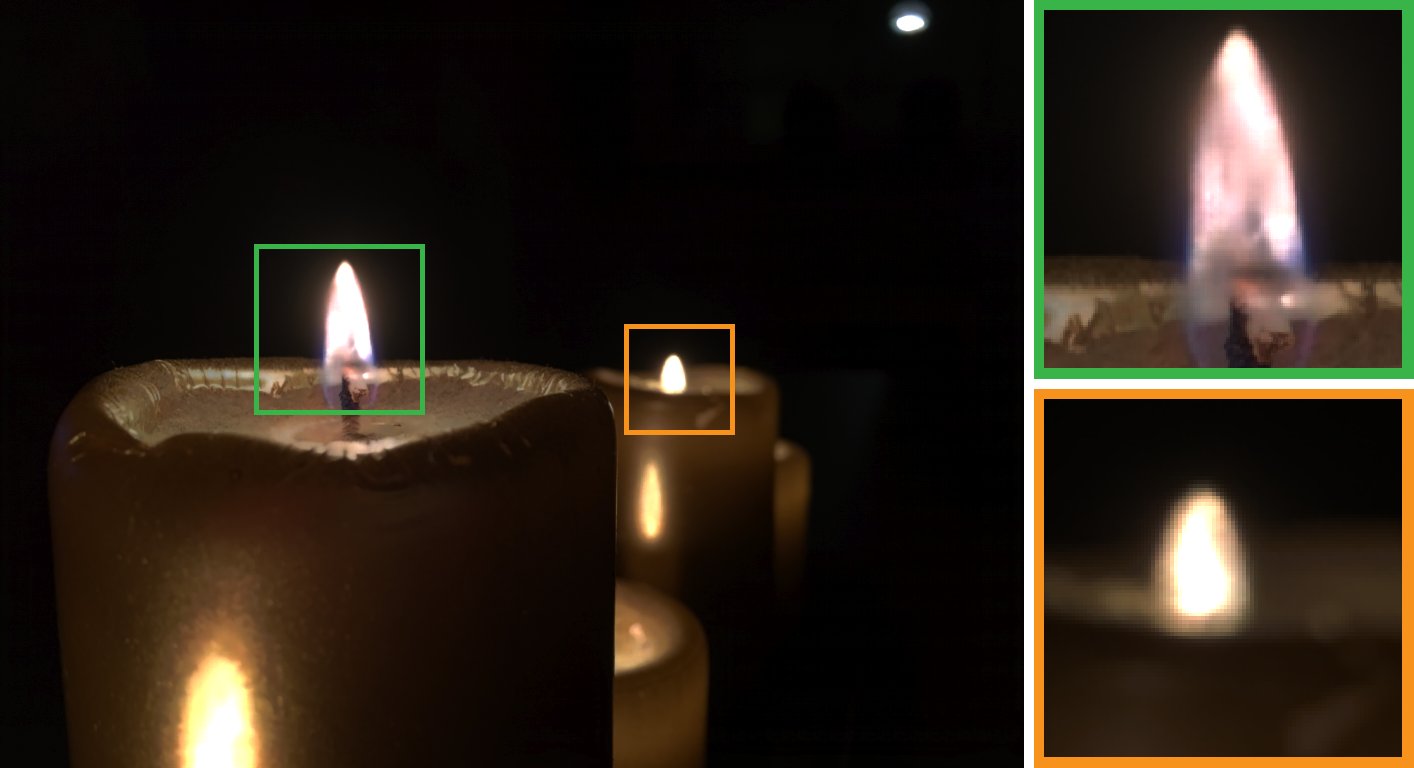}\vspace{3pt}\\
	\vspace{-5pt}
	\subfigure[Input]{\includegraphics[width=\ww\textwidth]{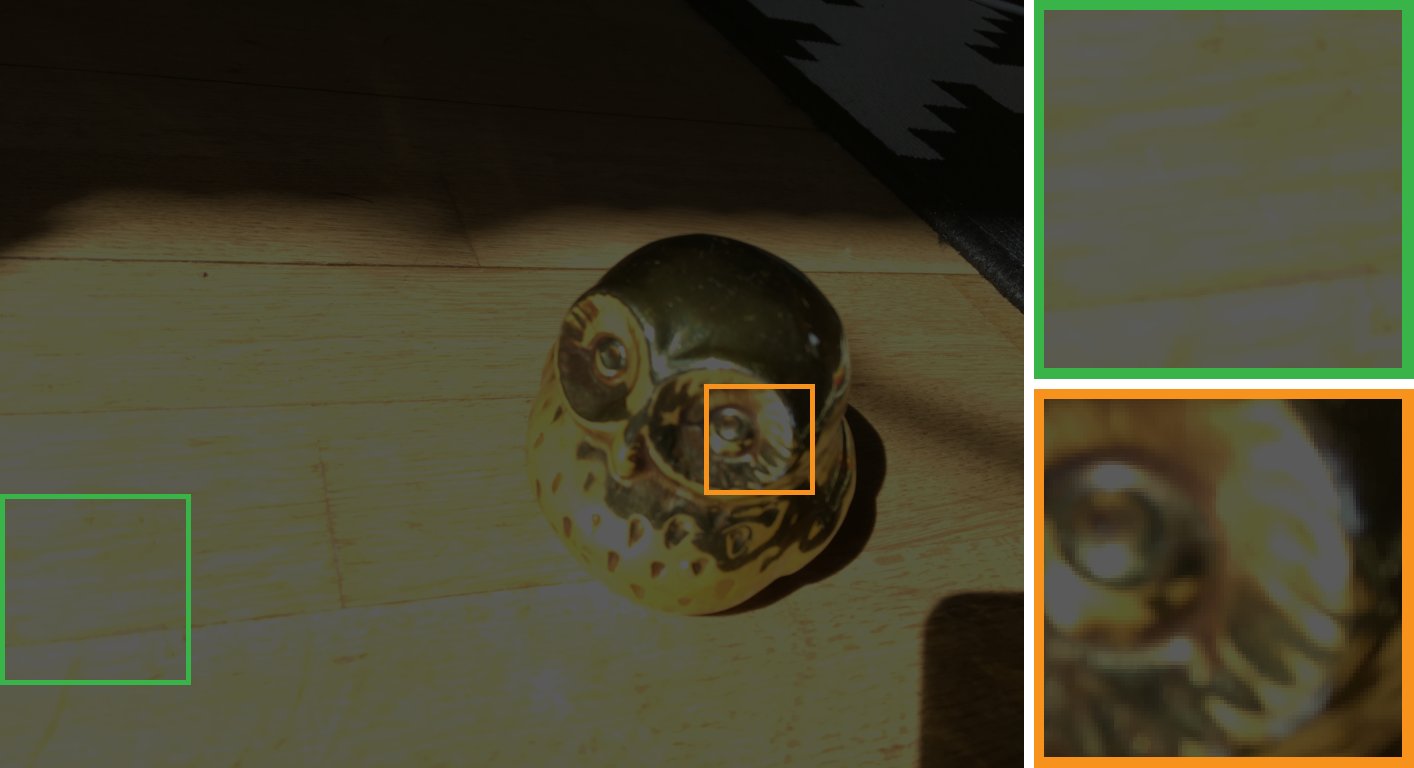}}
	\subfigure[Reconstruction]{\includegraphics[width=\ww\textwidth]{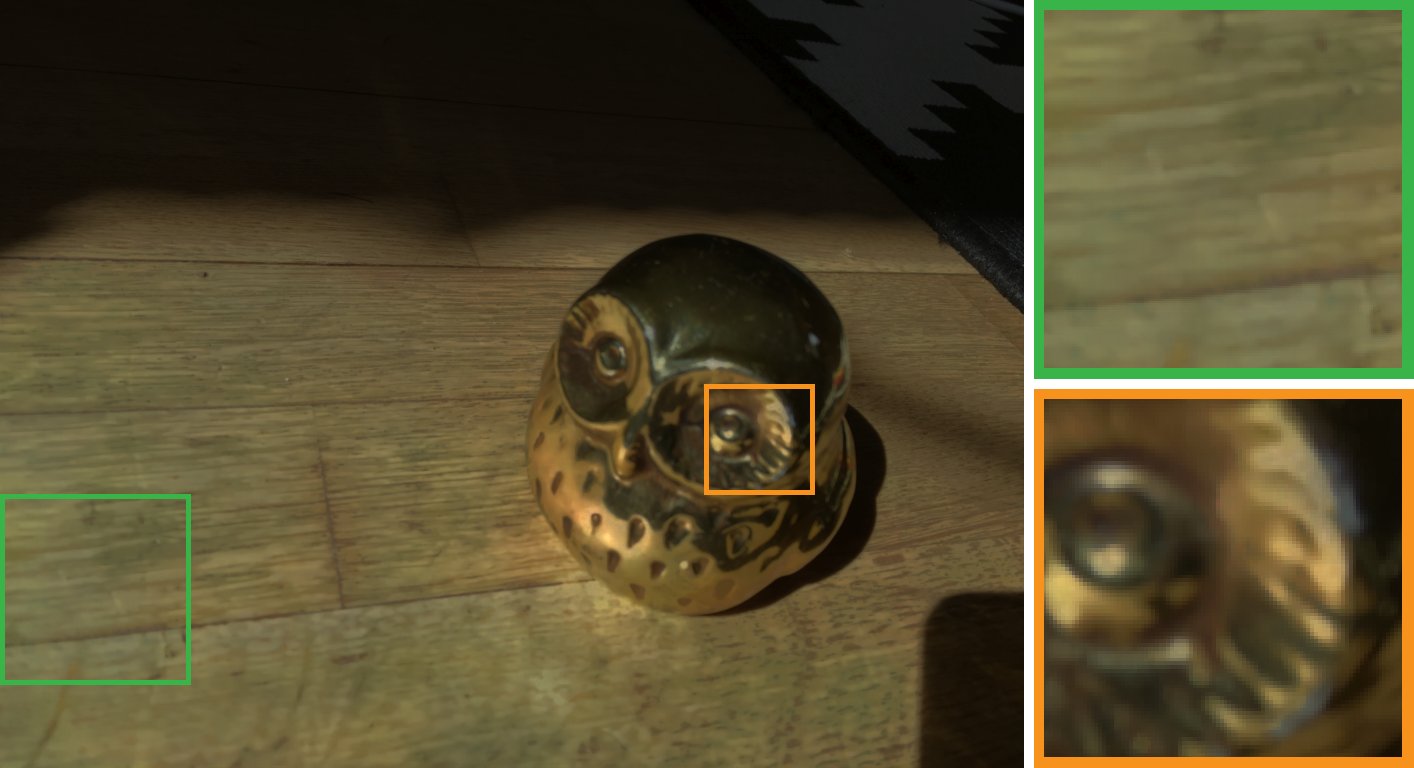}}
	\vspace{-10pt}
	\caption{\label{fig:itm_iphone} Predictions on iPhone camera images. Plausible reconstructions can be made of skin tones, caustics and fire (row 1-3). Large saturated areas can be recovered if there still is information in one of the color channels (bottom row).}
	\vspace{\belowfigspace}
\end{figure}

\customsection{Reconstruction with real-world cameras}
In order to show that the HDR reconstruction model generalizes to real-world cameras, \figref{itm_ldr} shows two of the scenes from \figref{itm_hdr}, captured using a Canon 5DS R camera's JPEG mode (top row) and with an iPhone 6S camera (bottom row). Both these cameras provide more realistic scenarios as compared to the virtual camera.
Nevertheless, reconstructions of equal quality can be done from camera JPEGs. The iPhone images are more degraded, shot in dark conditions without a tripod, but the reconstructed information comply well with the image characteristics. To further explore the possibilities in reconstructing everyday images, \figref{itm_iphone} displays a set of iPhone images, taken in a variety of situations. The examples not only demonstrate the method's ability to generalize to a different camera, but also to a wide range of situations, such as skin tones, fire and caustics. The most limiting factor when reconstructing from iPhone images is the hard JPEG compression that has been applied, which results in small blocking artifacts close to highlights, affecting the final reconstruction. In order to compensate for these, the brightness of the images has been increased by a small factor, followed by clipping. This removes the artifacts caused by harder compression in the brightest pixels, which improves the final reconstruction quality.

\begin{figure}
	\vspace{5pt}
	\newcommand\ww{0.116}
	\centering
	\includegraphics[width=\ww\textwidth]{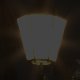}
	\includegraphics[width=\ww\textwidth]{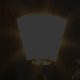}
	\includegraphics[width=\ww\textwidth]{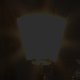}
	\includegraphics[width=\ww\textwidth]{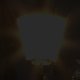}\\
	\vspace{-2pt}
	\subfigure[4\%]{\includegraphics[width=\ww\textwidth]{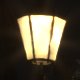}}
	\subfigure[6\%]{\includegraphics[width=\ww\textwidth]{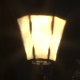}}
	\subfigure[8\%]{\includegraphics[width=\ww\textwidth]{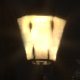}}
	\subfigure[10\%]{\includegraphics[width=\ww\textwidth]{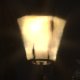}}
	\vspace{-10pt}
	\caption{\label{fig:exp} Zoom-ins of reconstructions (bottom row) with different exposure settings of the input (top row). The numbers indicate how large fraction of the total number of pixels are saturated in the input. The images have then been scaled to have the same exposure after clipping has been applied. Although illuminance is predicted at approximately the same level, more details are available in reconstruction of the shorter exposure images.}
	\vspace{\belowfigspace}
\end{figure}

\customsection{Changing clipping point}
In order to demonstrate the behavior of the reconstruction with varying amount of information loss, \figref{exp} shows predictions using different virtual exposure times. As the training of the CNN uses a virtual camera with different exposure settings, part of the objective is to minimize the difference between these, apart from a scaling factor. However, since more information is available in highlights with shorter exposure, in most cases there will be visible differences in the reconstruction quality, as exemplified by the figure.

\begin{figure}
	\newcommand\ww{0.23}
	\centering
	\subfigure[Input LDR]{\includegraphics[width=\ww\textwidth]{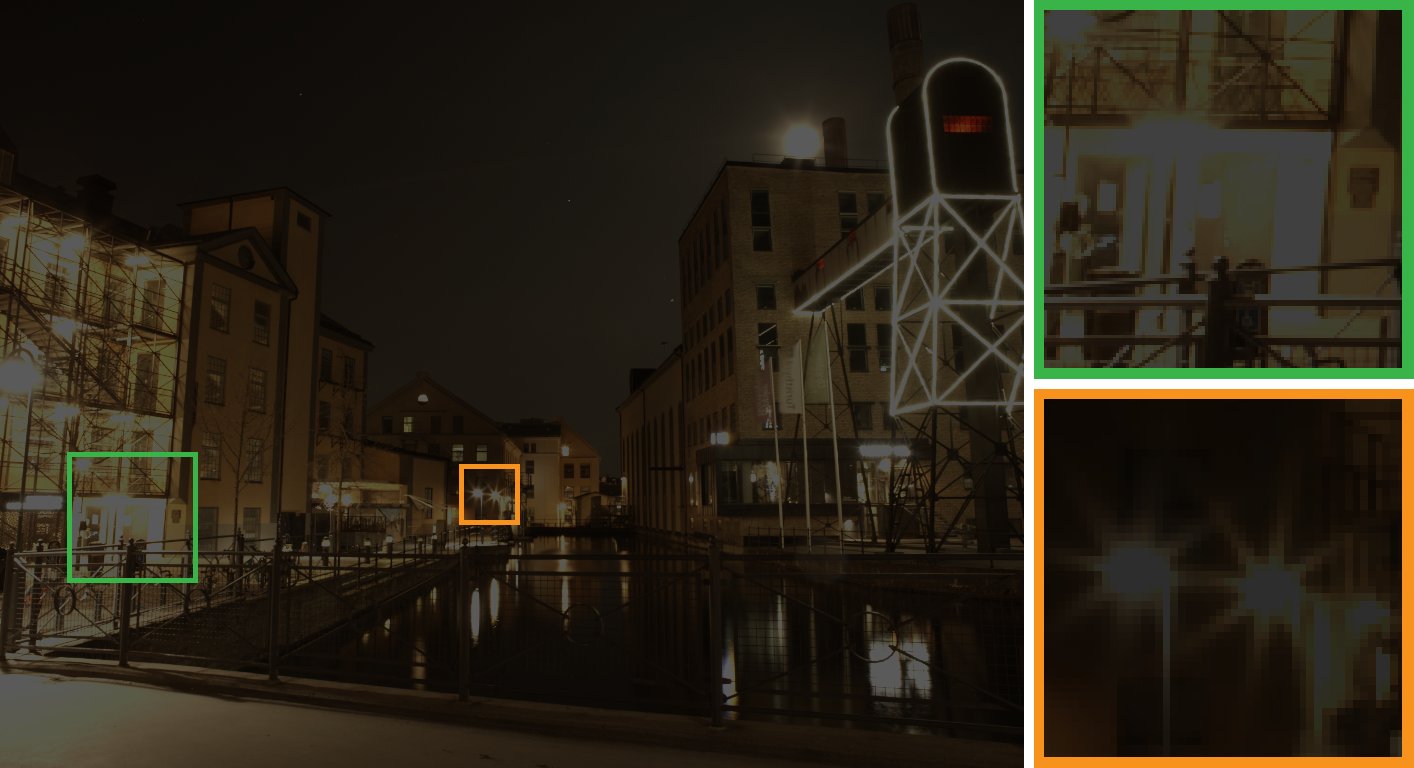}}
	\subfigure[\cite{Banterle2008}]{\includegraphics[width=\ww\textwidth]{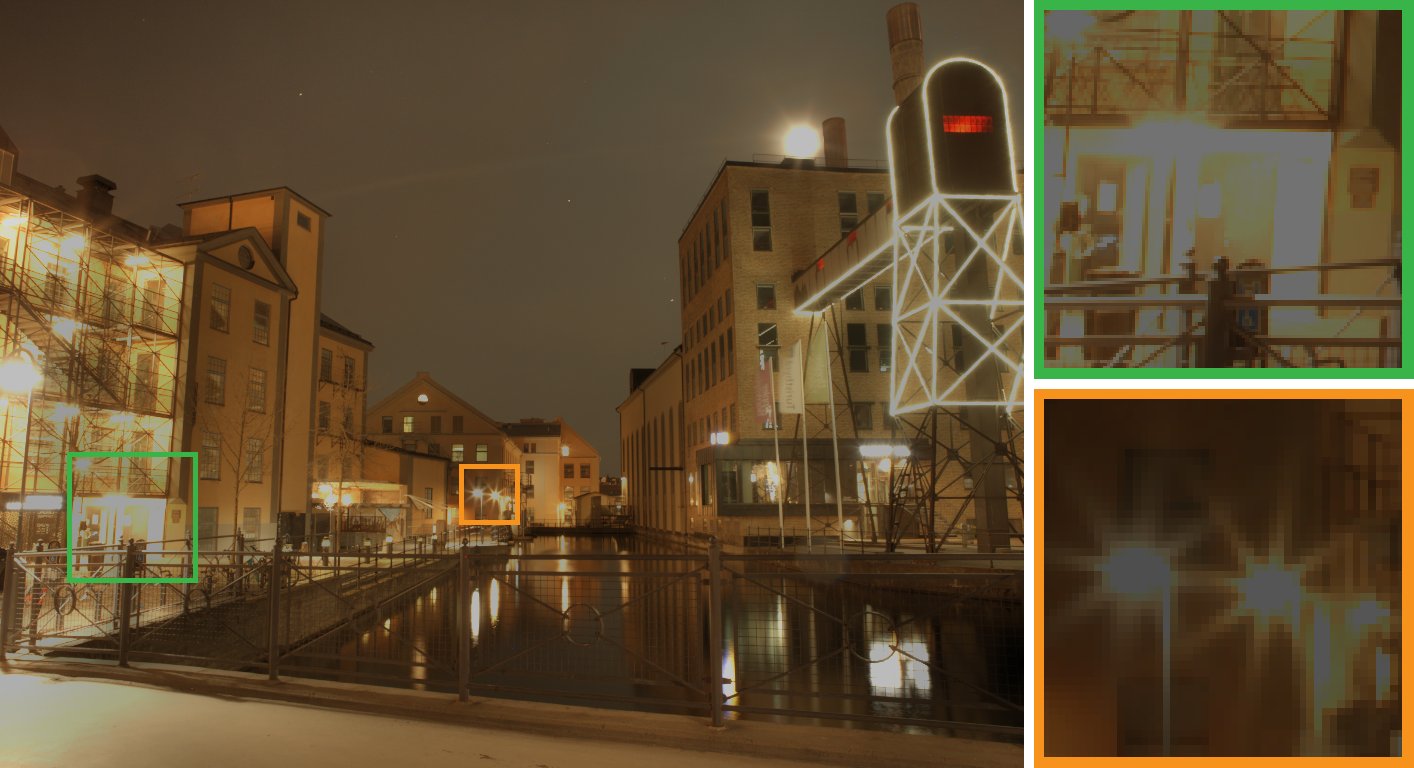}}\\
	\vspace{-5pt}
	\subfigure[\cite{Meylan2006}]{\includegraphics[width=\ww\textwidth]{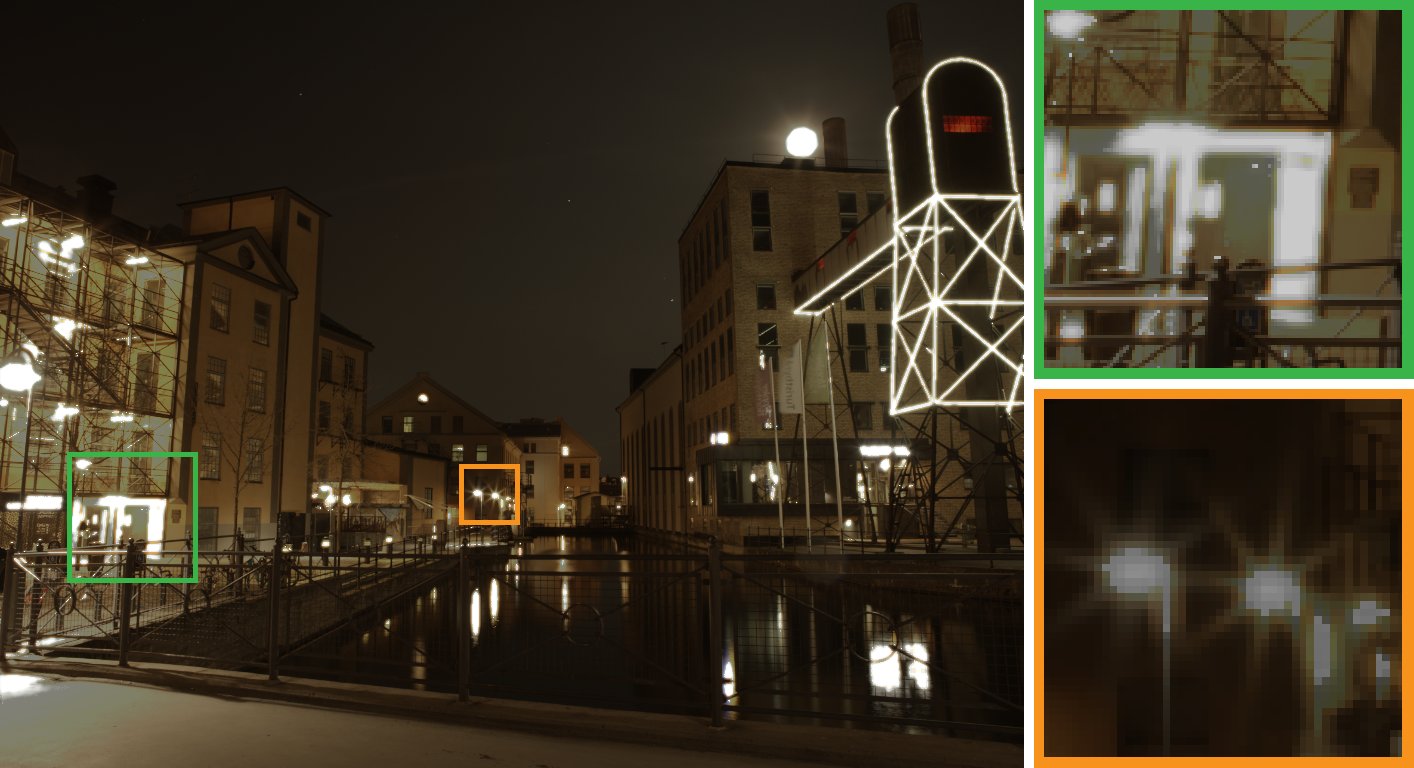}}
	\subfigure[\cite{Rempel2007}]{\includegraphics[width=\ww\textwidth]{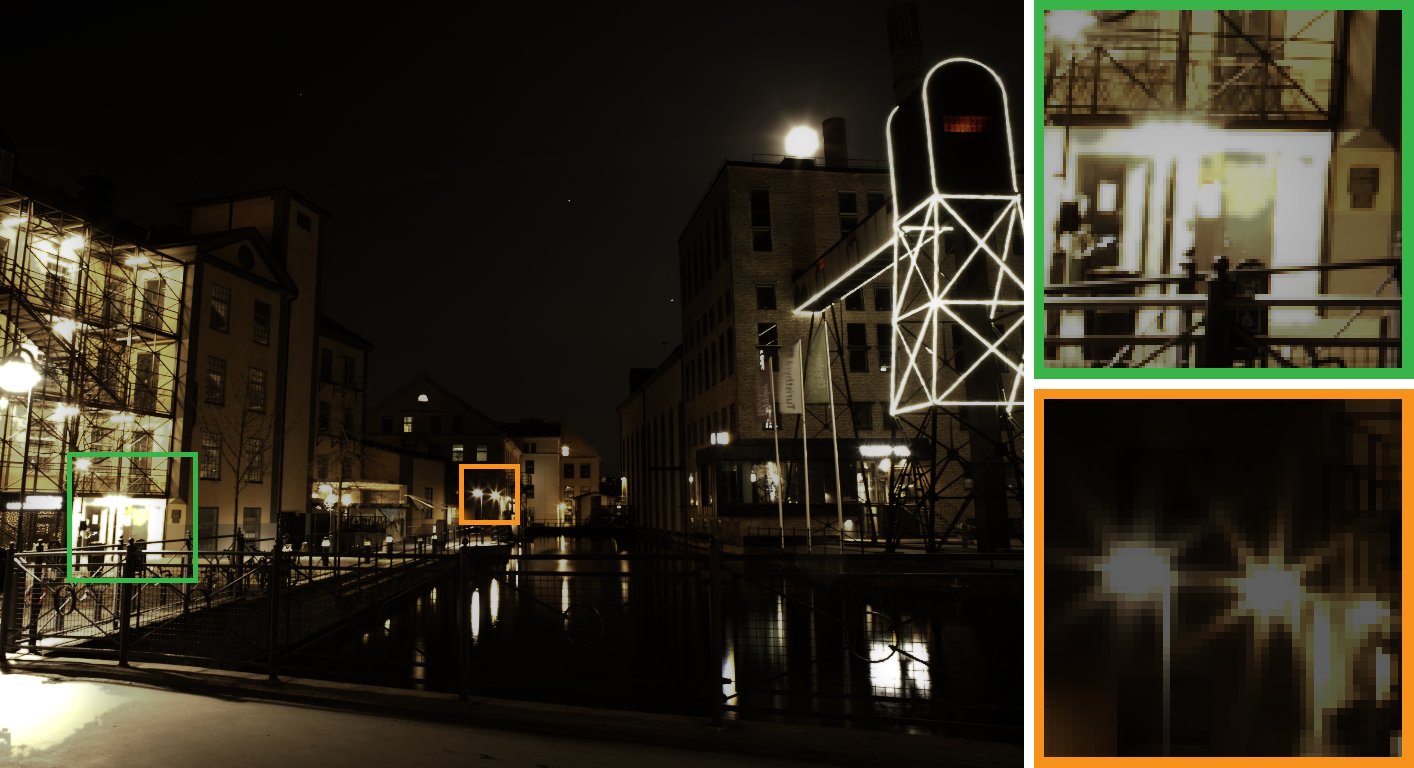}}\\
	\vspace{-5pt}
	\subfigure[Ours]{\includegraphics[width=\ww\textwidth]{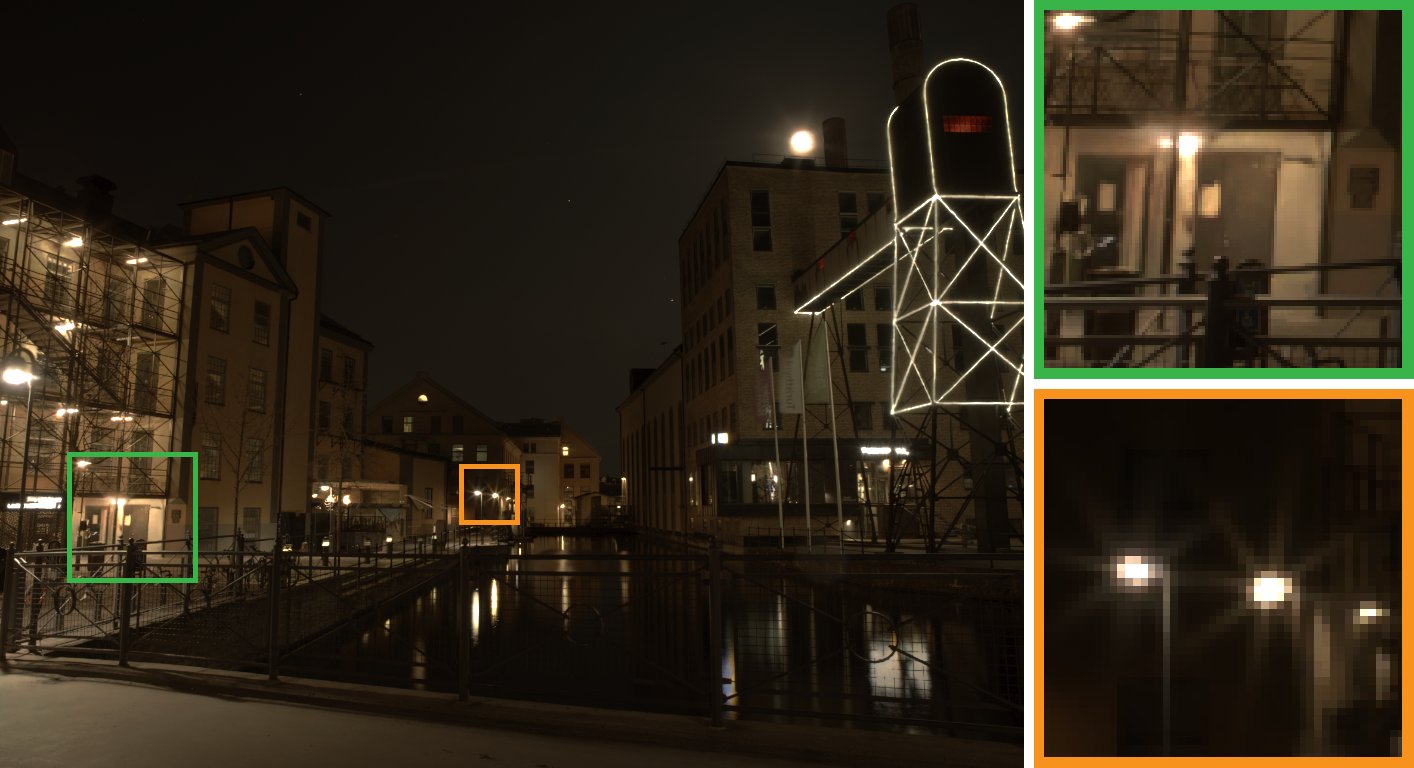}}
	\subfigure[Ground truth]{\includegraphics[width=\ww\textwidth]{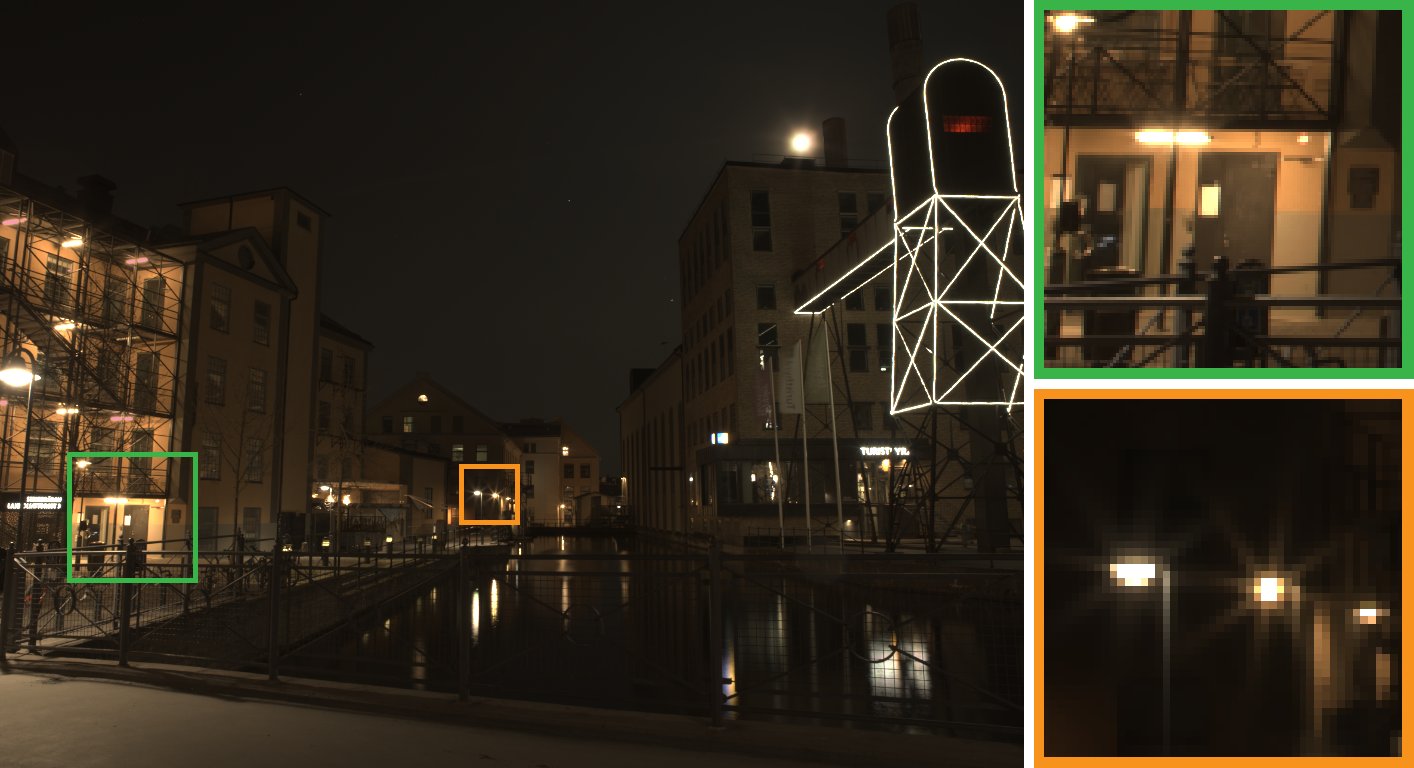}}
	\vspace{-10pt}
	\caption{\label{fig:itmos} Comparison to some existing iTMOs. Since the iTMO results usually are calibrated for an HDR display, they have been scaled to the same range for comparison. Although the they can deliver an impression of increased dynamic range by boosting highlights, when inspecting the saturated image regions little information have actually been reconstructed. The CNN we use can make a prediction that is significantly closer to the true HDR image.}
	\vspace{\belowfigspace}
\end{figure}

\begin{figure*}
	\vspace{5pt}
	\newcommand\ww{0.246}
	\centering
	\includegraphics[width=\ww\textwidth]{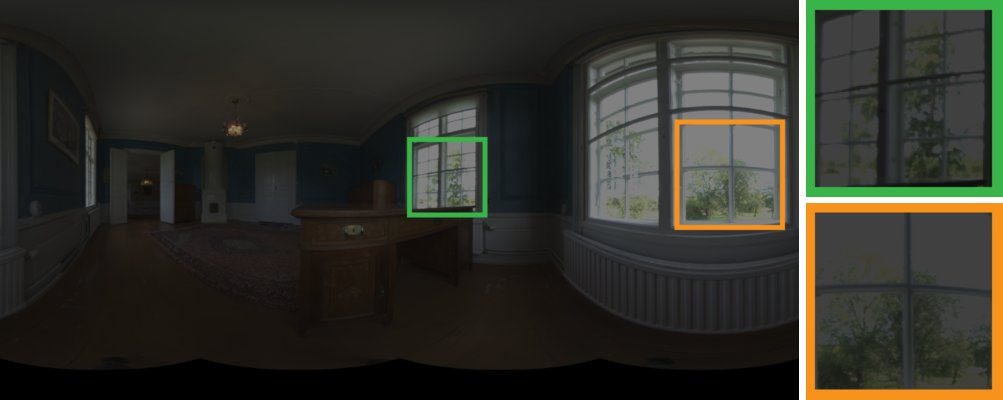}
	\includegraphics[width=\ww\textwidth]{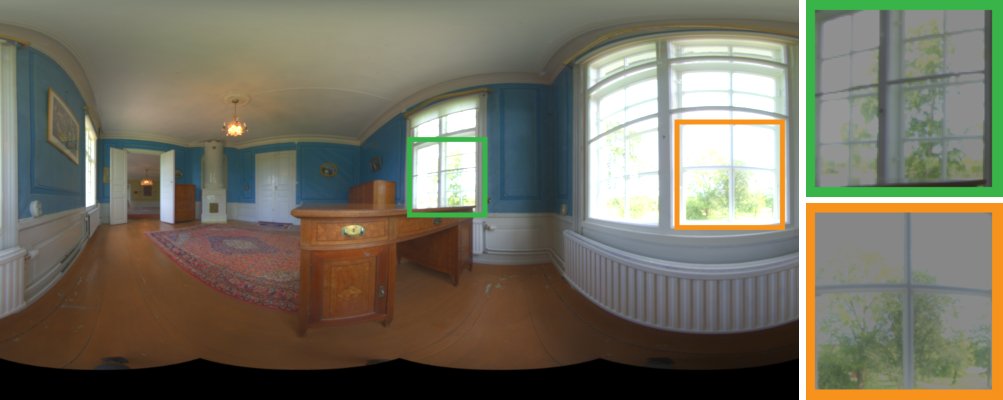}
	\includegraphics[width=\ww\textwidth]{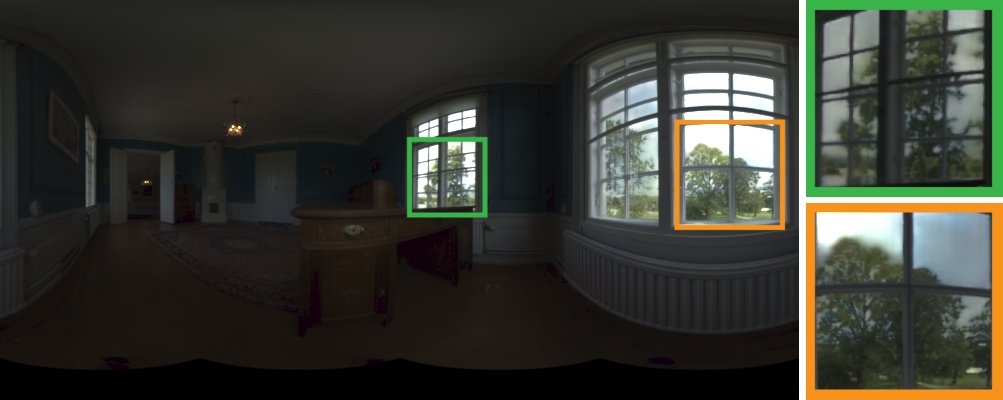}
	\includegraphics[width=\ww\textwidth]{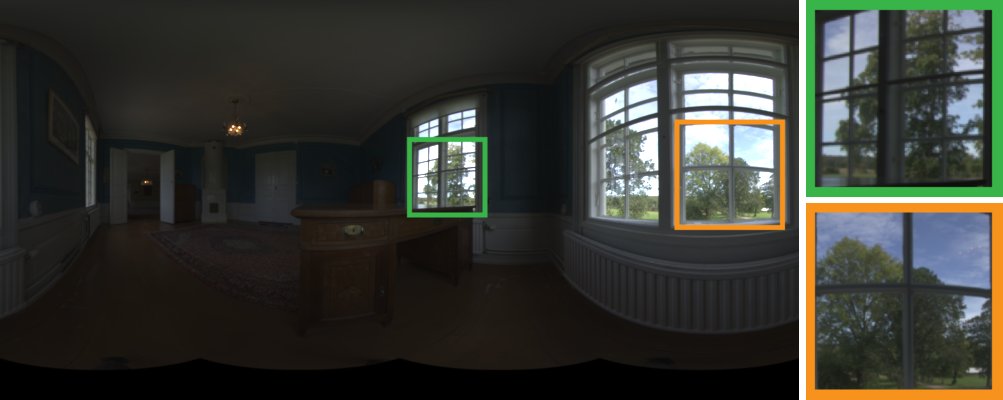}\\
	\vspace{-2pt}
	\subfigure[Input LDR]{\includegraphics[width=\ww\textwidth]{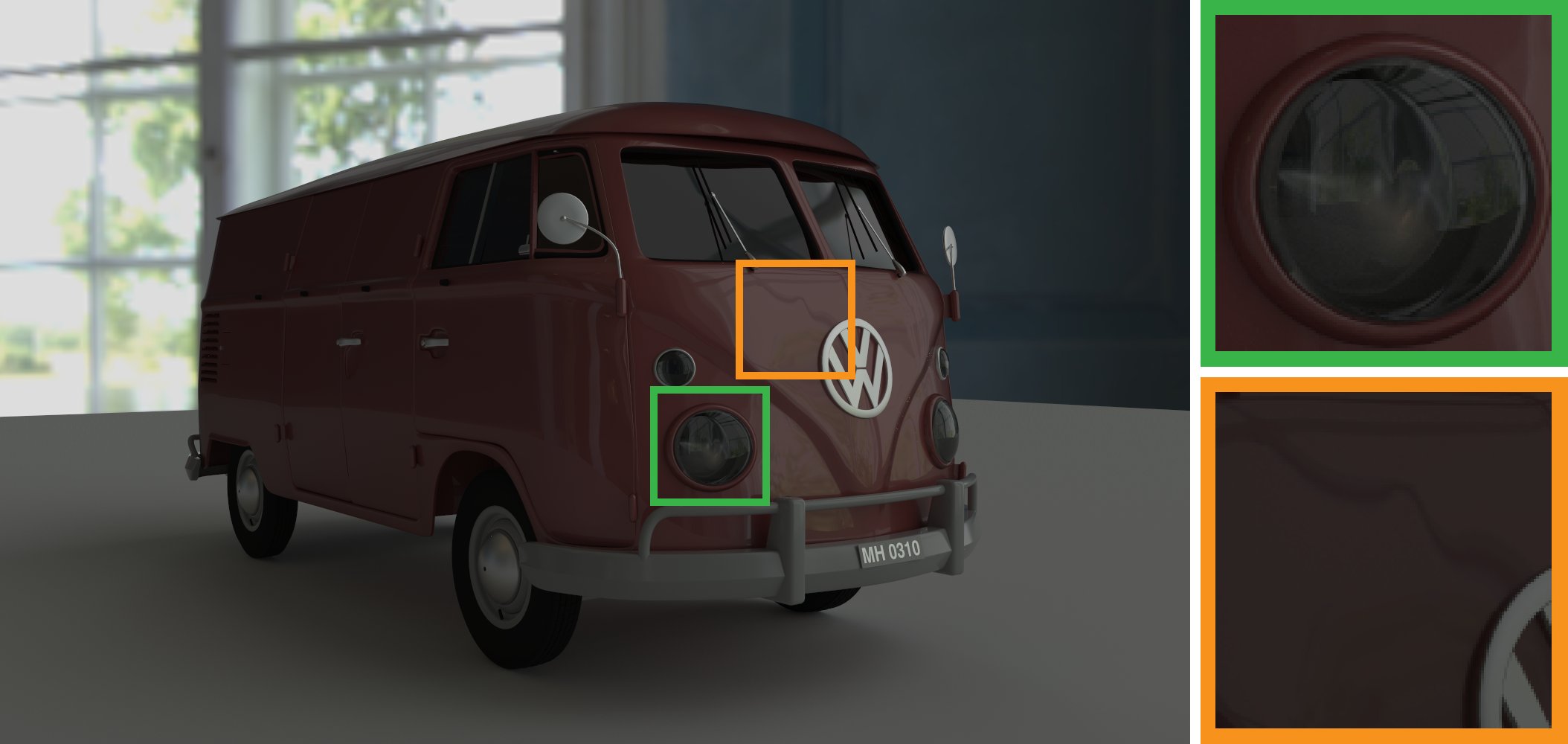}}
	\subfigure[iTMO, \cite{Banterle2008}]{\includegraphics[width=\ww\textwidth]{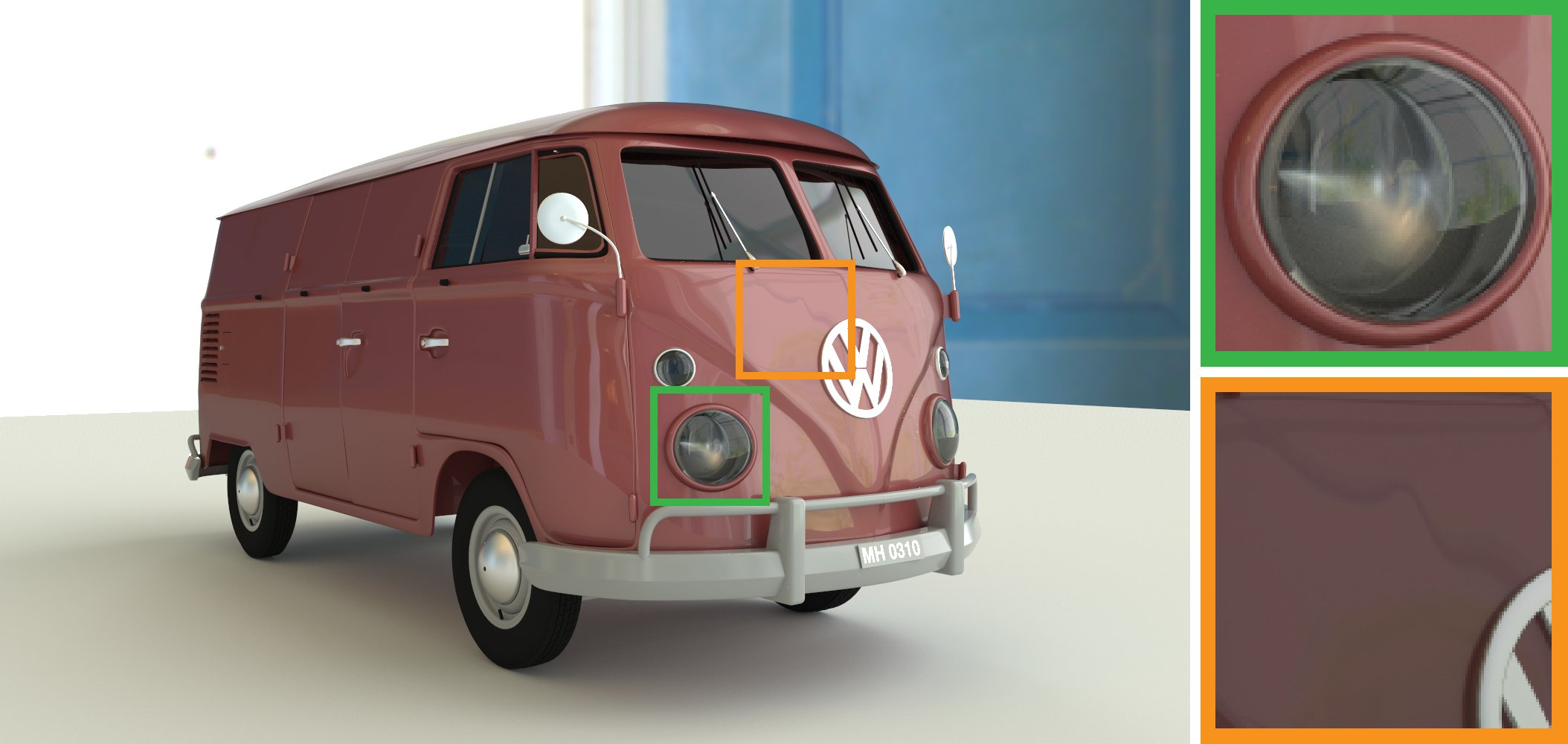}}
	\subfigure[Ours]{\includegraphics[width=\ww\textwidth]{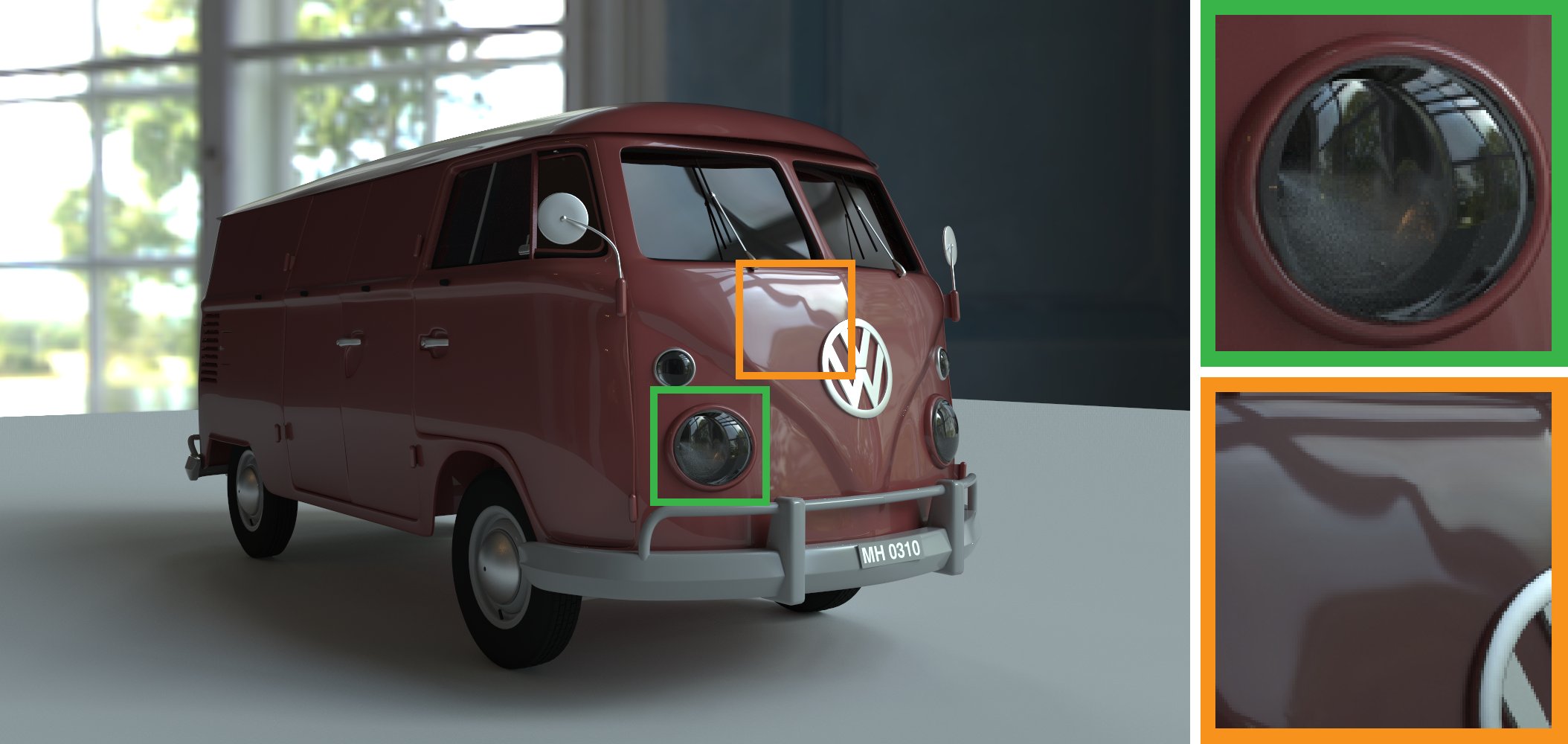}}
	\subfigure[Ground truth]{\includegraphics[width=\ww\textwidth]{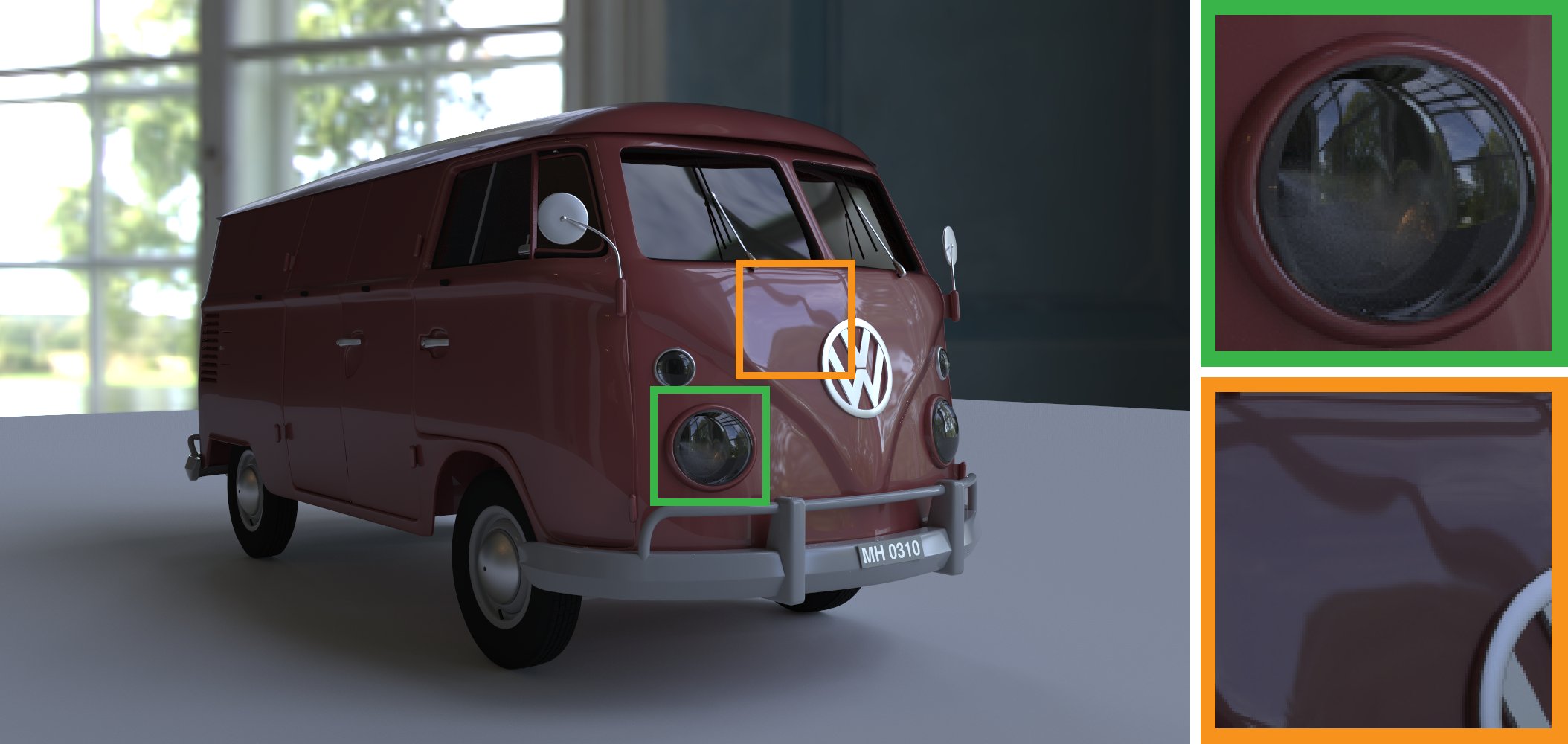}}
	\vspace{-12pt}	
	\caption{\label{fig:ibl} IBL using reconstructed highlights. The top row shows the panoramas that are used for IBL in the bottom row. Rendering with the LDR input gives a dark and undynamic result. The iTMO boosts brightness to alleviate the problems. With our reconstruction, although all details cannot be recovered in the large saturated areas of the windows, the estimated luminance enables a visually convincing rendering that is much closer to the ground truth.}
	\vspace{\belowfigspace}
\end{figure*}

\customsection{Comparison to iTMOs}
\figref{itmos} shows the HDR reconstruction compared to three existing methods for inverse tone-mapping. These are examples of local methods that apply different processing in saturated areas in order to boost the dynamic range of an LDR image. The results can successfully convey an impression of HDR when viewed on an HDR capable display. However, when inspecting the highlights, local information is not recovered by a na\"ive scaling of image highlights. With our CNN we are able to predict saturated regions based on a high level understanding of the context around the area, which makes it possible to actually recover convincing colors and structural content.

\customsection{Image based lighting}
An important application of HDR imaging is image based lighting (IBL), where the omnidirectional lighting of a particular scene is captured in a panorama and used for re-lighting of a computer graphics model. For this task it is of major importance that the entire range of luminances are present within the image, in order to convey a faithful and dynamic rendering. \figref{ibl} shows a panorama from an indoor scene, where the majority of illuminance is due to two windows. In the LDR image the windows are saturated, and the result when used for IBL is overall dark and of low contrast. The iTMO by Banterle \etal \citeyear{Banterle2008} can accomplish a more dynamic rendering by boosting image highlights, although the result is far from the ground truth. With our learning-based method, the reconstructed panorama shows some loss of details in the large saturated areas of the windows, but the illuminance is estimated convincingly. This makes it possible to render a result that is very close to the ground truth.


\begin{figure*}[]
	\centering
	\includegraphics[width=\textwidth]{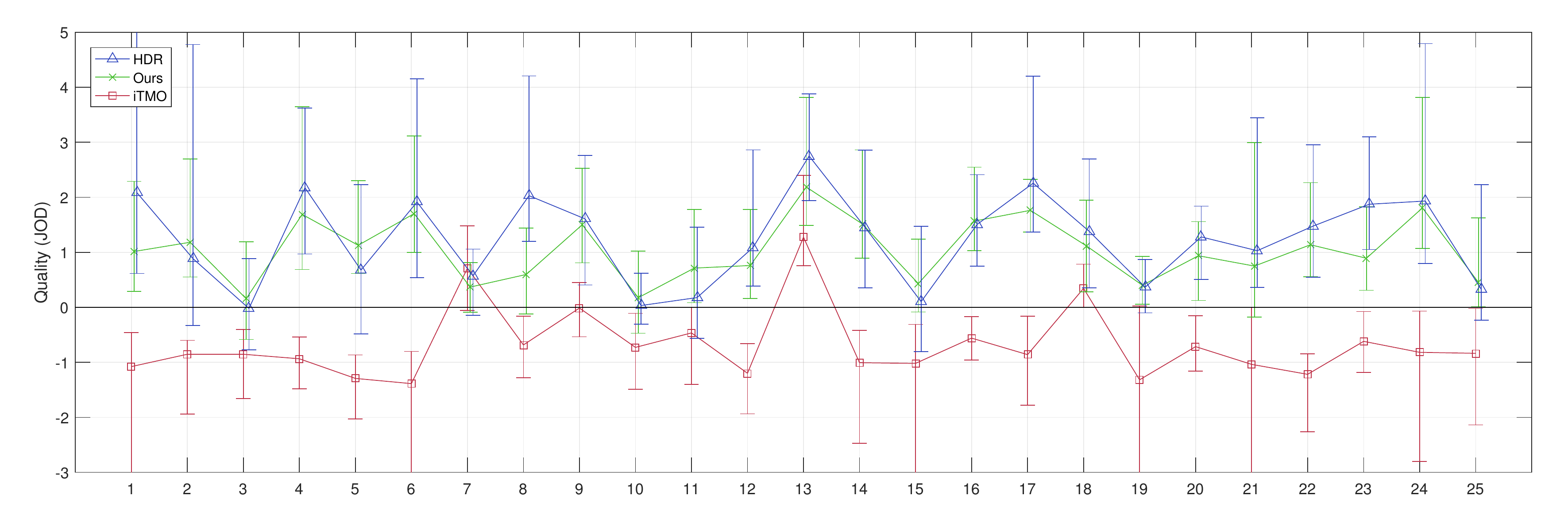}
	\vspace{-15pt}
	\caption{\label{fig:eval_pwc} Results of the subjective quality experiment. The error bars represent 95\% confidence intervals computed by bootstrapping. All values on the JOD scale are relative to the LDR images. Negative scores indicate a lower perceived naturalness for the iTMO technique when compared with LDR images. The output of our CNN-based HDR reconstruction method surpasses LDR and is comparable to the original HDR images in most cases.}
	\vspace{\belowfigspace}
\end{figure*}

\section{Evaluation}\label{sec:evaluation}
In order to assess the perceived visual quality of our novel HDR reconstruction, we performed a subjective pairwise comparison experiment. 15 participants took part in the experiment, aged 19 -- 40 with normal or corrected-to-normal full color vision. 

\customsection{Setup}
We used a calibrated projector-based HDR display, similar to the one described by Seetzen et al. \citeyear{Seetzen:2004:HDR:1015706.1015797}, at a viewing distance of ~$80\,cm$. The display consisted of an Acer P1276 $1024\times768$ DLP projector with removed color wheel, and a 9.7" $2048\times1536$ iPad Retina display panel with removed backlight. The maximal and minimal luminance of the display was $5000\,cd/m^2$ and $0.1\,cd/m^2$, yielding a maximal contrast range of 50\,000:1. 

\customsection{Stimuli}
Each stimulus consisted of a pair of images identical in terms of content, but reproducing one of the four processing or display scenarios: (1) LDR image with its dynamic range clamped to that of a typical DSLR camera (10.5 stops); (2) ground truth HDR image; (3) iTMO technique by Banterle \etal \citeyear{Banterle2008}, shown to perform the best in the evaluation \cite{Banterle2009}; (4) output from our CNN-based HDR reconstruction (Pre-train + I/R loss). To prevent overall image brightness from affecting the outcome of the experiment, we fixed the luminance of the 90$^{th}$ percentile pixel value for all methods to 180\,cd/m$^2$. To avoid bias in the selection of images, we used a randomly selected sample of 25 images from a pool of the 95 images in the test set.

\customsection{Task} We used a two-alternative forced choice experiment, where in each trial the participants were shown a pair of images side-by-side. The task was to select the image that looked more \emph{natural}. Our definition of natural involved ``the depth and vividness that bears most resemblance to a view that you could experience looking through a window''. Participants were given unlimited time to look at the images and to make their decisions. Before each session, participants were briefed about their task both verbally and in writing, followed by a short training session to gain familiarity with the display equipment and the task. We used a full pairwise comparison design, in which all pairs were compared. Each observer compared each pair three times, resulting in 450 trials per observer. The order of the stimuli as well as their placement on the screen were randomized.

\customsection{Results} The result of the pairwise comparison experiment is scaled in just-objectionable-differences (JODs), which quantify the relative quality differences. The scaling is performed using publicly available software\footnote{Pairwise comparison scaling software for Matlab: \url{https://github.com/mantiuk/pwcmp}}, which formulates the problem as a Bayesian inference under the Thurstone Case V assumptions, and uses a maximum-likelihood-estimator to find relative JOD values. Unlike standard scaling procedures, the Bayesian approach robustly scales pairs of conditions for which there is unanimous agreement. Since JOD values are relative, we fix the starting point of the JOD scale at 0 for the LDR images. When two points are a single unit apart on the JOD space, approximately 75\% of the population are expected to perceive an objectionable quality difference.

\begin{figure}
	\newcommand\ww{0.48}
	\centering
	\includegraphics[width=\ww\textwidth]{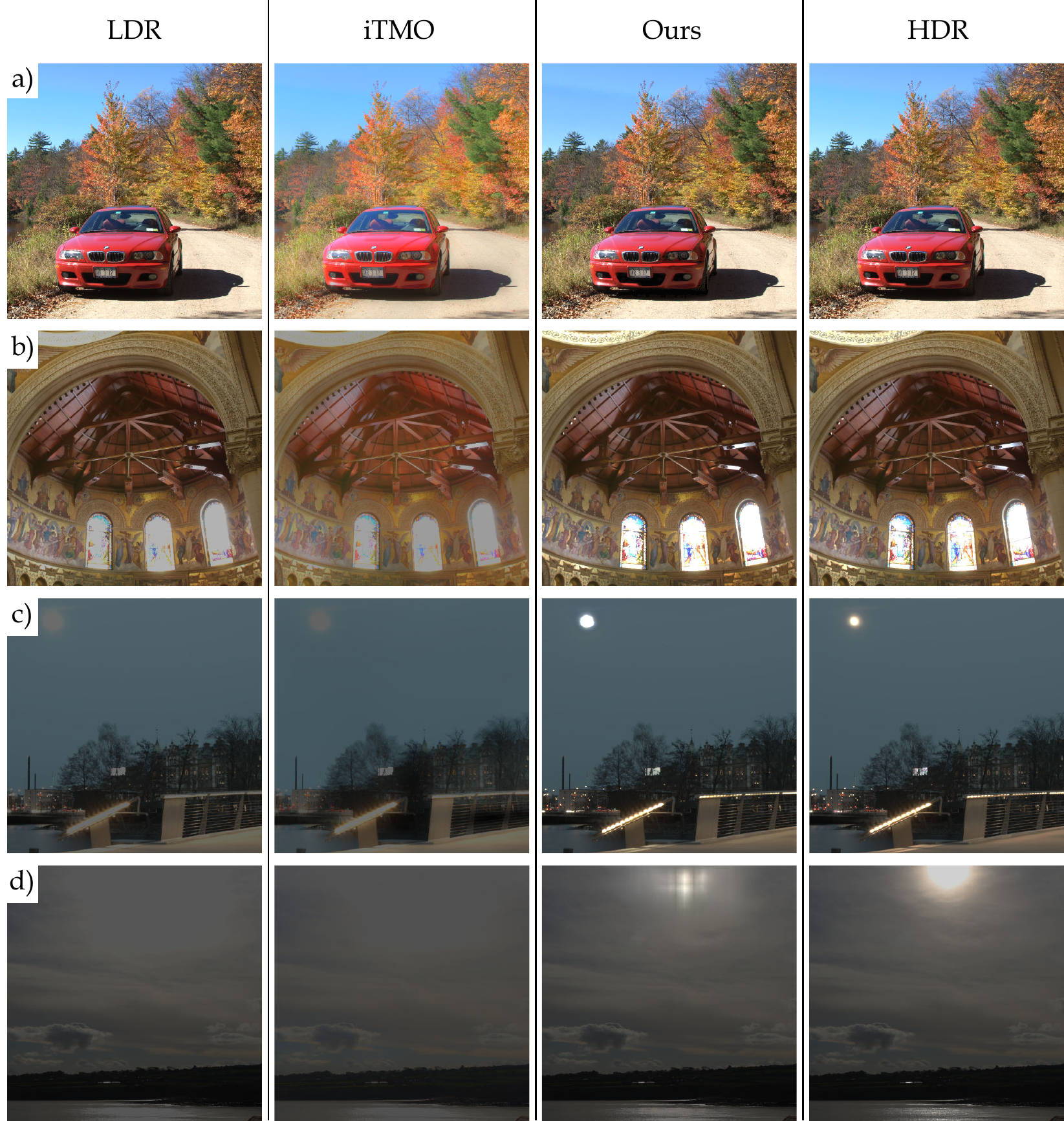}
	\caption{\label{fig:eval_images} Examples of typical images in the subjective experiment. a) \emph{image 19} demonstrates how the inaccurate colors of iTMO reduces perceived realism. b) \emph{image 21} with a successful HDR reconstruction. c) \emph{image 4} shows how even an inaccurate estimate of the highlight luminance still produces plausible results. d) \emph{image 8} is an example of unsuccessful reconstruction, producing artifacts that heavily affect the perceived naturalness. }
	\vspace{\belowfigspace}
\end{figure}

The results of the subjective quality experiment are shown in \figref{eval_pwc}. Unexpectedly, the iTMO technique by Banterle \etal was judged to be the least natural, even less so than the LDR images. This can be explained by the operator's attempt to inverse the camera response curve, which often results in reduced contrast and inaccurate colors. \figref{eval_images} row (a) shows the 19th image from the evaluation, where the effect of over-boosted colors can be easily observed.
As expected, LDR images were rated worse than the original HDR images in almost all cases. Most participants were mainly accustomed to standard display monitors, which, as reflected upon by some subjects during the unstructured post-experiment interview, might have affected their perception of ``natural''. With more exposure to HDR displays we expect the perceived quality of LDR images to drop in the future. 
According to the data, our CNN-based images are very likely to perform better than their original LDR counterparts. The number of times our CNN images were picked is slightly less but comparable to the original HDR images. \figref{eval_images} rows (b) and (c) illustrate two scenes, where the algorithm succeeds in estimating the spatial texture of high luminance areas, producing plausible results.
The performance of the CNN is scene-dependent, and can sometimes introduce artifacts in the highlights that affect perceived naturalness. 
One example is depicted in \figref{eval_images} row (d), where the artificial texture of the sun is immediately obvious. Overall, the CNN-based reconstruction improves the subjective quality as compared to the input LDR images, which is not always the case for the state-of-the-art iTMO techniques. The reconstructed images are in most cases comparable to ground truth HDR images, as evidenced by the  quality differences of less than 1 JOD.

\section{Conclusion and Future Work}\label{sec:conclusion}
HDR reconstruction from an arbitrary single exposed LDR image is a challenging task. To robustly solve this problem, we have presented a hybrid dynamic range autoencoder. This is designed and trained taking into account the characteristics of HDR images in the model architecture, training data and optimization procedure. The quality and versatility of the HDR reconstruction have been demonstrated through a number of examples, as well as in a subjective experiment.

\begin{figure}[t]
	\newcommand\ww{0.155}
	\centering
	\includegraphics[width=\ww\textwidth]{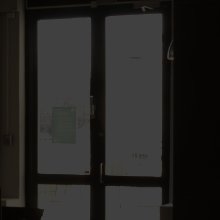}
	\includegraphics[width=\ww\textwidth]{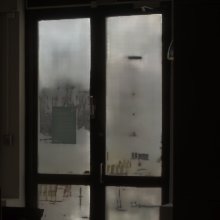}
	\includegraphics[width=\ww\textwidth]{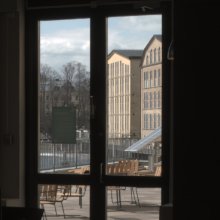}\\
	\vspace{-2pt}
	\subfigure[Input]{\includegraphics[width=\ww\textwidth]{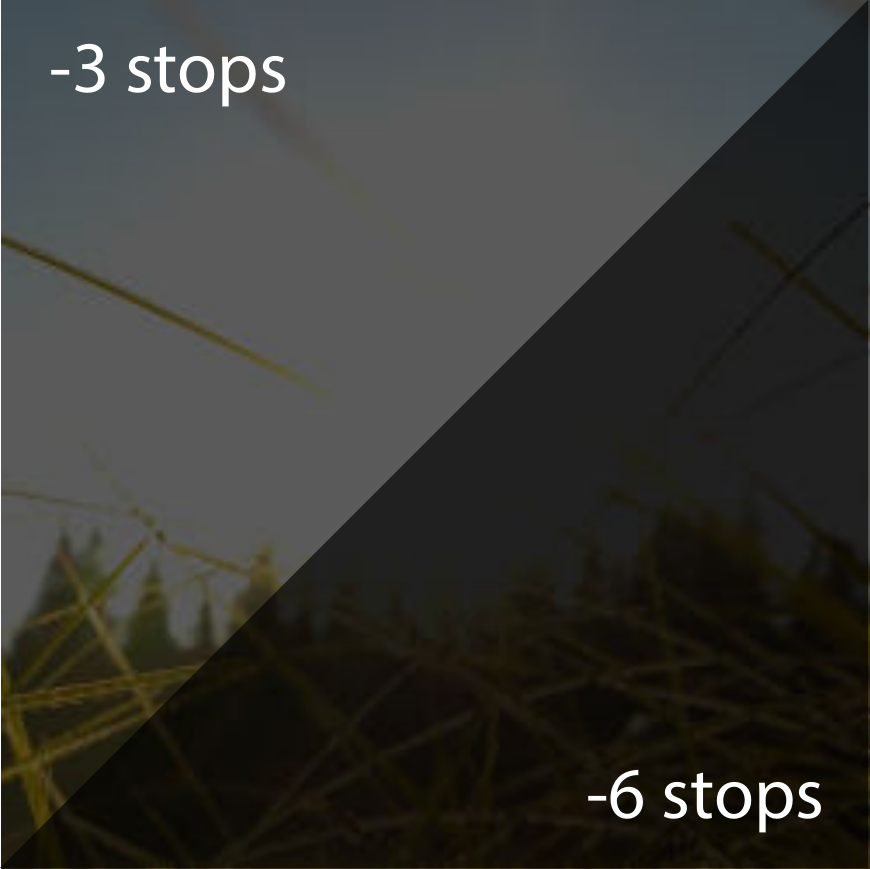}}
	\subfigure[Reconstruction]{\includegraphics[width=\ww\textwidth]{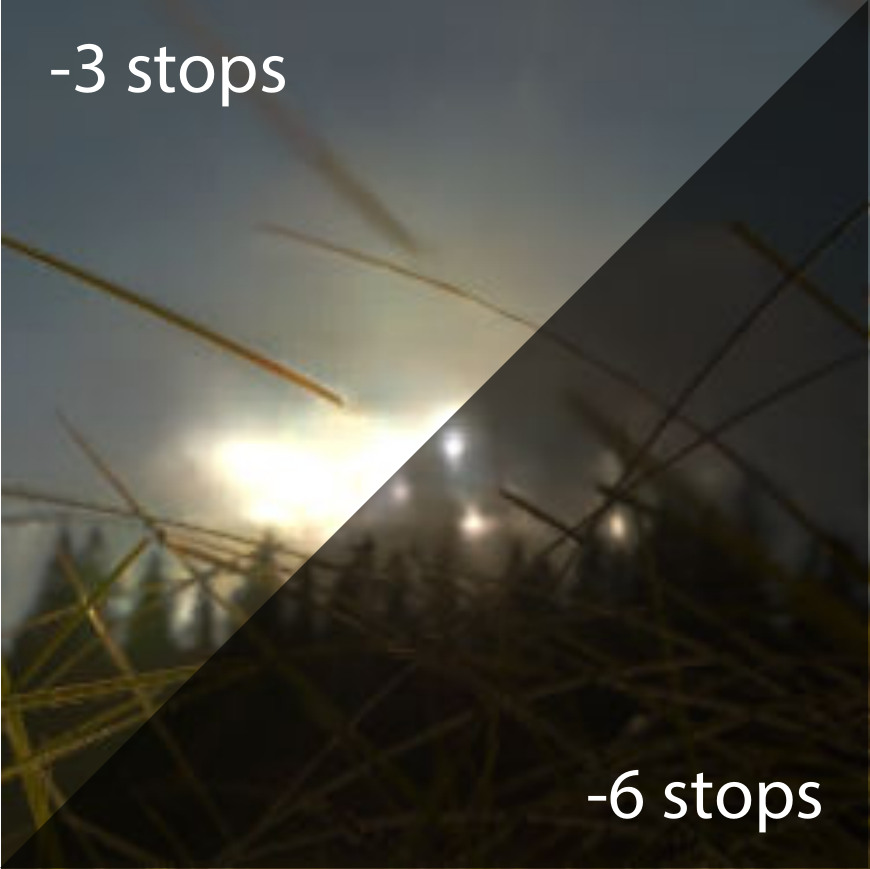}}
	\subfigure[Ground truth]{\includegraphics[width=\ww\textwidth]{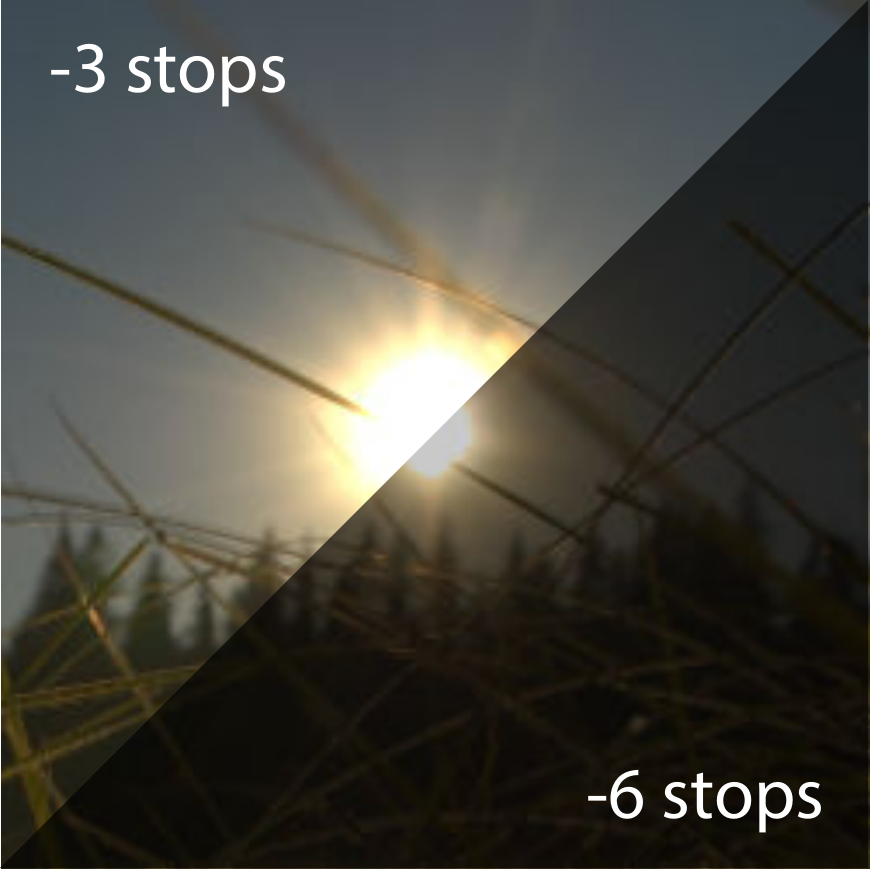}}
	\vspace{-10pt}
	\caption{\label{fig:failure} Zoom-ins of areas where reconstruction fails. (Top) A large fraction of the pixels are saturated, and the structures cannot be recovered properly. (Bottom) the intensity and shape of the sun are not estimated correctly.}
	\vspace{\belowfigspace}
\end{figure}

\customsection{Limitations} There is a content-dependent limit on how much missing information the network can handle which is generally hard to quantify. \figref{failure} shows two examples of difficult scenes that are hard to reconstruct. The first row has a large region with saturation in all color channels, so that structures and details cannot be inferred. However, illuminance may still be estimated well enough in order to allow for high quality IBL, as demonstrated in \figref{ibl}. The second row of \figref{failure} shows a situation where besides a similar loss of spatial structures, extreme intensities are also underestimated. This is also demonstrated in \figref{residual} with the intense spotlight. The plot also shows that the problem can be alleviated by altering the illuminance weight $\lambda$ in \eqnref{ir_loss}. However, underestimation of highlights is also an inherent problem of the training data. 
Some of the HDR images used to create the training data show saturated pixels in high intensity regions. For example, the sun in \figref{failure} is saturated in the ground truth HDR image.

There is also a limitation on how much compression artifacts that can be present in the input image. If there are blocking artifacts around highlights, these will impair the reconstruction performance to some extent.

\customsection{Future work} Recovering saturated image regions is only one of the problems in reconstructing HDR images. Another, less prominent issue is quantization artifacts, which can be alleviated using existing methods \cite{Daly2003}. However, we believe that bit-depth extension also can be achieved by means of deep learning. For this purpose existing architectures for compression artifact reduction \cite{Svoboda2016} or super resolution \cite{Ledig2016} are probably better suited.

The complementary problem of reconstructing saturated pixels, is the recovery of dark regions of an image, that have been lost due to quantization and noise. This problem is also significantly different from ours in that noise will be a main issue when increasing the exposure of an image.

Another direction for future work is to investigate how to improve reconstruction of images that are degraded by compression artifacts. The most straightforward solution would be to augment the training data with compression artifacts. However, this also runs the risk of lowering the quality for reconstructions of images without any compression applied.

Finally, although recent development in generative adversarial networks (GAN) \cite{Goodfellow2014,Radford2015} shows promising result in a number of imaging tasks \cite{Pathak2016,Ledig2016}, they have several limitations. An important challenge for future work is to overcome these, in order to enable high-resolution and robust estimation.

\begin{acks}
The authors would like to thank Francesco Banterle for the invaluable discussions and help with inverse tone-mapping operators, and the anonymous reviewers for helping in improving the manuscript. This work was funded through \grantsponsor{}{Link\"oping University Center for Industrial Information Technology (CENIIT)}{}, the~\grantsponsor{}{Swedish Science Council}{} through Grant~\grantnum{}{2015-05180}, and the~\grantsponsor{}{Wallenberg Autonomous Systems Program (WASP)}{}.
\end{acks}


\appendix
\section{Data augmentation}\label{app:augmentation}
In this appendix we specify the details of the virtual camera used in \secref{hdr_db} for augmentation of HDR images.

\customsection{Random cropping}
For each mega-pixel of HDR data, $N$ sub-images are selected at random positions and sizes.  For the trainings we perform, we choose $N=10$, which results in a final training set of $\sim\!\!125$K images. The sizes are drawn uniformly from the range $[20\%,60\%]$ of the size of an input image, followed by bilinear resampling to $320$x$320$ pixels. Since $320$ pixels corresponds to $20\%$--$60\%$ of the original images, the training is optimized to account for images that are between $320/0.6 = 533$ and $320/0.2 = 1600$ pixels.

\customsection{Exposure}
The exposure of each cropped image is set so that clipping removes a fraction $v$ of the image information. The fraction is uniformly drawn in the range $v\in[0.05,0.15]$. To accomplish this, $v$ is used to define an exposure scaling $s$,
\begin{equation}
s = \frac{1}{\hdrp_{th}}, \,\, s.t. \sum_{i = \hdrp_{min}}^{\hdrp_{th}}{p_{\hdr}(i)} = 1-v,
\end{equation}
where $p_{\hdr}$ is the histogram of the HDR image $\hdr$.  Thus, $\hdrp_{th}$ is the $1-v$ percentile, and this defines the scaling $s$ which is applied to the image in order to remove $5-15\%$ of the information when clipping is performed (see \eqnref{clip}).

\begin{figure}
	\centering
	\vspace{-2pt}
	\subfigure[\cite{Grossberg2003}]{\includegraphics[width=0.495\linewidth]{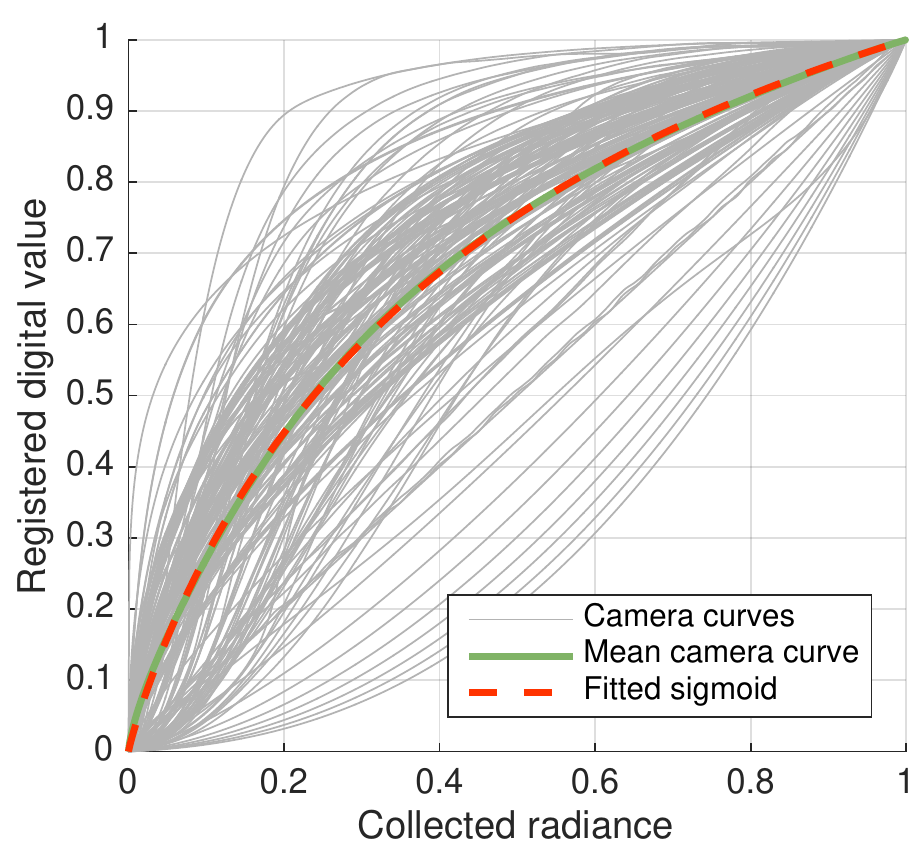}}
	\subfigure[Random sigmoid curves]{\includegraphics[width=0.495\linewidth]{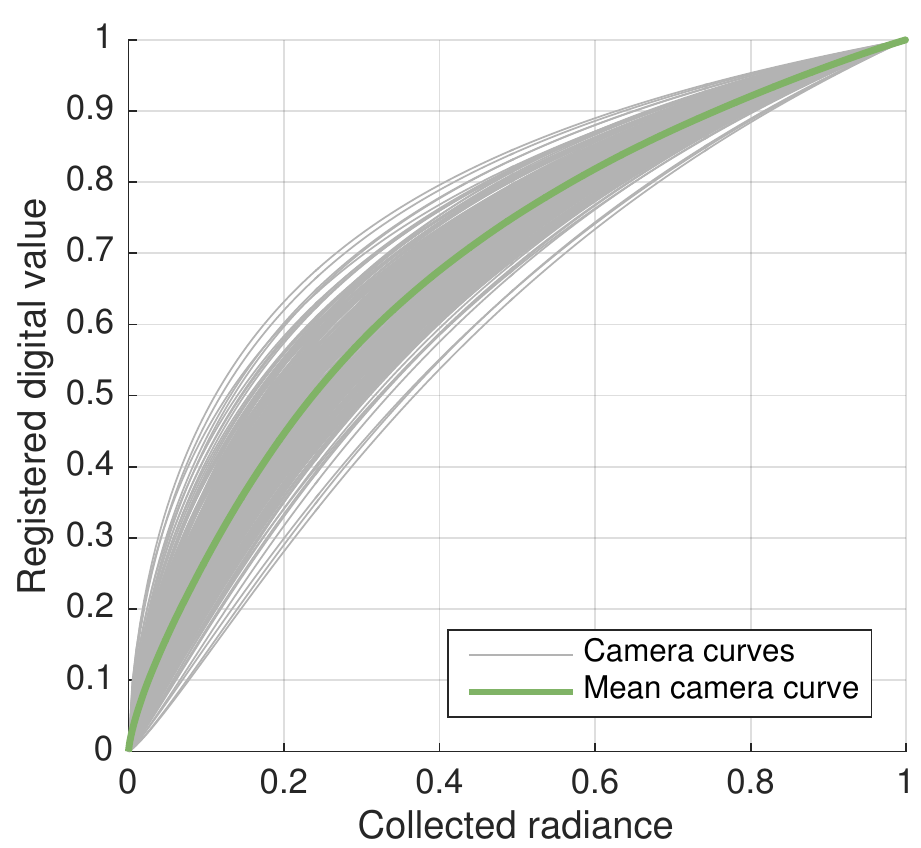}}
	\vspace{-5pt}
	\caption{\label{fig:cc} A sigmoid function fits well to the mean of the collected camera curves by Grossberg and Nayar \citeyear{Grossberg2003} (a). We simulate camera curves from a normal distribution centered at the fitted parameters (b).}
\end{figure}

\customsection{Camera curve}
To approximate different camera curves we use a parametric function, in form of a sigmoid,
\begin{equation}
\cc(\hdrp_{i,c}) = (1+\sigma) \frac{\hdrp_{i,c}^n}{\hdrp_{i,c}^n + \sigma}.
\label{eqn:cc}
\end{equation}
The scaling $1+\sigma$ is used to normalize the output so that $\cc(1) = 1$. We fit the parameters $n$ and $\sigma$ to the mean of the database of camera curves collected by Grossberg and Nayar \citeyear{Grossberg2003}, where $n = 0.9$ and $\sigma = 0.6$ gives a good fit as shown in \figref{cc}. For random selection of camera curves in the training data preparation, we draw the parameters from normal distributions around the fitted values, $n \sim \mathcal{N}(0.9,0.1)$ and $\sigma \sim \mathcal{N}(0.6,0.1)$. As demonstrated in \figref{cc} this creates a continuous range that do not include extreme functions such as gamma curves with $\gamma > 1$.

\customsection{Other}
Augmentation in terms of colors is accomplished in the HSV color space, where we modify the hue and saturation channels. The hue is altered by means of adding a random perturbation $\tilde{h} \sim \mathcal{N}(0,7)$. The same is done for the saturation, but with a narrower distribution, $\tilde{s} \sim \mathcal{N}(0,0.1)$.

Finally, a small amount of additive Gaussian noise is injected, with a standard deviation randomly selected in the range $\sigma\in[0,0.01]$. Also, images are flipped horizontally with a probability of $0.5$. The processed linear images $\hdr$ represent the ground truth data, while the inputs for training are clipped at $1$ and quantized, 
\begin{equation}
\ldrp_{i,c} = \lfloor 255 \min(1,\cc(\hdrp_{i,c})) + 0.5 \rfloor / 255.
\label{eqn:clip}
\end{equation}

\bibliography{hdrcnn} 

\end{document}